\documentclass[10pt, letterpaper]{article}

\usepackage[preprint]{corl_2026} 

\usepackage{pratik}
\usepackage[acronym]{glossaries}
\usepackage{tikz-timing}
\usetikzlibrary{arrows.meta,decorations.pathreplacing}
\usepackage{caption}

\lstdefinestyle{promptstyle}{
    basicstyle= \color{gray},
    frame=single,
    rulecolor=\color{black},
    breaklines=true,
    breakatwhitespace=true,
    columns=fullflexible,
    keepspaces=true,
    showstringspaces=false,
}

\usepackage[color=gray!10, size=tiny]{todonotes}
\definecolor{c1}{HTML}{E83A2E}   
\definecolor{c2}{HTML}{F5A623}   
\definecolor{c3}{HTML}{2E7D4F}   
\definecolor{c4}{HTML}{E91E8C}   
\definecolor{c5}{HTML}{7B61C4}   
\definecolor{c6}{HTML}{00AEEF}   
\definecolor{c7}{HTML}{2C2C2C}   

\begin{document}

\clearpage

\title{
OctoSense: Self-Supervised Learning for\\Multimodal Robot Perception\\
}

\author{Anthony Bisulco$^{1}$, Jeremy Wang$^{1}$,  Kostas Daniilidis$^{1}$, Randall Balestriero$^{2}$, Pratik Chaudhari $^{1}$\\
$^{1}$General, Robotics, Automation, Sensing and Perception (GRASP) Laboratory,\\
University of Pennsylvania\\
$^{2}$ Computer Science Department, Brown University\\
{\tt\small\{abisulco,jerewang,kostas,pratikac\}@upenn.edu, rbalestr@brown.edu} }%
\maketitle


\begin{abstract}
We present OctoSense, an open-source sensor platform with stereo RGB and event cameras, LiDAR, a thermal camera, an inertial measurement unit, RTK-corrected global positioning system, and proprioception (CAN bus data from a car, and joint angles for a quadruped robot). The eponymous OctoSense dataset contains 59 hours of time-synchronized driving data across different types of environments at different times of the day, including situations with highly degraded sensors. We demonstrate multi-modal self-supervised learning using such real-world robotics data, where sensors have different representations, frequencies, latencies and noise. Our approach, a ``late-fusion'' masked autoencoder, (i) uses modality-specific tokenizers to account for different spatiotemporal characteristics of these sensors, and (ii) caches modality-specific tokens at inference time to process new measurements as they come. This architecture (i) is fast (6.68 ms and 112 ms on NVIDIA 5090 and Orin NX respectively, to compute the representation), (ii) performs better than existing image-only foundation models on tasks such as estimation of optical flow, depth, semantic segmentation, and ego-motion (translation, rotation, and steering angle), and (iii) predicts robustly at nighttime or in situations where sensory data is degraded. See our project page for links to the dataset, code, and supplementary videos: \url{https://abisulco.com/octosense/}. 

\end{abstract}

\section{Introduction}

Vision-based foundation models trained using self-supervised learning~(SSL) such as DINO~\cite{caron2021emerging}, SigLIP~\cite{Zhai2023SigmoidLF, Tschannen2025SigLIP2M}, and Hiera~\cite{Ryali2023HieraAH} have transformed robotics.
However, in real-world robotic settings, no single sensor suffices~\cite{Liu2024ARO}.
Global-shutter cameras work poorly in situations with low light, high dynamic range, and rapid motion.
LiDAR provides accurate but very sparse depth at low frame rates, with poor semantics. And so on... Different sensors have different sensing modalities, angular resolutions, data types and information rates. They respond differently to weather degradation, noise, and packet loss. And they also fail in different ways. Effective robot perception requires representations that are resilient to these failures.
This is perhaps why the Global Robotics Technology Roadmap 2025--2035~\cite{christensen2026global} notes: ``The defining enabling technology across all field applications is multi-sensor fusion robust to environmental degradation---the problem that keeps the most capable field robots indoors''.
SSL has focused on visual and/or textual data. And although there is a growing literature on data from depth~\cite{bachmann2022multimae} or inertial~\cite{Girdhar2023ImageBindOE} sensors, we do not have SSL methods that work with all typical sensors a robot uses.
Our contributions towards addressing this gap are as follows.

\textbf{1. OctoSense hardware and dataset.}
A central challenge in multi-modal robot representation learning is the lack of large datasets with diverse sensors. To address this gap, we develop an open-source hardware platform called OctoSense that provides time-synchronized data from stereo RGB and event cameras, a thermal camera, LiDAR, an inertial measurement unit (IMU), RTK-corrected GPS, and proprioception information from the CAN bus of a vehicle and joint angles of a quadruped robot. We release 59 hours of data from OctoSense. This is one of the largest datasets of its kind. It covers urban, suburban, and rural environments across a mix of highway, residential, and urban driving; recordings span sunrise, daytime, sunset, and nighttime conditions, and include situations where sensors are degraded due to sunflares, dropped packets, etc.

\textbf{2. Multi-modal masked autoencoder.}
We account for different spatiotemporal characteristics
of LiDAR, event cameras and IMUs to develop modality-specific tokenizers which feed a shared transformer-based masked autoencoder (MAE). At inference time, we cache modality-specific features within a temporal window and only process new measurements. This makes it quite fast (6.68 ms and 112 ms on NVIDIA 5090 and Orin NX respectively, for computing the representation).

\textbf{3. Performance on downstream tasks.}
On tasks such as estimation of optical flow, depth, semantic segmentation and ego-motion (relative translation and rotation, linear and angular velocity, and steering position), we show that our multi-modal MAE is significantly better than existing image-only SSL approaches. Depth error improves by about 1.65 m and optical flow end-point error by about 7.16 pixels during daytime; segmentation is about on par.
Our approach can perform ego-motion tasks essentially perfectly.
Improvements are larger in nighttime data and when sensors are degraded.

\section{Related Work}

SSL has been instrumental in building highly general representations trained on unlabeled data. These techniques have led to visual foundation models such as DINO~\cite{caron2021emerging, oquab2023dinov2, simeoni2025dinov3}, Perception Encoder~\cite{Bolya2025PerceptionET}, SigLIP~\cite{Zhai2023SigmoidLF, Tschannen2025SigLIP2M}, SAM~\cite{Ravi2024SAM2S} that are the state of the art in computer vision today.
There are a number of different pre-training objectives, such as self-distillation~\cite{caron2021emerging}, masked autoencoders~\cite{He2021MAE,Xie2021SimMIMAS, Devlin2019BERTPO} and contrastive alignment~\cite{Oord2018RepresentationLW,Chen2020ASF}.
SSL has also been extended beyond image modalities, e.g., to event cameras~\cite{Cao2026GenerativeEP, Klenk2022MaskedEM, Yang2023EventCD, das2025fast}, depth data/point clouds~\cite{Patel2026DeFMLF, pang2022masked, Yu2021PointBERTP3,Wang2020UnsupervisedPC,Xie2020PointContrastUP}, thermal cameras~\cite{Munir2021SSTNSD, zurn2024selfsupervised} and IMUs~\cite{Narayanswamy2024ScalingWF, Xu2021LIMUBERTUT}.
These existing efforts have demonstrated strong results on individual sensors. Our work builds upon a number of techniques and best practices in this literature, but our goal is to learn a fused representation from multiple sensors that inherits their individual strengths.

Multi-modal SSL has employed two key ideas~\cite{Zong2023SelfSupervisedML}. The first, contrastive alignment, anchors each modality to a shared embedding, e.g., CLIP~\cite{Radford2021LearningTV} and SigLIP for vision-language models, ImageBind~\cite{Girdhar2023ImageBindOE} for multiple modalities. Recent variants for images and text such as SigLIP 2~\cite{Tschannen2025SigLIP2M} and Perception Encoder~\cite{Bolya2025PerceptionET} observe that these embeddings are not sufficiently sensitive to local image structures~\cite{Wang2020DenseCL} and improve this aspect using masking and distillation objectives.
Our approach is based on the second key idea, namely masked multi-modal modeling. MultiMAE~\cite{bachmann2022multimae} uses a Dirichlet masking strategy for multiple input modalities, and 4M~\cite{Mizrahi20234MMM} builds upon this to use modality-specific tokenizers~\cite{bao2022beit,lu2022unified,Lu2023UnifiedIO2S}.
Some works on driving such as UniM$^2$AE~\cite{Zou2023UniM2AEMM} and CALICO~\cite{Sun2023CALICOSC} combine cameras and LiDAR to perform tasks such as  bird's eye view (BEV) map segmentation.
In this paper, we extend masked multi-modal modeling beyond image and text modalities to data from typical robotic sensors, and beyond the camera-LiDAR BEV setting to dense perception and egomotion tasks.

\clearpage   

\begin{figure}[p]
  \centering
  \includegraphics[width=0.9\linewidth,height=\textheight,keepaspectratio]{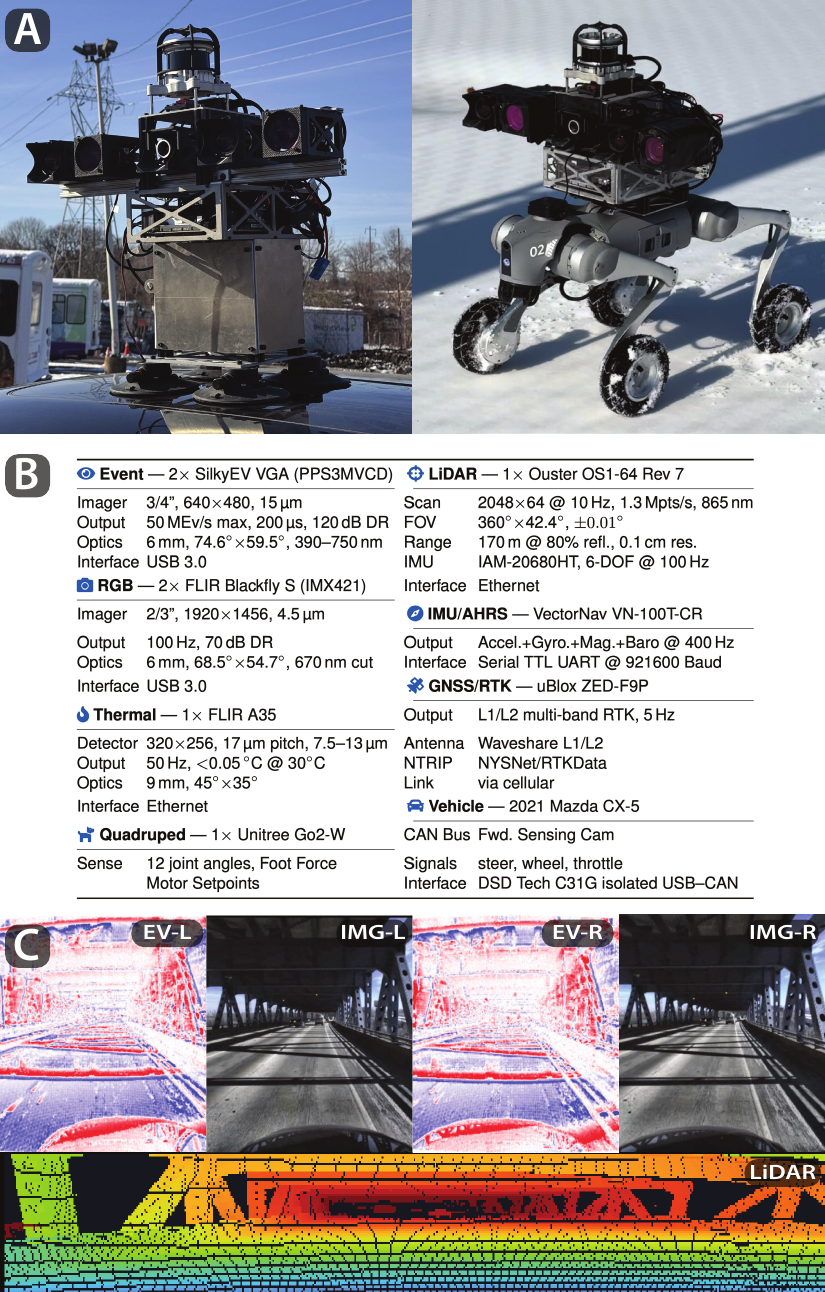}
  \caption{A) OctoSense has been deployed across a car and a Unitree Go2-W. This initial dataset release focuses on car driving sequences with a very small amount of data collected using the Unitree.
B) This platform is unique because it contains diverse sensors such as stereo RGB and event cameras, a thermal imager, LiDAR, IMU, GPS and CAN bus data. These sensors have very different data rates, frequencies, and information structures. C) Examples of time-synchronized data collected from OctoSense.}
  \label{fig:panel_1}
\end{figure}

\begin{figure}[p]\ContinuedFloat
  \centering
  \includegraphics[width=\linewidth,keepaspectratio]{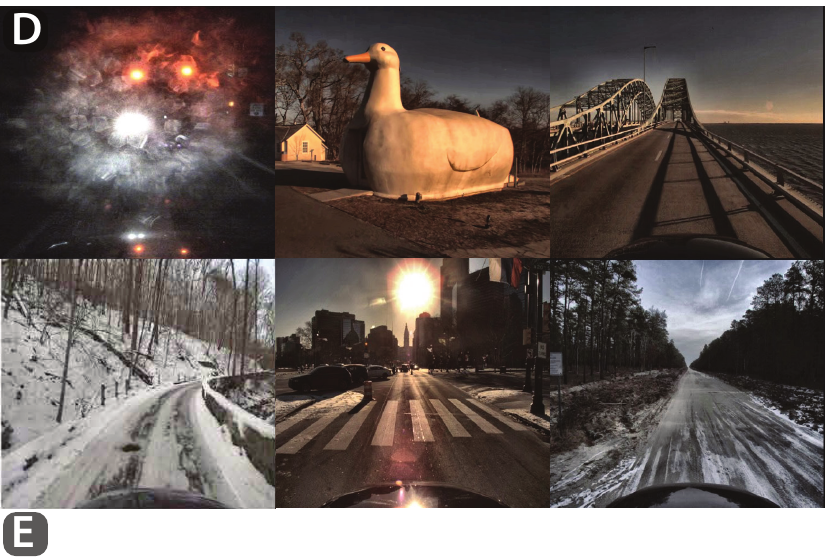}
  \par\vspace{1.0ex}
  \scalebox{0.83}{

{\small
\setlength{\tabcolsep}{4pt}%
\renewcommand{\arraystretch}{1.0}%
\fontfamily{phv}\selectfont
\begin{tabular}{@{}>{\raggedright\arraybackslash}p{3.0cm} @{\hspace{4pt}} l @{\hspace{4pt}} >{\raggedright\arraybackslash}p{3.5cm} p{5.0cm}@{}}
\toprule
\textbf{Dataset} & \textbf{Duration / Dist.} & \textbf{Sensors} & \textbf{Tasks} \\
\midrule
\multicolumn{4}{@{}l}{\textit{Autonomous Vehicles, long-term, off-road, and ground-robot datasets}} \\
\midrule
KITTI~\cite{Geiger2012AreWR}                  & 6\,h        & RGB, Gray., LiDAR, IMU, GPS                       & Depth, flow, stereo, 2D/3D det., track., seg., egomotion, lane det. \\
nuScenes~\cite{Caesar2019nuScenesAM}          & 5\,h 33\,min  & RGB, LiDAR, Radar, IMU, GPS, CAN                  & 3D det., track., seg., egomotion \\
Waymo Perc.~\cite{Sun2019ScalabilityIP}       & 6\,h 24\,min  & RGB, LiDAR                                        & 2D/3D det., track., seg., egomotion \\
AV2-Sensors~\cite{LidarArgo}                  & 4\,h 10\,min  & RGB, LiDAR                                        & 3D det., track., egomotion \\
Oxford RobotCar~\cite{Maddern20171Y1}         & 1000\,km    & RGB, LiDAR, GPS/INS                               & Egomotion \\
Boreas Road Trip~\cite{Lisus2026BoreasRT}     & 643\,km     & RGB, LiDAR/FMCW, Radar, IMU, GPS/INS, wheel enc.  & Egomotion \\
NCLT~\cite{CarlevarisBianco2016UniversityOM}  & 34\,h 54\,min & RGB, LiDAR, IMU, fiber gyro., GPS, wheel enc.     & Egomotion \\
TartanDrive 1.0/2.0~\cite{Triest2022TartanDriveAL,Sivaprakasam2024TartanDrive2M} & 12\,h & RGB, Gray., LiDAR, IMU, GPS, shock, wheel enc., Ctrl. & Egomotion, roughness \\
Ithaca365~\cite{DiazRuiz2022Ithaca365DA}      & $\sim$600\,km & RGB, LiDAR, GPS/INS                             & Amodal seg., 3D det., disparity \\
Comma2k19~\cite{Schafer2018ACI}               & 33\,h       & RGB, IMU, GPS, CAN                               & Egomotion \\
NVIDIA PAI~\cite{nvidia2025physicalai}        & 1700\,h     & RGB, LiDAR, Radar                                & 3D det., track., egomotion, reasoning \\
\midrule
\multicolumn{4}{@{}l}{\textit{Event-inclusive multimodal datasets}} \\
\midrule
DSEC~\cite{dsec}                              & 53\,min       & Events, RGB, LiDAR, IMU                          & Seg., disparity, flow, 2D det. \\
MVSEC~\cite{MVSEC}                            & 1\,h 08\,min  & Events, Gray., LiDAR, IMU                        & Depth, flow, egomotion \\
VECtor~\cite{vector}                          & 21\,min       & Events, Gray., Depth, LiDAR, IMU                 & Egomotion \\
ECMD~\cite{Chen2023ECMDAE}                    & 3\,h 21\,min  & Events, RGB, IR, LiDAR, IMU, GPS/INS             & Egomotion \\
M3ED~\cite{m3ed}                              & 3\,h 46\,min  & Events, RGB, Gray., LiDAR, IMU, GPS              & Seg., depth, egomotion \\
ViViD++~\cite{Lee2022ViViDV}                & 4\,h 31\,min  & Events, RGB-D, IR, LiDAR, IMU, GPS               & Egomotion \\
Prophesee 1MP~\cite{Perot2020LearningTD}      & 14\,h 39\,min & Events                                          & 2D det. \\
DDD17/20~\cite{Binas2017DDD17ED, Hu2020DDD20EE} & 51\,h     & Events, Gray.                                   & Steering angle \\
\midrule
\rowcolor{gray!12}
\textbf{OctoSense (Ours)} & \textbf{59\,h / 2474\,km} & \textbf{Events, RGB, IR, LiDAR, IMU, GPS, CAN} & Depth, flow, seg., egomotion, steering angle \\
\bottomrule
\end{tabular}}}
  \caption*{Figure~\ref{fig:panel_1} (continued). D) We have many sequences with degraded sensors during off-road and city driving. E) OctoSense is currently the largest multi-modal robotics dataset that includes event camera data. It contains 59 hours of driving data across 2474 km. Alongside NVIDIA's Physical AI, which came out concurrently with this work, OctoSense is also one of the largest driving datasets with dense sensory data. Sensors listed are those released as data streams in the dataset.}
\end{figure}

\clearpage   

\section{OctoSense Hardware and Dataset}
\label{sec:dataset}

OctoSense has eight sensors: stereo RGB and event cameras, LiDAR, a thermal camera, an inertial measurement unit (IMU) that measures accelerations, angular velocities, magnetic fields, and barometric pressure, RTK-corrected global positioning system (GPS), and the proprioception state.
See picture in~\cref{fig:panel_1}A.
\cref{fig:panel_1}B shows that these sensors have very different data rates, bandwidth requirements and resolutions.
OctoSense has a desktop-class CPU for onboard computation. A 24 V battery provides about one hour of operation per charge.
\cref{app:bom} provides a bill of materials.

\textbf{Engineering challenges and our solutions.} Some examples include online adaptation of exposure and sensor gain for RGB cameras, a bandpass filter on the event camera to block infrared interference from the LiDAR, foam insulation around the event cameras to prevent corrupted timestamps in cold weather, soft-mounting sensors to suppress LiDAR packet loss and IMU noise from vibrations, and splicing the car's camera connector for CAN access.
\cref{app:octo_detail} gives more details.
Many sensor drivers required substantial modifications, which we discuss in \cref{sec:software}. Composing drivers and data recorder as ROS 2 Zenoh composable nodes gives zero-copy transfer, which allows us to record those sensors at native rates without dropping any data. We built a web app to monitor system health during data collection, specifically to track per-sensor data rates and timestamps, and to start/stop data recording. OctoSense produces $\sim$1.7 GB/s of data; we therefore develop drivers that record compressed sensor streams, e.g., LiDAR/event packets and H.265~\cite{ITUT_H265} video without B frames for RGB and thermal cameras. This reduces the required bandwidth, by 21$\times$, to a more manageable 78.7 MB/s.

\textbf{Temporal synchronization.}
Sensors typically timestamp measurements using their individual clocks---with a different start time, frequency, latency and clock drift. However, tasks such as stereo vision or projecting LiDAR depth upon RGB images for obtaining ground-truth measurements require sensors that are synchronized to the same clock.
Let the timestamps of sensor $s$ be $(t^s_i)_{i \in \naturals}$. We built custom circuits to hardware-trigger sensors using a pulse per second~(PPS) signal, let us denote these by $(t^*_i)_{i \in \naturals}$.%
\footnote{If this is done naively, a single missed trigger signal due to hardware issues can cause the entire subsequent sensor stream to be out of sync. We therefore create a unique six-pulse sequence as a global identifier every four minutes and fifteen seconds, see~\cref{fig:pps-sync}. This way, we can unambiguously realign sensor data within each four-minute and fifteen second window during post-hoc clock synchronization.}
We post-process recordings to synchronize sensor clocks by fitting an affine transformation $t^s_i \mapsto t^*_i$ using Kalman smoothing, see~\cref{app:kalman_filt}.

\textbf{Calibration.}
We use a retro-reflective circle target centered on an AprilGrid~\cite{Olson2011AprilTagAR} for calibration, see \cref{fig:cal}.
The grid gives correspondences for RGB/event cameras while the retro-reflective circle gives a feature visible to both cameras and LiDAR.
We first reconstruct event frames using a second-order IIR filter~\cite{pfrommer2022frequencycam} and optimize for RGB and event camera intrinsics and extrinsics with Kalibr~\cite{rehder2016extending,furgale2013unified}.
We then solve for LiDAR-camera extrinsics using three types of correspondences: (i) reprojected circle-center error, (ii) circle radius computed using LiDAR points, and (iii) distance to the board. We recover the circle center in LiDAR data by fitting a plane and circle to the thresholded reflectivity, and in the camera by detecting AprilTags and solving PnP. LiDAR-camera extrinsics are initialized using these circle centers via RANSAC with the Kabsch–Umeyama algorithm~\cite{Kabsch1976ASF,Umeyama1991LeastSquaresEO} and refined using Levenberg–Marquardt~\cite{Levenberg1944AMF}.
We use Kalibr for obtaining camera-IMU extrinsics.%

\begin{wrapfigure}{R}{0.5\linewidth}
    \centering
    \includegraphics[width=\linewidth]{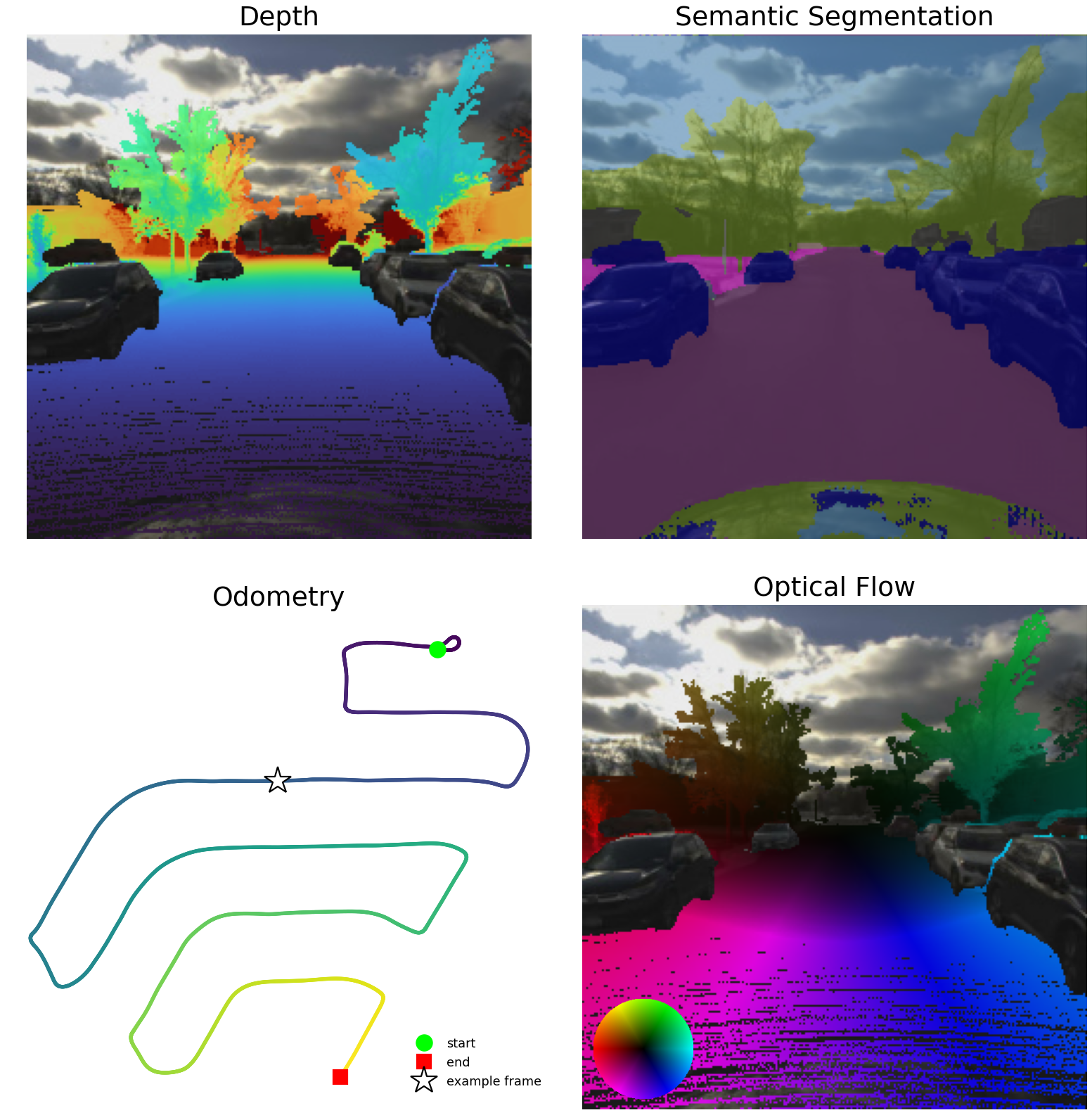}
    \caption{Ground-truth of different downstream tasks in OctoSense}
    \label{fig:gt_perception}
\end{wrapfigure}

\textbf{This first release of OctoSense data} covers urban, suburban, and rural environments across a mix of highway, residential, and city driving on Long Island and in Philadelphia, see~\cref{fig:gps_routes}. Recordings span sunrise, daytime, sunset, and nighttime conditions, with most sequences lasting for about 10 minutes. Altogether, OctoSense contains 59 hours and 8.4 TB of time-synchronized data.%
\footnote{
From 75 hours of recorded raw data, we removed parts where LiDAR packets were missing for more than 5 sec, event-camera timestamps or thermal data were corrupted, or data packets were dropped.}
\cref{fig:panel_1}D shows some examples. \cref{fig:panel_1}E shows that OctoSense is one of the largest datasets of its kind.
We build software to caption each 5-second window with Gemma 4~\cite{gemma4_2026} from three representative frames (see~\cref{sec:caption}), embed the caption with Qwen3-Embedding-8B~\cite{Zhang2025Qwen3EA}, and index embeddings with FAISS~\cite{douze2024faiss}. This is useful to explore the data using natural language.

\textbf{Ground-truth for odometry, depth, optical flow, and semantic segmentation.}
We use a LiDAR-inertial odometry method named RKO-LIO~\cite{malladi2025arxiv} to estimate the pose of the platform in $\mathrm{SE}(3)$, along with linear and angular velocities. This method also deskews each 360° LiDAR scan to transform points as if they were captured at a single timestamp. We accumulate points across 61 consecutive LiDAR scans (a window of 6 sec around a reference time) to obtain relatively dense ground-truth depth.
We handle independently moving objects separately. We use YOLO26-medium~\cite{Sapkota2025YOLO26KA} on the rectified RGB image closest to each LiDAR scan and remove points that correspond to vehicles or pedestrians.
Similarly, if some pixels have multiple depth measurements after projecting the accumulated points onto the rectified RGB image, we keep the minimum depth (this respects disocclusions).
In the accumulated point cloud, we prune 3D voxels that are not observed by at least $N$ different scans.
We use the odometry pose and LiDAR-camera extrinsics and intrinsics to project the accumulated points into a future RGB frame to compute per-pixel displacement as the optical flow induced by ego-motion. We obtain segmentation pseudo-labels by running a recent segmentation model, EoMT~\cite{kerssies2025eomt}, trained on Cityscapes on the rectified RGB images.


\section{Methods}
\label{sec:method}

We next discuss a multi-modal MAE architecture for sensors in OctoSense.
These sensors have different spatiotemporal characteristics: dense 2D arrays for RGB, a high-frequency point process for the event camera, an unordered and sparse point cloud for LiDAR and a rapidly varying multi-channel time series for the IMU.
This heterogeneity makes it challenging to build a fused representation.
Our approach has four key facets: (i) low-level input representations for these diverse sensory modalities, (ii) appropriate per-modality tokenization to ensure effective reconstruction, (iii) an MAE architecture (encoder and decoder) that combines multiple modalities into a shared representation, and (iv) task-specific probes that make predictions from that representation.

\subsection{Input representation}
\label{subsec:input_representation}

We undistort and rectify RGB images before building ViT-style patches~\cite{dosovitskiy2020image}. Event cameras produce an event stream $(t_i, u_i,p_i)_{i \in \naturals}$ for time $t_i$ and pixel coordinates $u_i \in \integers^2$ and polarity $p_i$ indicating the sign of intensity change at the pixel since the last such event~\cite{async_delta,ev_survey}. Events have a very high frequency, bandwidth and spatiotemporal noise, which makes it difficult to incorporate them directly into MAE architectures. After undistortion and rectification, we use a spatiotemporal filter to remove isolated events~\cite{delbruck2008frame}. We then use a bank of leaky integrators at each pixel with different bandwidths to obtain a multi-channel image-like representation \cite{gerstner2014neuronal, Lagorce2017HOTSAH, Bisulco2020FastMU}. See \cref{app:ev_cam}.%
\footnote{
We normalize each channel individually to account for the fact that filters with different bandwidths result in different statistics at each channel.
}
The LiDAR point cloud gives a 64 $\times$ 2048 range image which is cropped to a 90° field of view to get a 64 $\times$ 512 range image.
We convolve accelerometer and gyroscope data within a 1.6 sec time window and tokenize it using cross-attention pooling~\cite{jaegle2021perceiver} (see \cref{app:imu_arch}).

\subsection{Training tokenizers}
\label{subsec:tokentarget}

MAEs for RGB data are trained to minimize the mean squared error~(MSE)  between the predicted pixel intensities and those in the original RGB image. This leads to poor training in multi-modal settings because different modalities have different spatiotemporal statistics and token counts.
Works such as BEiT~\cite{bao2022beit}, Unified-IO~\cite{lu2022unified} and 4M~\cite{Mizrahi20234MMM} mitigate this issue by training a tokenizer where each patch is represented by a token supported on a finite set. Then, instead of minimizing the MSE on pixel intensities, these methods minimize the discrepancy between the predicted token for masked patches and the ground-truth token of that patch. We follow this approach, but compute the loss on all patches rather than only the masked ones, which V-JEPA 2.1~\cite{MurLabadia2026VJEPA2U} shows is important for dense prediction tasks. Each tokenizer uses finite scalar quantization (FSQ)~\cite{mentzer2024finite}.

Modality-specific tokenizers are built using a transformer encoder with RoPE~\cite{su2021roformer} in two dimensions for image-like modalities of RGB, event filter-bank channels, LiDAR range image, and in one dimension for IMU data.
The decoder is a hybrid transformer-convolutional architecture which we found to produce sharper reconstructions, see \cref{fig:convtoken}.
Tokenizers are trained to minimize the mean $\ell_1$ loss with modality-specific modifications. For LiDAR, in addition to $\ell_1$ on log-depth, we use a depth-validity mask with binary-cross-entropy loss and a SiLog loss~\cite{Eigen2014DepthMP}. The validity mask accounts for ray drops which produce holes in the range image~\cite{Manivasagam2020LiDARsimRL}.
For event cameras where the input is sparse, we minimize a weighted reconstruction loss on activations of the filter bank that balances active (event-containing) and inactive pixels.
For the IMU, we regress upon the time series after removing the mean and applying a fixed normalization to the dataset.
\cref{app:hparams} gives details.

\subsection{Fusing individual modalities in the MAE}
\label{subsec:arch}

\begin{wrapfigure}{R}{0.6\linewidth}
    \centering
    \includegraphics[width=\linewidth]{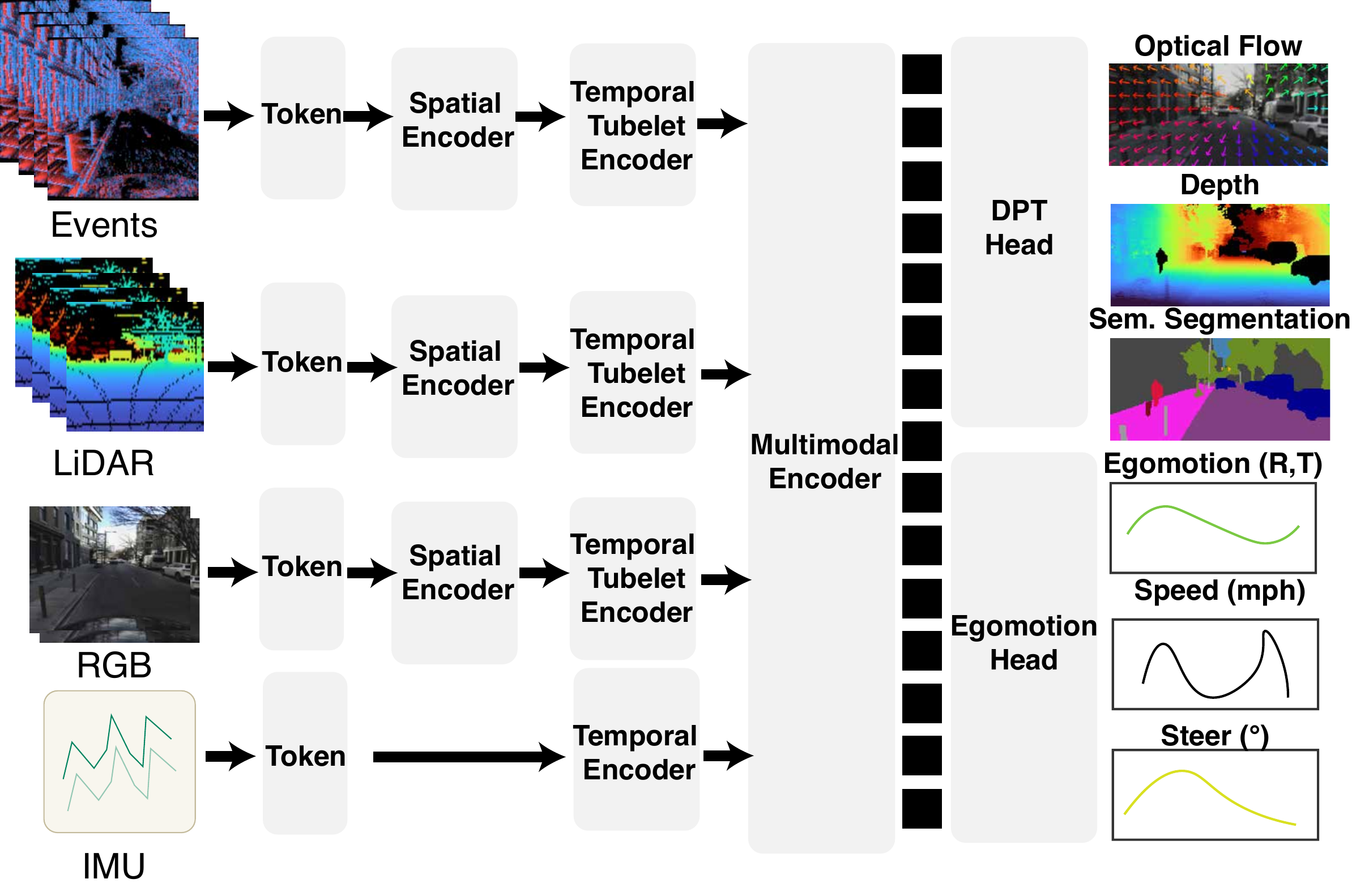}
    \caption{A schematic of the late-fusion MAE encoder and probe architecture. The text provides more details. \cref{app:hparams} elaborates upon the architecture with detailed schematics in  \cref{fig:method}.
}
    \label{fig:mini_arch}
\end{wrapfigure}

The multi-modal MAE~(\cref{fig:mini_arch}) takes as input tokens from each sensor over a 1.4 sec window, with the exception of the IMU, which is tokenized over a slightly wider 1.6 sec window. Each datum consists of 8 uniformly spaced LiDAR scans within this window, 8 RGB images from the left camera with timestamps closest to these scans, 8 left event camera filter bank responses at these timestamps, along with accelerometer and gyroscope time series.\footnote{Experiments in this paper do not use right RGB/event cameras, or the thermal camera.}

\textbf{Early vs. late-fusion.}
Building a representation of this data requires reasoning over spatial structure within each modality, temporal structure across the time window, and cross-modal structure across modalities.
It is possible to take the input representations in \cref{subsec:input_representation} and process them ``directly'' using a transformer architecture---we call this an early fusion architecture. The upside of this approach is that it can exploit synergistic or complementary information across sensors early. The downside is that it puts severe demands on GPU memory and bandwidth.
We therefore propose a late-fusion architecture with modality-specific spatial and temporal tubelet encoders that feed into a cross-modal encoder to create the representation.

\textbf{Masking strategy.}
First consider a sensor $s \in \{1, 2, 3\}$ which could be RGB/event camera or the LiDAR. The masking ratio is $r_s = 1-\min(1, \f{3}{5}\; p_s)$ where $p \sim \text{Dirichlet}(\g, \g, \g)$ with $\g$ drawn uniformly randomly from $\{0.5, 1, 2\}$.%
\footnote{A draw from a Dirichlet distribution with $n$ parameters is a point in the $(n-1)$-dimensional simplex. The distribution is uniform for $\g = 1$. For $\g > 1$ the distribution is concentrated near the centroid, i.e., on all modalities. Distributions with $\g < 1$ concentrate on the boundaries, i.e., drop one or more modalities.}
Masks are spatio-temporal tubes that span the entire time window for each modality.
For one-dimensional IMU data, tokens are kept (left unmasked) independently with Bernoulli parameter $r_s=0.2 + \f{1}{2} r_0$ where $r_0 \sim \text{Beta}(2, 5)$.

\textbf{Architecture.}
For the fusion MAE, we use a four-dimensional variant of RoPE corresponding to coordinates $(t, u_1, u_2, s)$ where $t \in \{0, 1, \dots, 7\}$ indicates position within the temporal window, $(u_1, u_2)$ indicates patch coordinate for image-like inputs (set to zero for IMU) and $s$ indicates the sensor index.
The encoder in our late-fusion MAE applies a per-modality spatial encoder at each time instant, followed by a per-modality temporal encoder across each unmasked tube (let us call this ``tubelet attention'').%
\footnote{In contrast to tubelet attention, ``full attention'' attends to all spatio-temporal tokens within a modality.}
Both of these are self-attention based transformer architectures.
A cross-modal encoder processes features from all these modalities further to create the representation.
The decoder receives the encoder output together with a learned mask token inserted at every masked patch.
See \cref{fig:mini_arch} for rough schematic of the architecture.
The multi-modal MAE is trained to regress the (frozen) FSQ tokenizer's values using an $\ell_1$ loss.
\cref{app:hparams} gives more details of the architecture and \cref{app:masked_recon} discusses some examples of reconstruction of original sensory data.

\subsection{Task-specific probes}
\label{subsec:dpt}

After pretraining the encoder, we can use the representation to train task-specific probes.
For dense prediction tasks such as segmentation, optical flow and depth estimation, we train a dense prediction transformer (DPT)-style ~\cite{Ranftl2021VisionTF} fusion head using features from four encoder layers uniformly separated in depth. Before feeding into the DPT, for each of these layers, we (i) process features using a self-attention block to benefit from temporal information within the time window of each modality, and (ii) select a fixed number of features (as many as the number of patches in one RGB frame) as input to the DPT.%
\footnote{
There are two things to note here. First, when we run baseline methods, say DINO, we encode each of the 8 RGB images within the time window and then run the self-attention block in step (i); this helps in tasks such as optical flow which require temporal context. Second, we select a fixed number of tokens in step (ii) to ensure that the DPT receives a fixed-size grid as input.}
The same DPT network is used to predict multiple tasks using task-specific heads, trained with SiLog loss~\cite{Eigen2014DepthMP} for depth, Charbonnier loss~\cite{Charbonnier} for flow, and cross-entropy loss for segmentation.
\cref{fig:method} provides a schematic of the DPT and \cref{app:eval} provides more details.
For ego-motion prediction, we train three independent cross-attention modules whose queries attend to the last layer of the encoder. These probes predict linear and angular velocity, relative translation and rotation across time instants within the window, car speed and steering position. For the rotation heads, we minimize the squared Frobenius norm between the
predicted and ground-truth rotation matrices; at inference, the predicted matrix
is projected onto $\mathrm{SO}(3)$~\cite{Geist2024LearningW3,Levinson2020AnAO}. The remaining
heads are trained with a Huber loss on Z-scored targets.

\section{Results}
\label{sec:results}

This section provides experimental evidence and analysis to support our claim that multimodal sensors provide complementary information that vision-only foundation models cannot recover. We are particularly interested in understanding performance when sensors are degraded or missing.


\begin{table}
\centering
\Large
\caption{
\textbf{Late-fusion MAE architecture predicts multiple tasks robustly across varied conditions.}
We report root mean square error of depth prediction at pixels with depth within [3, 200] m.
For flow, we report the average end-point error in pixels.
For segmentation, we report mean intersection-over-union (mIoU) over 19 Cityscapes classes.
Ground-truth segmentation labels are unavailable at night.
For ego-motion we report RMSE on translation~(m), rotation angle~(°) and steering position~(°).
All metrics are reported on data where speed is at least 5 mph.
These metrics are a very heavily summarized evaluation of the performance, e.g., RMSE depth could be measured in terms of SiLog loss, mean absolute error, errors on objects at different distances from the platform etc.
We report such detailed metrics in \cref{app:error_analysis}.
Our approach also performs well on these detailed metrics.
For V-JEPA 2.1 (stereo) $^*$, we ran the original model on both the left and right RGB images.
}
\vspace*{1ex}
\label{tab:v9-main}
\begin{adjustbox}{width=\linewidth}
\renewcommand{\arraystretch}{1.2}
\begin{tabular}{l rr rr rr rr rr rr}
\toprule
\textbf{Encoder}  & \multicolumn{2}{c}{\textbf{Depth (m) $\downarrow$}} & \multicolumn{2}{c}{\textbf{Flow (px) $\downarrow$}} & \multicolumn{2}{c}{\textbf{Seg. (mIoU $\uparrow$)}} & \multicolumn{2}{c}{\textbf{Trans. (m) $\downarrow$}} & \multicolumn{2}{c}{\textbf{Rotation ($^\circ$) $\downarrow$}} & \multicolumn{2}{c}{\textbf{Steering ($^\circ$) $\downarrow$}} \\
\cmidrule(lr){2-3} \cmidrule(lr){4-5} \cmidrule(lr){6-7} \cmidrule(lr){8-9} \cmidrule(lr){10-11} \cmidrule(lr){12-13}
& day & night & day & night & & day & day & night & day & night & day & night \\
\midrule
\textbf{Existing approaches}\\
DINOv2 & 6.82 & 9.95 & 19.46 & 18.36 &  & 0.410 & 0.96 & 0.84 & 0.80 & 0.75 & 22.62 & 23.11 \\
DINOv3 & 6.93 & 10.03 & 19.43 & 16.97 &  & 0.382 & 0.93 & 0.79 & 0.79 & 0.76 & 25.24 & 21.93 \\
SigLIP 2 & 7.22 & 10.36 & 20.22 & 18.12 &  & 0.377 & 0.98 & 0.78 & 0.95 & 0.94 & 36.12 & 36.00 \\
Perception Enc. & 7.12 & 10.20 & 19.90 & 17.84 &  & 0.388 & 0.96 & 0.91 & 0.94 & 0.93 & 35.62 & 34.54 \\
V-JEPA 2.1 & 6.38 & 9.51 & 9.13 & 10.90 &  & 0.402 & 0.77 & 0.76 & 0.47 & 0.47 & 5.98 & 6.83 \\
\\
\textbf{Our baselines}\\
V-JEPA 2.1 (stereo)$^*$ & 6.40 & 9.51 & 9.03 & 10.55 &  & 0.405 & 0.77 & 0.71 & 0.48 & 0.47 & 5.69 & 6.47 \\
RGB-only image MAE & 7.78 & 10.58 & 20.88 & 18.98 &  & 0.296 & 0.92 & 1.05 & 0.88 & 0.77 & 31.57 & 21.39 \\
RGB-only video MAE & 7.41 & 10.20 & 10.11 & 11.56 &  & 0.327 & 0.76 & 0.72 & 0.50 & 0.50 & 12.96 & 12.45 \\
Early-fusion MAE & 4.98 & 6.30 & 2.01 & 2.45 &  & 0.379 & 0.14 & 0.15 & 0.56 & 0.53 & 4.58 & 4.48 \\
\\
\textbf{Late-fusion MAE} & \textbf{4.73} & 6.08 & 1.97 & 2.39 &  & \textbf{0.411} & 0.06 & 0.06 & 0.24 & 0.23 & 2.51 & 2.27 \\
w/ full attention & \textbf{4.73} & \textbf{6.07} & \textbf{1.96} & \textbf{2.38} &  & 0.405 & \textbf{0.05} & \textbf{0.05} & \textbf{0.17} & \textbf{0.17} & \textbf{2.46} & \textbf{2.25} \\
\\
\multicolumn{6}{l}{\textbf{Leave one modality out/in at test time in late-fusion MAE}}\\
w/o RGB & 4.70 & 5.99 & 2.02 & 2.43 &  & 0.366 & 0.06 & 0.06 & 0.24 & 0.23 & 2.65 & 2.34 \\
w/o EV & 4.88 & 6.32 & 2.00 & 2.42 &  & 0.403 & 0.05 & 0.05 & 0.21 & 0.20 & 2.59 & 2.42 \\
w/o LiDAR & 6.74 & 9.55 & 7.19 & 8.05 &  & 0.351 & 0.73 & 0.57 & 0.46 & 0.45 & 3.97 & 4.00 \\
w/o IMU & 4.67 & 6.00 & 2.05 & 2.49 &  & 0.402 & 0.06 & 0.06 & 0.24 & 0.22 & 2.78 & 2.61 \\
w/ only RGB & 7.20 & 10.09 & 11.31 & 12.81 &  & 0.310 & 0.83 & 0.77 & 0.47 & 0.47 & 8.29 & 7.75 \\
\bottomrule
\end{tabular}
\end{adjustbox}
\end{table}

\textbf{Baselines.}
We compare our approach to existing vision-only methods (DINOv2, DINOv3, SigLIP 2, Perception Encoder and V-JEPA 2.1) across five tasks (depth, flow, segmentation, egomotion and steering position).
We also develop four other baselines that are based on our approach:
(i) RGB-only MAE (with only the left camera) trained on FSQ targets for single frames which only has access to spatial information,
(ii) RGB-only video MAE (with only the left camera) trained on FSQ targets for all 8 frames within the temporal window and thereby also receives some temporal information, and
(iii) early fusion MAE as described in \cref{subsec:arch} which uses a monolithic transformer architecture and has access to multi-modal information,
(iv) late-fusion MAE (w/ full attention) that performs temporal self-attention over all tokens in a modality, as opposed to our proposed architecture which only performs self-attention over spatiotemporal tubes independently.
To evaluate vision-only baselines, we encode each of the 8 RGB images within the time window and then proceed to the self-attention-block implemented before the DPT, as discussed in \cref{subsec:dpt}. \cref{app:eval} provides more details.

\textbf{Training and evaluation methodology.}
We split OctoSense data into 293 training and 78 test sequences, each about 10 mins long. Test set contains data from both daytime and nighttime conditions, with sensor-degraded sequences reserved as a separate held-out set in \cref{app:error_analysis}. During training we split data into 1.4 sec windows that have an overlap of 1 sec with each other; temporal windows for test data do not have any overlap.
We train our late-fusion MAE
(ViT-B scale) for 150 epochs with a batch size of 512 on four NVIDIA B200 GPUs; hyper-parameters are provided in \cref{app:hparams}. The experiments in this paper required $\sim$7,000 GPU-hours. All task probes are trained for 60 epochs. Secondary metrics are reported in \cref{app:error_analysis}.

\textbf{Overall results.}
\cref{tab:v9-main} summarizes the performance of our proposed approach as well as some of the baselines.
Overall, on all tasks, the late-fusion MAE performs much better than other approaches.
Depth RMSE is better by 1.65 m compared to the best existing method (V-JEPA 2.1). This difference is more than twice as large in nighttime testing, which demonstrates the utility of having multiple modalities as well as the robustness of our approach.
Trends are very similar for flow estimation, more than 7.16 and 8.51 pixel improvement over V-JEPA 2.1 in daytime and nighttime respectively. On segmentation, the late-fusion MAE marginally exceeds the best existing model (0.411 vs. 0.410). The narrow margin is likely because our pseudo-labels in OctoSense are generated by a DINOv2-based model. Our early-fusion MAE matches DINOv2 on frequent categories, but the former does poorly on rare categories (poles, buses, people), see \cref{tab:seg_per_class}. Predicting ego-motion from RGB-only models is quite difficult, and our approach performs significantly better because it has access to LiDAR as well as inertial information. We achieve a prediction error of 0.06 m in translation and 0.23--0.24° in rotation. The relative advantage of having multiple sensory modalities is higher at nighttime when visual sensors are degraded (\cref{tab:main:degraded}).

\begin{wraptable}{R}{0.65\linewidth}
\centering
\caption{Performance of different encoders on degraded data from OctoSense. We report RMSE (m) for depth, end-point error (px) for flow, translation (m) and rotation (°) for ego-motion estimation.}
\label{tab:main:degraded}
\begin{adjustbox}{width=\linewidth}
\renewcommand{\arraystretch}{1.2}
\begin{tabular}{lcccc}
\toprule
\textbf{Encoder} & \textbf{Depth (m) $\downarrow$} & \textbf{Flow (px) $\downarrow$} & \textbf{Trans.\@(m) $\downarrow$} & \textbf{Rot.\@ (°) $\downarrow$}\\
\midrule
DINOv2          & 7.41 & 19.97 & 0.65 & 0.58 \\
  DINOv3          & 7.49 & 18.82 & 0.65 & 0.61 \\
  SigLIP 2        & 7.55 & 19.02 & 0.71 & 0.70 \\
  Perception Enc. & 7.56 & 19.52 & 0.72 & 0.71 \\
  V-JEPA 2.1      & 7.40 & 10.85 & 0.51 & 0.43 \\
  Late-fusion MAE & \textbf{4.77} & \textbf{2.12} & \textbf{0.06} & \textbf{0.23} \\
\bottomrule
\end{tabular}
\end{adjustbox}
\vspace*{1ex}

\caption{Latency of the forward pass for late-fusion MAE with similar number of parameters as that of the early-fusion MAE (batch-size of 1). \cref{tab:hmae_latency,tab:hmae_latency_jetson} elaborate upon this.}
\label{tab:main:timing}
\begin{adjustbox}{width=\linewidth}
\renewcommand{\arraystretch}{1.2}
\begin{tabular}{lrrr}
\toprule
\textbf{Fusion strategy} & \textbf{Nvidia GPU} & \textbf{\# weights} (M) & \textbf{Latency (ms)}\\
\midrule
Early & 5090 & 85 & 11.35\\
Late & 5090 & 83.7 & 6.68\\
Early & Orin NX & 85   & 288\\
Late & Orin NX & 83.7 & 112\\
\bottomrule\\
\end{tabular}
\end{adjustbox}

\caption{Late-fusion MAE trained on OctoSense evaluated on M3ED~\cite{m3ed}.}
\label{tab:main:m3ed}
\begin{adjustbox}{width=\linewidth}
\renewcommand{\arraystretch}{1.2}
\begin{tabular}{lcccc}
\toprule
\textbf{Encoder} & \textbf{Depth (m) $\downarrow$} & \textbf{Flow (px) $\downarrow$} & \textbf{Trans.\@(m) $\downarrow$} & \textbf{Rot.\@ (°) $\downarrow$}\\
\midrule
DINOv2          & 9.53 & 5.52 & 0.46 & 1.13 \\
DINOv3          & 9.29 & 5.09 & 0.55 & 1.24 \\
SigLIP 2        & 9.69 & 5.43 & 0.70 & 1.24 \\
Perception Enc. & 9.75 & 5.26 & 0.53 & 1.32 \\
V-JEPA 2.1      & 9.72 & \textbf{1.59} & \textbf{0.22} & \textbf{0.64} \\
Late-fusion MAE & \textbf{9.05} & 1.60 & 0.36 & 0.82 \\
\bottomrule
\end{tabular}
\end{adjustbox}
\end{wraptable}

\textbf{Comparison to baselines.}
RGB-only image MAE performs on par with existing methods, although the former is trained on our data while the latter are being evaluated zero-shot on our data. This is a positive datapoint to suggest that one can train useful self-supervised learning methods using data at the OctoSense scale. The video MAE's gain over the image MAE (${\sim}11$ px on daytime flow) matches V-JEPA 2.1's gain over the image-based existing methods, both reflect pretraining with temporal information. Late-fusion MAE performs about as well as the early-fusion MAE. This is somewhat remarkable because it indicates that a monolithic transformer can adequately represent information across these modalities. As \cref{tab:main:timing} shows, the late-fusion MAE has about 60\% (Orin) smaller latency and is much more amenable to implementation on NVIDIA Orin.
Late-fusion MAE (w/ full attention) which uses full temporal self-attention performs on-par with our proposed tubelet attention.

\textbf{Leaving out modalities at test time.}
Leaving out a specific modality at test time in~\cref{tab:v9-main} helps understand the relative contribution from that modality.
LiDAR is a dominant sensor for ego-motion estimation tasks, without it the error in translation, rotation and steering angle is very high compared to late-fusion MAE. Dropping event images or the IMU has a smaller effect on the performance compared to dropping LiDAR. LiDAR is also the most important sensor for dense prediction: dropping it sharply degrades both depth and flow (RMSE $4.73 \rightarrow 6.74$~m, EPE $1.97 \rightarrow 7.19$~px), whereas dropping RGB leaves them essentially unchanged. The RGB camera instead matters most for segmentation, where leaving it out is what hurts performance the most. Event cameras and IMU contribute marginally on most tasks. It is remarkable that well-trained foundation models like V-JEPA 2.1 that are trained only on RGB data perform comparably to our late-fusion MAE that does not have access to LiDAR data (w/o LiDAR row in \cref{tab:v9-main}). In view of the fact that OctoSense is a much smaller dataset than those used to train industry-scale foundation models, this suggests that having LiDAR information can go a long way. 

Indeed, LiDAR provides a direct geometric signal for ego-motion while RGB images serve to spatially anchor the tokens of other modalities for pixel-aligned semantic prediction. The fact that we do not see event cameras and IMU contributing to the performance much is perhaps due to the fact that we evaluate these tasks at 5 Hz. This is far below the native temporal resolution of these sensors~(microsecond-scale for event cameras and $\sim$400 Hz for IMU). These sensors will be useful for high-speed tracking and low-latency maneuvers; we leave this for future work.

\textbf{Zero-shot generalization to new datasets.}
Foundation models like DINO or V-JEPA generalize well to new settings, in part because they are trained on large datasets. To check whether OctoSense data can lead to such effective generalization, we evaluate on the M3ED dataset~\cite{m3ed}. M3ED is the only other large dataset with event camera data, a similar sensor suite and ego-motion ground truth.%
\footnote{Cross-dataset generalization requires a dataset with a sensor suite close to ours and reliable ground-truth odometry. To our knowledge, M3ED is the only publicly available dataset that meets both criteria. Evaluating generalization across substantially different sensor configurations e.g., 16/32/64/128-beam LiDARs, is itself an open research problem that we leave to future work. } As \cref{tab:main:m3ed} shows, our late-fusion architecture is competitive on dense perception tasks compared to existing foundation models, and second best on ego-motion tasks. Our model performs quite well on a new sensor configuration (different intrinsics/extrinsics and RGB/event-camera sensor).

\section{Limitations and Directions for Future Work}

Our paper demonstrates the utility and viability of multi-modal SSL across a variety of robot perception tasks and experimental settings. But it is a research prototype. Our dataset is orders-of-magnitude smaller than the datasets used to train foundation models such as DINO and V-JEPA. Training on more data can improve the generalizability of our learned representation---and in effect allow us to build a ``foundation'' multi-modal robot perception model. For our approach to replace existing foundation models, we will need a community-wide effort to pool data.

Our computing budget restricted us from training with larger networks, which can improve performance. One of the key avenues for future work is to use temporal windows longer than 1.4 sec with recurrent architectures. This can enable new tasks such as object tracking and 3D instance detection (after appropriate pseudo annotation), and also further reduce the computational complexity of our approach. We are also interested in questions regarding the unique, synergistic and redundant information in multi-modal sensors in the context of prior work which has noticed that many perception tasks are highly redundant functions of their inputs \cite{ramesh2024many}. This can shed new light on normative principles for multi-modal robot perception.

\acknowledgments{
This work was supported by grants from the National Science Foundation (IIS-2145164, CCF-2212519), the NSF and DoD OUSD (R\&E) under Agreement PHY-2229929 (The NSF AI Institute for Artificial and Natural Intelligence), DSO National Laboratories, Singapore and the Office of Naval Research DURIP.
}

\renewcommand{\bibfont}{\small}
\setlength{\bibsep}{2ex}
\bibliography{ref}

@string{cvpr  = {IEEE/CVF Conference on Computer Vision and Pattern Recognition}}

@string{iccv  = {IEEE/CVF International Conference on Computer Vision}}

@string{eccv  = {European Conference on Computer Vision}}

@string{wacv  = {IEEE/CVF Winter Conference on Applications of Computer Vision}}

@string{nips  = {Advances in Neural Information Processing Systems}}

@string{icml  = {International Conference on Machine Learning}}

@string{iclr  = {International Conference on Learning Representations}}

@string{icra  = {IEEE International Conference on Robotics and Automation}}

@string{iros  = {IEEE/RSJ International Conference on Intelligent Robots and Systems}}

@string{itsc  = {IEEE International Conference on Intelligent Transportation Systems}}

@string{pami  = {IEEE Transactions on Pattern Analysis and Machine Intelligence}}

@string{ral   = {IEEE Robotics and Automation Letters}}

@article{das2025fast,
  title={Fast feature field ({F}3): A predictive representation of events},
  author={Das, Richeek and Daniilidis, Kostas and Chaudhari, Pratik},
  journal={arXiv preprint 2509.25146},
  year={2025}
}

@inproceedings{jaegle2021perceiver,
  title={Perceiver: General perception with iterative attention},
  author={Jaegle, Andrew and Gimeno, Felix and Brock, Andy and Vinyals, Oriol and Zisserman, Andrew and Carreira, Joao},
  booktitle=ICML,
  year={2021}
}

@article{Klenk2022MaskedEM,
  title={{Masked Event Modeling}: Self-Supervised Pretraining for Event Cameras},
  author={Simone Klenk and David Bonello and Lukas Koestler and Daniel Cremers},
  journal=WACV,
  year={2022},
}

@inproceedings{Sun2023CALICOSC,
  title={{CALICO}: Self-Supervised Camera-LiDAR Contrastive Pre-training for BEV Perception},
  author={Jiachen Sun and Haizhong Zheng and Qingzhao Zhang and Atul Prakash and Zhuoqing Morley Mao and Chaowei Xiao},
  booktitle=ICLR,
  year={2024},
}

@inproceedings{Lu2023UnifiedIO2S,
  title={{Unified-IO 2}: Scaling Autoregressive Multimodal Models with Vision, Language, Audio, and Action},
  author={Jiasen Lu and Christopher Clark and Sangho Lee and Zichen Zhang and Savya Khosla and Ryan Marten and Derek Hoiem and Aniruddha Kembhavi},
  booktitle=CVPR,
  year={2024},
}

@inproceedings{Chen2020ASF,
  title={A Simple Framework for Contrastive Learning of Visual Representations},
  author={Ting Chen and Simon Kornblith and Mohammad Norouzi and Geoffrey E. Hinton},
  booktitle=ICML,
  year={2020},
}

@inproceedings{Hamilton2008ACESAC,
  title={{ACES}: adaptive clock estimation and synchronization using Kalman filtering},
  author={Benjamin R. Hamilton and Xiaoli Ma and Qi Zhao and Jun Xu},
  booktitle={ACM/IEEE International Conference on Mobile Computing and Networking},
  year={2008},
}

@article{Xu2025WODE2EWO,
  title={{WOD-E2E}: Waymo Open Dataset for End-to-End Driving in Challenging Long-tail Scenarios},
  author={Runsheng Xu and Hubert Lin and Wonseok Jeon and Hao Feng and Yuliang Zou and Liting Sun and John Gorman and Kate Tolstaya and Sarah Tang and Brandyn White and Benjamin Sapp and Mingxing Tan and Jyh-Jing Hwang and Drago Anguelov},
  journal={arXiv preprint 2510.26125},
  year={2025},
}

@techreport{ITUT_H265,
  title = {Series H: Audiovisual and Multimedia Systems: Infrastructure of audiovisual services - Coding of moving video: High efficiency video coding},
  author = {{ITU-T}},
  institution = {International Telecommunication Union},
  number = {ITU-T H.265},
  year = {2026},
  note = {Version 01/2026},
}

@article{Lagorce2017HOTSAH,
  title={{HOTS}: A Hierarchy of Event-Based Time-Surfaces for Pattern Recognition},
  author={Xavier Lagorce and G. Orchard and Francesco Galluppi and Bertram E. Shi and Ryad B. Benosman},
  journal={IEEE Transactions on Pattern Analysis and Machine Intelligence},
  year={2017},
  volume={39},
  pages={1346-1359},
}

@book{gerstner2014neuronal,
  title     = {Neuronal Dynamics: From Single Neurons to Networks and Models of Cognition},
  author    = {Gerstner, Wulfram and Kistler, Werner M. and Naud, Richard and Paninski, Liam},
  year      = {2014},
  publisher = {Cambridge University Press},
}

@inproceedings{Devlin2019BERTPO,
  title={{BERT}: Pre-training of Deep Bidirectional Transformers for Language Understanding},
  author={Jacob Devlin and Ming-Wei Chang and Kenton Lee and Kristina Toutanova},
  booktitle={North American Chapter of the Association for Computational Linguistics},
  year={2019},
}

@techreport{christensen2026global,
  author      = {Christensen, Henrik I.},
  title       = {Global Robotics Technology Roadmap 2025--2035: A Multi-Regional, Cross-Domain Strategic Perspective for Europe, Asia, and the United States},
  institution = {University of California San Diego},
  year        = {2026},
  month       = {April},
  note        = {Version 1.02},
  type        = {Technology Roadmap}
}

@article{Munir2021SSTNSD,
  title={SSTN: Self-Supervised Domain Adaptation Thermal Object Detection for Autonomous Driving},
  author={Farzeen Munir and Shoaib Azam and Moongu Jeon},
  journal=IROS,
  year={2021},
}

@InProceedings{Ryali2023HieraAH,
  title = 	 {Hiera: A Hierarchical Vision Transformer without the Bells-and-Whistles},
  author =       {Ryali, Chaitanya and Hu, Yuan-Ting and Bolya, Daniel and Wei, Chen and Fan, Haoqi and Huang, Po-Yao and Aggarwal, Vaibhav and Chowdhury, Arkabandhu and Poursaeed, Omid and Hoffman, Judy and Malik, Jitendra and Li, Yanghao and Feichtenhofer, Christoph},
  booktitle = 	 ICML,
  year = 	 {2023},
}

@inproceedings{Bolya2025PerceptionET,
  title={Perception Encoder: The best visual embeddings are not at the output of the network},
  author={Daniel Bolya and Po-Yao Huang and Peize Sun and Jang Hyun Cho and Andrea Madotto and Chen Wei and Tengyu Ma and Jiale Zhi and Jathushan Rajasegaran and Hanoona Abdul Rasheed and Junke Wang and Marco Monteiro and Hu Xu and Shiyu Dong and Nikhila Ravi and Shang-Wen Li and Piotr Doll'ar and Christoph Feichtenhofer},
  booktitle=NIPS,
  year={2025},
}

@inproceedings{Radford2021LearningTV,
  title={Learning Transferable Visual Models From Natural Language Supervision},
  author={Alec Radford and Jong Wook Kim and Chris Hallacy and Aditya Ramesh and Gabriel Goh and Sandhini Agarwal and Girish Sastry and Amanda Askell and Pamela Mishkin and Jack Clark and Gretchen Krueger and Ilya Sutskever},
  booktitle=ICML,
  year={2021},
}

@article{Tschannen2025SigLIP2M,
  title={{SigLIP 2}: Multilingual Vision-Language Encoders with Improved Semantic Understanding, Localization, and Dense Features},
  author={Michael Tschannen and Alexey Gritsenko and Xiao Wang and Muhammad Ferjad Naeem and Ibrahim M. Alabdulmohsin and Nikhil Parthasarathy and Talfan Evans and Lucas Beyer and Ye Xia and Basil Mustafa and Olivier H'enaff and Jeremiah Harmsen and Andreas Steiner and Xiao-Qi Zhai},
  journal={arXiv preprint 2502.14786},
  year={2025},
}

@article{Oord2018RepresentationLW,
  title={Representation Learning with Contrastive Predictive Coding},
  author={A{\"a}ron van den Oord and Yazhe Li and Oriol Vinyals},
  journal={arXiv preprint 1807.03748},
  year={2018},
}

@inproceedings{caron2021emerging,
  title={Emerging Properties in Self-Supervised Vision Transformers},
  author={Caron, Mathilde and Touvron, Hugo and Misra, Ishan and J\'egou, Herv\'e  and Mairal, Julien and Bojanowski, Piotr and Joulin, Armand},
  booktitle=ICCV,
  year={2021}
}

@inproceedings{Zou2023UniM2AEMM,
  title={{UniM2AE}: Multi-modal Masked Autoencoders with Unified 3D Representation for 3D Perception in Autonomous Driving},
  author={Jian Zou and Tianyu Huang and Guanglei Yang and Zhenhua Guo and Wangmeng Zuo},
  booktitle=ECCV,
  year={2024},
}

@article{Zong2023SelfSupervisedML,
  title={Self-Supervised Multimodal Learning: A Survey},
  author={Yongshuo Zong and Oisin Mac Aodha and Timothy M. Hospedales},
  journal={IEEE Transactions on Pattern Analysis and Machine Intelligence},
  year={2023},
  volume={47},
  pages={5299-5318},
}

@inproceedings{Yu2021PointBERTP3,
  title={{Point-BERT}: Pre-training 3D Point Cloud Transformers with Masked Point Modeling},
  author={Xumin Yu and Lulu Tang and Yongming Rao and Tiejun Huang and Jie Zhou and Jiwen Lu},
  booktitle=CVPR,
  year={2022},
}

@inproceedings{Xie2020PointContrastUP,
  title={{PointContrast}: Unsupervised Pre-training for 3D Point Cloud Understanding},
  author={Saining Xie and Jiatao Gu and Demi Guo and C. Qi and Leonidas J. Guibas and Or Litany},
  booktitle=ECCV,
  year={2020},
}

@inproceedings{Xu2021LIMUBERTUT,
  title={{LIMU-BERT}: Unleashing the Potential of Unlabeled Data for IMU Sensing Applications},
  author={Huatao Xu and Pengfei Zhou and Rui Tan and Mo Li and Guobin Shen},
  booktitle={ ACM Conference on Embedded Networked Sensor Systems},
  year={2021},
}

@inproceedings{Wang2020UnsupervisedPC,
  title={Unsupervised Point Cloud Pre-training via Occlusion Completion},
  author={Hanchen Wang and Qi Liu and Xiangyu Yue and Joan Lasenby and Matt J. Kusner},
  booktitle=ICCV,
  year={2021},
}

@inproceedings{Eigen2014DepthMP,
  title={Depth Map Prediction from a Single Image using a Multi-Scale Deep Network},
  author={David Eigen and Christian Puhrsch and Rob Fergus},
  booktitle=NIPS,
  year={2014},
}

@inproceedings{Yang2023EventCD,
  title={Event Camera Data Dense Pre-training},
  author={Yan Yang and Liyuan Pan and Liu Liu},
  booktitle=ECCV,
  year={2024},
}

@ARTICLE{async_delta,
  author={Hawkes, T. and Simonpieri, P.},
  journal={IEEE Trans. on Comm.}, 
  title={Signal Coding Using Asynchronous Delta Modulation}, 
  year={1974},
  volume={22},
  number={5},
  month={March},
  pages={729-731}}

@ARTICLE{MVSEC,
  author={Zhu, Alex Zihao and Thakur, Dinesh and Özaslan, Tolga and Pfrommer, Bernd and Kumar, Vijay and Daniilidis, Kostas},
  journal={IEEE Robt. and Autom. Lett.}, 
  title={The Multivehicle Stereo Event Camera Dataset: An Event Camera Dataset for 3D Perception},
  month={Feb.},
  year={2018},
  volume={3},
  pages={2032-2039}}

@inproceedings{adamw,
  title={Decoupled Weight Decay Regularization},
  author={Ilya Loshchilov and Frank Hutter},
  booktitle={International Conference on Learning Representations},
  year={2017}
}

@ARTICLE{ev_survey,
  author={Gallego, Guillermo and Delbrück, Tobi and Orchard, Garrick and Bartolozzi, Chiara and Taba, Brian and Censi, Andrea and Leutenegger, Stefan and Davison, Andrew J. and Conradt, Jörg and Daniilidis, Kostas and Scaramuzza, Davide},
  journal={IEEE Transactions on Pattern Analysis and Machine Intelligence}, 
  title={Event-Based Vision: A Survey}, 
  year={2022},
  volume={44},
  number={1},
  pages={154-180}}

@article{malladi2025arxiv,
  author      = {M.V.R. Malladi and T. Guadagnino and L. Lobefaro and C. Stachniss},
  title       = {A Robust Approach for LiDAR-Inertial Odometry Without Sensor-Specific Modeling},
  journal     = RAL,
  year        = {2026},
  volume      = {11},
  number      = {6},
  pages       = {7420--7427},
}

@inproceedings{kerssies2025eomt,
  author    = {Kerssies, Tommie and Cavagnero, Niccol\`{o} and Hermans, Alexander and Norouzi, Narges and Averta, Giuseppe and Leibe, Bastian and Dubbelman, Gijs and {de Geus}, Daan},
  title     = {{Your ViT is Secretly an Image Segmentation Model}},
  booktitle = CVPR,
  year      = {2025},
}

@inproceedings{rehder2016extending,
  title={Extending kalibr: Calibrating the extrinsics of multiple {IMUs} and of individual axes},
  author={Rehder, Joern and Nikolic, Janosch and Schneider, Thomas and Hinzmann, Timo and Siegwart, Roland},
  booktitle=ICRA,
}

@inproceedings{furgale2013unified,
  title={Unified temporal and spatial calibration for multi-sensor systems},
  author={Furgale, Paul and Rehder, Joern and Siegwart, Roland},
  booktitle={IEEE/RSJ International Conference on Intelligent Robots and Systems},
  pages={1280--1286},
  year={2013},
}

@phdthesis{zurn2024selfsupervised,
  author       = {Z{\"u}rn, Jannik},
  title        = {Self-supervised and Multi-modal Learning for Perception in Mobile Robots and Autonomous Driving},
  school       = {University of Freiburg},
  year         = {2024},
  doi          = {10.6094/UNIFR/260783},
}

@INPROCEEDINGS{Bisulco2020FastMU,
  author={ Bisulco, Anthony and Ojeda, Fernando Cladera and Isler, Volkan and Lee, Daniel D.},
  booktitle=ICRA, 
  title={{ Fast Motion Understanding with Spatiotemporal Neural Networks and Dynamic Vision Sensors}}, 
  year={2021},
}

@Article{dsec,
  author  = {Mathias Gehrig and Willem Aarents and Daniel Gehrig and Davide Scaramuzza},
  title   = {{DSEC}: A Stereo Event Camera Dataset for Driving Scenarios},
  journal = {IEEE Robot. and Autom. Lett.},
  year    = {2021},
  month={March}
}

@misc{nvidia2025physicalai,
  author       = {{NVIDIA Corporation}},
  title        = {{PhysicalAI-Autonomous-Vehicles} Dataset},
  year         = {2025},
  howpublished = {\url{https://huggingface.co/datasets/nvidia/PhysicalAI-Autonomous-Vehicles}},
}

@article{Geiger2012AreWR,
  title={Are we ready for autonomous driving? The KITTI vision benchmark suite},
  author={Andreas Geiger and Philip Lenz and Raquel Urtasun},
  journal={2012 IEEE Conference on Computer Vision and Pattern Recognition},
  year={2012},
  pages={3354-3361},
}

@article{Sivaprakasam2024TartanDrive2M,
  title={{TartanDrive 2.0}: More Modalities and Better Infrastructure to Further Self-Supervised Learning Research in Off-Road Driving Tasks},
  author={Matthew Sivaprakasam and Parv Maheshwari and Mateo Guaman Castro and Samuel Triest and Micah Nye and Steven Willits and Andrew Saba and Wenshan Wang and Sebastian A. Scherer},
  journal={2024 IEEE International Conference on Robotics and Automation (ICRA)},
  year={2024},
  pages={12606-12606},
}

@article{Schafer2018ACI,
  title={A Commute in Data: The comma2k19 Dataset},
  author={Harald Schafer and Eder Santana and Andy Haden and Riccardo Biasini},
  journal={ArXiv},
  year={2018},
  volume={abs/1812.05752},
}

@article{Caesar2019nuScenesAM,
  title={{nuScenes}: A Multimodal Dataset for Autonomous Driving},
  author={Holger Caesar and Varun Bankiti and Alex H. Lang and Sourabh Vora and Venice Erin Liong and Qiang Xu and Anush Krishnan and Yuxin Pan and Giancarlo Baldan and Oscar Beijbom},
  journal={2020 IEEE/CVF Conference on Computer Vision and Pattern Recognition (CVPR)},
  year={2019},
  pages={11618-11628},
}

@article{Sun2019ScalabilityIP,
  title={Scalability in Perception for Autonomous Driving: Waymo Open Dataset},
  author={Pei Sun and Henrik Kretzschmar and Xerxes Dotiwalla and Aurelien Chouard and Vijaysai Patnaik and Paul Tsui and James Guo and Yin Zhou and Yuning Chai and Benjamin Caine and Vijay Vasudevan and Wei Han and Jiquan Ngiam and Hang Zhao and Aleksei Timofeev and Scott M. Ettinger and Maxim Krivokon and Amy Gao and Aditya Joshi and Yu Zhang and Jonathon Shlens and Zhifeng Chen and Dragomir Anguelov},
  journal={2020 IEEE/CVF Conference on Computer Vision and Pattern Recognition (CVPR)},
  year={2019},
  pages={2443-2451},
}

@article{Girdhar2023ImageBindOE,
  title={{ImageBind} One Embedding Space to Bind Them All},
  author={Rohit Girdhar and Alaaeldin El-Nouby and Zhuang Liu and Mannat Singh and Kalyan Vasudev Alwala and Armand Joulin and Ishan Misra},
  journal=cvpr,
  year={2023},
}

@article{Patel2026DeFMLF,
  title={{DeFM}: Learning Foundation Representations from Depth for Robotics},
  author={Manthan Patel and Jonas Frey and Mayank Mittal and Fan Yang and Alexander Hansson and Amir Bar and C{\'e}sar Cadena and Marco Hutter},
  journal={arXiv preprint 2601.18923},
  year={2026},
  volume={abs/2601.18923},
}

@inproceedings{Mizrahi20234MMM,
  title={{4M}: Massively Multimodal Masked Modeling},
  author={David Mizrahi and Roman Bachmann and Ouguzhan Fatih Kar and Teresa Yeo and Mingfei Gao and Afshin Dehghan and Amir Zamir},
  booktitle=NIPS,
  year={2023}
}

@INPROCEEDINGS{m3ed,
  author={Chaney, Kenneth and Cladera, Fernando and Wang, Ziyun and Bisulco, Anthony and Hsieh, M. Ani and Korpela, Christopher and Kumar, Vijay and Taylor, Camillo J. and Daniilidis, Kostas},
  booktitle={IEEE Conf. Comput. Vis. Pattern Recog. Workshop}, 
  title={{M3ED}: Multi-Robot, Multi-Sensor, Multi-Environment Event Dataset}}

@article{LidarArgo,
  title={Argoverse 2: Next Generation Datasets for Self-Driving Perception and Forecasting},
  author={Benjamin Wilson and William Qi and Tanmay Agarwal and John Lambert and Jagjeet Singh and Siddhesh Khandelwal and Bowen Pan and Ratnesh Kumar and Andrew Hartnett and Jhony Kaesemodel Pontes and Deva Ramanan and James Hays},
  journal={ArXiv},
  year={2023},
  volume={abs/2301.00493},
}

@ARTICLE{vector,
  author={ Gao, Ling and Liang, Yuxuan and Yang, Jiaqi and Wu, Shaoxun and Wang, Chenyu and Chen, Jiaben and Kneip, Laurent},
  journal={IEEE Robot. and Autom. Lett.}, 
  title={ {VECtor}: A Versatile Event-Centric Benchmark for Multi-Sensor SLAM}, 
  year={ 2022},
  volume={ 7},
  number={ 3},
  month={June},
  pages={ 8217-8224},}

@article{He2021MAE,
  title={Masked Autoencoders Are Scalable Vision Learners},
  author={Kaiming He and Xinlei Chen and Saining Xie and Yanghao Li and Piotr Doll'ar and Ross B. Girshick},
  journal=CVPR,
  year={2022},
}

@article{Xie2021SimMIMAS,
  title={{SimMIM}: a Simple Framework for Masked Image Modeling},
  author={Zhenda Xie and Zheng Zhang and Yue Cao and Yutong Lin and Jianmin Bao and Zhuliang Yao and Qi Dai and Han Hu},
  journal=CVPR,
  year={2022},
}

@inproceedings{pang2022masked,
  title={Masked autoencoders for point cloud self-supervised learning},
  author={Pang, Yatian and Wang, Wenxiao and Tay, Francis EH and Liu, Wei and Tian, Yonghong and Yuan, Li},
  booktitle=ECCV,
  year={2022},
}

@inproceedings{bachmann2022multimae,
  title={Multimae: Multi-modal multi-task masked autoencoders},
  author={Bachmann, Roman and Mizrahi, David and Atanov, Andrei and Zamir, Amir},
  booktitle=eccv,
  year={2022},
}

@inproceedings{lu2022unified,
  title={{Unified-IO}: A unified model for vision, language, and multi-modal tasks},
  author={Lu, Jiasen and Clark, Christopher and Zellers, Rowan and Mottaghi, Roozbeh and Kembhavi, Aniruddha},
  booktitle=ICLR,
  year={2023}
}

@inproceedings{oquab2023dinov2,
  title={{DINOv2}: Learning Robust Visual Features without Supervision},
  author={Oquab, Maxime and Darcet, Timothée and Moutakanni, Theo and Vo, Huy V. and Szafraniec, Marc and Khalidov, Vasil and Fernandez, Pierre and Haziza, Daniel and Massa, Francisco and El-Nouby, Alaaeldin and Howes, Russell and Huang, Po-Yao and Xu, Hu and Sharma, Vasu and Li, Shang-Wen and Galuba, Wojciech and Rabbat, Mike and Assran, Mido and Ballas, Nicolas and Synnaeve, Gabriel and Misra, Ishan and Jegou, Herve and Mairal, Julien and Labatut, Patrick and Joulin, Armand and Bojanowski, Piotr},
  booktitle=ICLR,
  year={2025}
}

@article{pfrommer2022frequencycam,
      title={Frequency Cam: Imaging Periodic Signals in Real-Time}, 
      author={Bernd Pfrommer},
      year={2022},
      journal={arXiv preprint 2211.00198},
}

@article{ramesh2024many,
  title={Many Perception Tasks are Highly Redundant Functions of their Input Data},
  author={Ramesh, Rahul and Bisulco, Anthony and DiTullio, Ronald W and Wei, Linran and Balasubramanian, Vijay and Daniilidis, Kostas and Chaudhari, Pratik},
  journal={Computational and Systems Neuroscience(COSYNE)},
  year={2025}
}

@article{dosovitskiy2020image,
  title={An image is worth 16x16 words: Transformers for image recognition at scale},
  author={Dosovitskiy, Alexey and Beyer, Lucas and Kolesnikov, Alexander and Weissenborn, Dirk and Zhai, Xiaohua and Unterthiner, Thomas and Dehghani, Mostafa and Minderer, Matthias and Heigold, Georg and Gelly, Sylvain and Uszkoreit, Jakob and Houlsby, Neil},
  journal=ICLR,
  year={2021}
}

@inproceedings{Narayanswamy2024ScalingWF,
  title={Scaling Wearable Foundation Models},
  author={Girish Narayanswamy and Xin Liu and Kumar Ayush and Yuzhe Yang and Xuhai Xu and Shun Liao and Jake Garrison and Shyam Tailor and Jacob Sunshine and Yun Liu and Tim Althoff and S. Narayanan and Pushmeet Kohli and Jiening Zhan and Mark Malhotra and Shwetak N. Patel and Samy Abdel-Ghaffar and Daniel McDuff},
  booktitle=ICLR,
  year={2025},
}

@inproceedings{Wang2020DenseCL,
  title={Dense Contrastive Learning for Self-Supervised Visual Pre-Training},
  author={Xinlong Wang and Rufeng Zhang and Chunhua Shen and Tao Kong and Lei Li},
  booktitle=CVPR,
  year={2021},
}

@inproceedings{Xiong2020OnLN,
  title={On Layer Normalization in the Transformer Architecture},
  author={Ruibin Xiong and Yunchang Yang and Di He and Kai Zheng and Shuxin Zheng and Chen Xing and Huishuai Zhang and Yanyan Lan and Liwei Wang and Tie-Yan Liu},
  booktitle=ICML,
  year={2020},
}

@article{Triest2022TartanDriveAL,
  title={{TartanDrive}: A Large-Scale Dataset for Learning Off-Road Dynamics Models},
  author={Samuel Triest and Matthew Sivaprakasam and Sean J. Wang and Wenshan Wang and Aaron M. Johnson and Sebastian A. Scherer},
  journal=ICRA,
  year={2022},
}

@article{Manivasagam2020LiDARsimRL,
  title={{LiDARsim}: Realistic LiDAR Simulation by Leveraging the Real World},
  author={Sivabalan Manivasagam and Shenlong Wang and K. Wong and Wenyuan Zeng and Mikita Sazanovich and Shuhan Tan and Binh Yang and Wei-Chiu Ma and Raquel Urtasun},
  journal=CVPR,
  year={2020},
}

@article{Binas2017DDD17ED,
  title={{DDD17}: End-To-End DAVIS Driving Dataset},
  author={Jonathan Binas and Daniel Neil and Shih-Chii Liu and Tobi Delbruck},
  journal={ArXiv},
  year={2017},
  volume={abs/1711.01458},
}

@article{DiazRuiz2022Ithaca365DA,
  title={Ithaca365: Dataset and Driving Perception under Repeated and Challenging Weather Conditions},
  author={Carlos Diaz-Ruiz and Youya Xia and Yurong You and Jose Nino and Junan Chen and Josephine Monica and Xiangyu Chen and Katie Luo and Yan Wang and Marc Emond and Wei-Lun Chao and Bharath Hariharan and Kilian Q. Weinberger and Mark E. Campbell},
  journal=CVPR,
  year={2022},
}

@article{Chang2022MaskGITMG,
  title={{MaskGIT}: Masked Generative Image Transformer},
  author={Huiwen Chang and Han Zhang and Lu Jiang and Ce Liu and William T. Freeman},
  journal=CVPR,
  year={2022},
  pages={11305-11315},
}

@article{Liu2024ARO,
  title={A Review of Sensing Technologies for Indoor Autonomous Mobile Robots},
  author={Yu Liu and Shuting Wang and Yuanlong Xie and Tifan Xiong and Mingyuan Wu},
  journal={Sensors},
  year={2024},
  volume={24},
}

@article{Cao2026GenerativeEP,
      title={Generative Event Pretraining with Foundation Model Alignment}, 
      author={Jianwen Cao and Jiaxu Xing and Nico Messikommer and Davide Scaramuzza},
      year={2026},
      journal={arXiv preprint 2603.23032}
}

@inproceedings{
Ravi2024SAM2S,
title={{SAM 2}: Segment Anything in Images and Videos},
author={Nikhila Ravi and Valentin Gabeur and Yuan-Ting Hu and Ronghang Hu and Chaitanya K. Ryali and Tengyu Ma and Haitham Khedr and Roman R{\"a}dle and Chlo{\'e} Rolland and Laura Gustafson and Eric Mintun and Junting Pan and Kalyan Vasudev Alwala and Nicolas Carion and Chao Wu and Ross B. Girshick and Piotr Doll'ar and Christoph Feichtenhofer},
booktitle=ICLR,
year={2025},
}

@inproceedings{
Zhai2023SigmoidLF,
title={Sigmoid Loss for Language Image Pre-Training},
  author={Xiaohua Zhai and Basil Mustafa and Alexander Kolesnikov and Lucas Beyer},
booktitle=iccv,
year={2023},
}

@article{Hu2020DDD20EE,
  title={{DDD20 End-to-End Event Camera Driving Dataset}: Fusing Frames and Events with Deep Learning for Improved Steering Prediction},
  author={Yuhuang Hu and Jonathan Binas and Daniel Neil and Shih-Chii Liu and Tobi Delbruck},
  journal={2020 IEEE 23rd International Conference on Intelligent Transportation Systems (ITSC)},
  year={2020},
  pages={1-6},
}

@article{Perot2020LearningTD,
  title={Learning to Detect Objects with a 1 Megapixel Event Camera},
  author={Etienne Perot and Pierre de Tournemire and Davide Oscar Nitti and Jonathan Masci and Amos Sironi},
  journal={Neural Information Processing Systems},
  year={2020},
}

@article{Lee2022ViViDV,
  title={{ViViD++} : Vision for Visibility Dataset},
  author={Alex Junho Lee and Younggun Cho and Young-sik Shin and Ayoung Kim and Hyun Myung},
  journal={IEEE Robotics and Automation Letters},
  year={2022},
  volume={7},
  pages={6282-6289},
}

@article{Chen2023ECMDAE,
  title={{ECMD}: An Event-Centric Multisensory Driving Dataset for SLAM},
  author={Peiyu Chen and Weipeng Guan and Feng Huang and Yihan Zhong and Weisong Weisong Wen and Li-Ta Hsu and Peng Lu},
  journal={IEEE Transactions on Intelligent Vehicles},
  year={2023},
  volume={9},
  pages={407-416},
  url={https://api.semanticscholar.org/CorpusID:265033288}
}

@article{CarlevarisBianco2016UniversityOM,
  title={{University of Michigan North Campus} long-term vision and lidar dataset},
  author={Nicholas Carlevaris-Bianco and Arash K. Ushani and Ryan M. Eustice},
  journal={The International Journal of Robotics Research},
  year={2016},
  volume={35},
  pages={1023 - 1035},
}

@article{Maddern20171Y1,
  title={1 year, 1000 km: The Oxford RobotCar dataset},
  author={William P. Maddern and Geoffrey Pascoe and Chris Linegar and Paul Newman},
  journal={The International Journal of Robotics Research},
  year={2017},
  volume={36},
  pages={15 - 3},
}

@article{Lisus2026BoreasRT,
  title={{Boreas Road Trip}: A Multi-Sensor Autonomous Driving Dataset on Challenging Roads},
  author={Daniil Lisus and Katya M. Papais and C{\'e}dric Le Gentil and Elliot Preston-Krebs and Andrew Lambert and Keith Y.K. Leung and Timothy D. Barfoot},
  journal={ArXiv},
  year={2026},
  volume={abs/2602.16870},
}

@inproceedings{
bao2022beit,
title={{BE}iT: {BERT} Pre-Training of Image Transformers},
author={Hangbo Bao and Li Dong and Songhao Piao and Furu Wei},
booktitle={International Conference on Learning Representations},
year={2022},
}

@article{Ranftl2021VisionTF,
  title={Vision Transformers for Dense Prediction},
  author={Ren{\'e} Ranftl and Alexey Bochkovskiy and Vladlen Koltun},
  journal=ICCV,
  year={2021},
}

@article{MurLabadia2026VJEPA2U,
  title={V-JEPA 2.1: Unlocking Dense Features in Video Self-Supervised Learning},
  author={Mur-Labadia, Lorenzo and Muckley, Matthew and Bar, Amir and Assran, Mahmoud and
Sinha, Koustuv and Rabbat, Michael and LeCun, Yann and Ballas, Nicolas and Bardes, Adrien},
  journal={arXiv preprint 2603.14482},
  year={2026}
}

@article{simeoni2025dinov3,
  title={{DINOv3}},
  author={Sim{\'e}oni, Oriane and Vo, Huy V. and Seitzer, Maximilian and Baldassarre, Federico and Oquab, Maxime and Jose, Cijo and Khalidov, Vasil and Szafraniec, Marc and Yi, Seungeun and Ramamonjisoa, Micha{\"e}l and Massa, Francisco and Haziza, Daniel and Wehrstedt, Luca and Wang, Jianyuan and Darcet, Timoth{\'e}e and Moutakanni, Th{\'e}o and Sentana, Leonel and Roberts, Claire and Vedaldi, Andrea and Tolan, Jamie and Brandt, John and Couprie, Camille and Mairal, Julien and J{\'e}gou, Herv{\'e} and Labatut, Patrick and Bojanowski, Piotr},
  year={2025},
  journal={arXiv preprint 2508.10104},
  year={2025}
}

@article{su2021roformer,
author = {Su, Jianlin and Ahmed, Murtadha and Lu, Yu and Pan, Shengfeng and Bo, Wen and Liu, Yunfeng},
title = {{RoFormer}: Enhanced transformer with Rotary Position Embedding},
year = {2024},
publisher = {Elsevier Science Publishers B. V.},
volume = {568},
issn = {0925-2312},
journal = {Neurocomput.},
}

@inproceedings{
mentzer2024finite,
title={Finite Scalar Quantization: {VQ}-{VAE} Made Simple},
author={Fabian Mentzer and David Minnen and Eirikur Agustsson and Michael Tschannen},
booktitle=ICLR,
year={2024},
}

@inproceedings{delbruck2008frame,
  author    = {Delbruck, Tobi},
  title     = {Frame-free dynamic digital vision},
  booktitle = {Proceedings of the International Symposium on Secure-Life Electronics, Advanced Electronics for Quality Life and Society},
  year      = {2008},
  pages     = {21--26},
}

@article{Geist2024LearningW3,
  title={{Learning with 3D rotations, a hitchhiker's guide to SO(3)}},
  author={Andreas Rene Geist and Jonas Frey and Mikel Zobro and Anna Levina and Georg Martius},
  journal=ICML,
  year={2024},
}

@article{Levinson2020AnAO,
  title={{An Analysis of SVD for Deep Rotation Estimation}},
  author={Jake Levinson and Carlos Esteves and Kefan Chen and Noah Snavely and Angjoo Kanazawa and Afshin Rostamizadeh and Ameesh Makadia},
  journal=NIPS,
  year={2020},
}

@ARTICLE{Charbonnier,
  author={Charbonnier, P. and Blanc-Feraud, L. and Aubert, G. and Barlaud, M.},
  journal={IEEE Transactions on Image Processing}, 
  title={Deterministic edge-preserving regularization in computed imaging}, 
  year={1997},
  volume={6},
  number={2},
  pages={298-311}}

@inproceedings{Vaswani2017AttentionIA,
  title={Attention is All you Need},
  author={Ashish Vaswani and Noam Shazeer and Niki Parmar and Jakob Uszkoreit and Llion Jones and Aidan N. Gomez and Lukasz Kaiser and Illia Polosukhin},
  booktitle=NIPS,
  year={2017},
}

@article{Zhang2025Qwen3EA,
  title={Qwen3 Embedding: Advancing Text Embedding and Reranking Through Foundation Models},
  author={Yanzhao Zhang and Mingxin Li and Dingkun Long and Xin Zhang and Huan Lin and Baosong Yang and Pengjun Xie and An Yang and Dayiheng Liu and Junyang Lin and Fei Huang and Jingren Zhou},
  journal={arXiv preprint 2506.05176},
  year={2025},
}

@article{Sapkota2025YOLO26KA,
  title={{YOLO26}: Key Architectural Enhancements and Performance Benchmarking for Real-Time Object Detection},
  author={Ranjan Sapkota and Rahul Harsha Cheppally and Ajay Sharda and Manoj Karkee},
  journal={arXiv preprint 2509.25164},
  year={2025},
}

@article{douze2024faiss,
      title={The {Faiss} library},
      author={Matthijs Douze and Alexandr Guzhva and Chengqi Deng and Jeff Johnson and Gergely Szilvasy and Pierre-Emmanuel Mazaré and Maria Lomeli and Lucas Hosseini and Hervé Jégou},
      year={2024},
      eprint={2401.08281},
      archivePrefix={arXiv},
      primaryClass={cs.LG}
}

@misc{gemma4_2026,
  title        = {Gemma 4},
  author       = {{Google DeepMind}},
  year         = {2026},
  howpublished = {\url{https://deepmind.google/models/gemma/gemma-4/}},
  note         = {Open model release}
}

@article{Levenberg1944AMF,
  title={A METHOD FOR THE SOLUTION OF CERTAIN NON – LINEAR PROBLEMS IN LEAST SQUARES},
  author={Kenneth Levenberg},
  journal={Quarterly of Applied Mathematics},
  year={1944},
  volume={2},
  pages={164-168},
}

@article{Kabsch1976ASF,
  title={A solution for the best rotation to relate two sets of vectors},
  author={Wolfgang Kabsch},
  journal={Acta Crystallographica Section A},
  year={1976},
  volume={32},
  pages={922-923},
}

@article{Olson2011AprilTagAR,
  title={{AprilTag}: A robust and flexible visual fiducial system},
  author={Edwin Olson},
  journal=ICRA,
  year={2011},
}

@article{Umeyama1991LeastSquaresEO,
  title={Least-Squares Estimation of Transformation Parameters Between Two Point Patterns},
  author={Shinji Umeyama},
  journal=pami,
  year={1991},
  volume={13},
  pages={376-380},
}

@inproceedings{Rauch1965OnTM,
  title={On the maximum likelihood estimates for linear dynamic systems},
  author={Harry Ernest Rauch and Fang-Wu Tung and Charlotte T. Striebel},
  year={1965},
  booktitle={AIAA Journal}
}

\onecolumn
\newpage
\pagenumbering{arabic}
\renewcommand{\thepage}{S\arabic{page}}
\setcounter{page}{1}
\setcounter{figure}{0}\renewcommand{\thefigure}{S\arabic{figure}}
\setcounter{table}{0}\renewcommand{\thetable}{S\arabic{table}}
\renewcommand{\thesection}{\Alph{section}}
\setcounter{section}{0}
\renewcommand{\thesection}{\Alph{section}}
\renewcommand{\thesubsection}{\thesection.\arabic{subsection}}

\addcontentsline{toc}{section}{Supplementary Material}

\appendix
\crefalias{section}{appendix}


\section{Bill of Materials}
\label{app:bom}

\begin{longtable}{p{6.5cm} l S[table-format=2.0] S[table-format=5.2] S[table-format=6.2]}
\caption{OctoSense platform bill of materials. Part codes link to vendor product pages where available.}
\label{tab:bom}\\
\toprule
\textbf{Item / Description} & \textbf{Code} & {\textbf{Qty}} & {\textbf{Unit (USD)}} & {\textbf{Ext.\ (USD)}}\\
\\
\textbf{\textsc{Sensing}} \\
\cmidrule(lr){1-5}
\footnotesize Ouster OS1-64 & \href{https://www.dataspeedinc.com/app/uploads/2019/11/DSDistrOusterBrochure.pdf}{\ttfamily\footnotesize OS1-64} & 1 & 13000.00 & 13000.00 \\
\footnotesize Blackfly S USB3: 2.8 MP, Color, C-mount & \href{https://www.teledynevisionsolutions.com/products/blackfly-s-usb3/?model=BFS-U3-28S5C-C\&vertical=machine\%20vision\&segment=iis}{\ttfamily\footnotesize BFS-U3-28S5C-C} & 2 & 982.00 & 1964.00 \\
\footnotesize Kowa LM6HC 6 mm, 1", 1 f/1.8-f/11, C Mount & \href{https://machinevisiondirect.com/products/kowa-lm6hc}{\ttfamily\footnotesize LM6HC} & 4 & 580.00 & 2320.00 \\
\footnotesize ICE UV IR Cut Thin Filter Optical Glass  & \href{https://www.amazon.com/Filter-Optical-Glass-Multi-Coated-Mirror/dp/B0DKCJ3WST/?\_encoding=UTF8\&pd\_rd\_w=BVOyC\&content-id=amzn1.sym.4efc43db-939e-4a80-abaf-50c6a6b8c631\%3Aamzn1.symc.5a16118f-86f0-44cd-8e3e-6c5f82df43d0\&pf\_rd\_p=4efc43db-939e-4a80-abaf-50c6a6b8c631\&pf\_rd\_r=JHCCYRDX1FXN08KANETV\&pd\_rd\_wg=BxRqu\&pd\_rd\_r=384240f7-2fd9-4783-b2ed-40b487ce9f6f\&th=1}{\ttfamily\footnotesize 840337507588} & 2 & 34.95 & 69.90 \\
\footnotesize VN-100T-CR & \href{https://www.vectornav.com/store/products/imu-ahrs/p/vn-100-rugged-imuahrs}{\ttfamily\footnotesize VN-100} & 1 & 1300.00 & 1300.00 \\
\footnotesize SilkyEvCam VGA& \href{https://centuryarks.com/en/silkyevcam-hd/}{\ttfamily\footnotesize EvC3A} & 2 & 650.00 & 1300.00 \\
\footnotesize Camera C-Mount Lens Adapter & \href{https://www.amazon.com/C-Mount-Adapter-ZCZQC-Extension-Security/dp/B09STVLLVY/ref=sr\_1\_3?crid=25OPTYD430QI3\&dib=eyJ2IjoiMSJ9.U1sxGBvTeqbzUS2U21W9derlKRaugxd3T1BLeO3al7itUC6YZ3kDJ1ftni48tnXzprt9eL0Cv7w7yWWiGx7-hJ7q9EBn883MR6F\_evtJmFh1NM18ZoUzmzkEkoTwXKvfu2FHOxg8x8RoAJUXPtMHxOyc1UeU2PIV9QoMdSNdLVEYNnRuSAI2djU1TwNcB3njIkH-kg-4mujiSqPNuasKcm8dFOJDmC0dMxsfoai3rOc.FzE5ODbcT2mtgiS2ZvzVpABONoWrUsqEBeYHPb2IqOo\&dib\_tag=se\&keywords=cs+mount+to+c+mount\&qid=1756742505\&sprefix=cs+mount+to+c+mount\%2Caps\%2C115\&sr=8-3}{\ttfamily\footnotesize B09STVLLVY} & 2 & 6.99 & 13.98 \\
\footnotesize SparkFun GPS-RTK-SMA Breakout, ZED-F9P & \href{https://www.amazon.com/SparkFun-GPS-RTK-SMA-Breakout-ZED-F9P-Concurrent-dimensional/dp/B087T7BV6K/ref=sr\_1\_1?crid=FGIDPBB7WCPE\&dib=eyJ2IjoiMSJ9.GWg\_UMNfW10c8usSIDr38C8\_\_8yvNb8v0dv9ISomqvH9xAwZwER79qyKb7CVVh8CXcu46hWpO\_A3NnP1KYiVblqLwWhQSlL-EPzB9qw9yf2FkeCvey9-sZMBqaNptkNjhWmas29oJIX1Cwa7rrSRFVqmQ7hnOjPq7MhwCE6wRMBFHXV51e6UodRhvjiqAHvGJih3gnLpF4jPIMAcFdVSq9KwK0fOdKROhn094M\_KF10.m-VoxVa5yFYeXVR0x0Kiv\_\_0zLKF\_t5sAW\_ZXT\_lQac\&dib\_tag=se\&keywords=SparkFun+GPS-RTK-SMA+Breakout+-+ZED-F9P\&qid=1756638025\&sprefix=sparkfun+gps-rtk-sma+breakout+-+zed-f9p\%2Caps\%2C101\&sr=8-1}{\ttfamily\footnotesize 845156010493} & 1 & 259.95 & 259.95 \\
\footnotesize VN-100 Pigtail Adapter Cable & \href{https://www.vectornav.com/store/products/imu-ahrs/p/vn-100-rugged-pigtail-adapter-cable}{\ttfamily\footnotesize NA} & 1 & 125.00 & 125.00 \\
\footnotesize waveshare GNSS L1+L2+L5 Multi-GNSS & \href{https://www.amazon.com/Multi-GNSS-Multi-Frequency-Positioning-Waterproof-Temperature/dp/B0CH8BSH5G/ref=sr\_1\_1?crid=I6JIA33YQWD1\&dib=eyJ2IjoiMSJ9.Zc2iR0f4E2mBZMPt2gOkH0G2WZfAP0qwJdK1bm\_uurYE\_xuyzt3DWh918KiZ2iug4CfV9RmFwrQt7iYe0Fy0AIIeLu7c4tl5DK4soJhYjqBnzKkFxPaKIPjicdkYdMK\_0KeggdCMb6xKNk87Rkq82BSlMvpaEkMBGqgrgd5UpLzXgionRKbXYTXHksrHXGJqUZt8Tl73V40kGQ13ubXWIVQdCeD8b6dVaNQjW0cm1Pk.Ffep9GeZs-Iw8NfuORLJZ6o2VxVSXD-JybZSjhpgqT0\&dib\_tag=se\&keywords=l1+l2+gps+antenna\&qid=1756842471\&sprefix=l1+l2+gps+a\%2Caps\%2C130\&sr=8-1}{\ttfamily\footnotesize B0CH8BSH5G} & 1 & 58.99 & 58.99 \\
\footnotesize Spectrum Smart Data Only SIM & \href{https://www.amazon.com/Spectrum-Smart-Data-Only-Card/dp/B0C9K5VLH7/ref=sr\_1\_4?crid=WNFIH47YZGFT\&dib=eyJ2IjoiMSJ9.1ZXv5VHO6iR3p4TFP2CRMgQHHx9pFThsr7jbCWdvU8DoSxo\_p8A2CQRfPZw4KAB-W4K\_PN-y5Zh4CaJRtbl2ht3nLbSuouPcppiNPM7LANalU9h9Q2j58HkLc\_8xGqRDm\_BTXQtXEuvM4yAKBoM6WIInRkhm-tL1nzmNT3mmS6EuqlKYUFQrd\_mipGKn4jstVzhQpGX5XGshhNhfgYCOWOPGDeSxaET7liPgLIltrCk.EjvooB-3ajG2-lIEb5Ik3St52ymweU493RBZie6xJNo\&dib\_tag=se\&keywords=5gb+data+sim+card+gps\&qid=1765320195\&refinements=p\_72\%3A2491149011\&rnid=2491147011\&sprefix=5gb+data+sim+card+\%2Caps\%2C587\&sr=8-4}{\ttfamily\footnotesize 197644272914} & 1 & 4.50 & 4.50 \\
\footnotesize Hotspot Device & \href{https://www.amazon.com/KuWFi-Version-Unlocked-Partner-Wireless/dp/B079FNC379/ref=sr_1_13?crid=15A7UYPYB7YN7&dib=eyJ2IjoiMSJ9.SDS5ekYrTBnp7_r34dWh40rQVEIm3kMbTWvtQKDJ9MbQYxzGSGa5629BAkWnOQVIysN8ki3xSqqR9Wy1yGuYWpR0C1Xzq_Qri3BTVUri3K-whEIRxZiLFO8csaVdrZ3_sJZ5BlAtOkTx-ZWZGTYJ6DQLOtrAfA20zi_odhMu_xQ72jO1SlJPpKvUFLA_iEJhLynnoW2r1pP_TzPHO5d9MQILje-I2u1_pkg3gUMybcc.erUazp-kH5Q7YPAxjLct6zZpx64FCsmsnhMVBMOaRME&dib_tag=se&keywords=wifi%2Bpuck&qid=1775992664&sprefix=wifi%2Bpuck%2Caps%2C136&sr=8-13&th=1}{\ttfamily\footnotesize B079FNC379} & 1 & 44.99 & 44.99 \\
\footnotesize DSD Tech C31-G & \href{https://www.amazon.com/dp/B0CDGD1FSH/ref=dp_iou_view_item?ie=UTF8&th=1}{\ttfamily\footnotesize 794251080989} & 1 & 17.99 & 17.99 \\
\addlinespace[2pt]\multicolumn{4}{r}{\textit{Sensing subtotal}} & 20479.30 \\

\textbf{\textsc{Compute}} \\
\cmidrule(lr){1-5}
\footnotesize Samsung 990 PRO SSD 4TB PCIe 4.0 & \href{https://www.amazon.com/SAMSUNG-Computing-Workstations-MZ-V9P4T0B-AM/dp/B0CHGT1KFJ/ref=sr\_1\_1?crid=VJ6OYNQ6EX6M\&dib=eyJ2IjoiMSJ9.h1VWh7AbYWtFKgCktjwnF2eN3L-N85PRaZju9yo8EktRTjHyewlP5e61KsXOBrM3n67aQ6tSWOyVTKFpSX3i2F2TmXde9w-SmgaB5f8CgpQ6FfIQGkQzrkh6fLlyPoOD5RXV\_3WwnhcWuCZdkThQAZ1eT9UnxZblAJg-ztb-TizHK3Tn7T6WDzXMOc-UWThHiGJl6zueux8\_yNBKMDUXMLJbcItSV4SvbfF49t2EhkQng5cQjOK4ZMuNSTcFDlCW0ui1FzIbBRVzexKHJMvVVtYV9iQ9703rh70G3dU9Xbs.Drkijh9D265tOsCAgE7Me75UBdJkppLj45qVFJcnfUY\&dib\_tag=se\&keywords=samsung\%2B990\%2Bpro\&qid=1756554049\&s=electronics\&sprefix=samasung\%2B990\%2Celectronics\%2C99\&sr=1-1\&th=1}{\ttfamily\footnotesize 887276795195} & 1 & 899.99 & 899.99 \\
\footnotesize Samsung SSD 9100 PRO 4TB & \href{https://www.amazon.com/Samsung-Computing-Workstations-VAP1T0B-AM/dp/B0DX2GJ1YR/ref=sr\_1\_1?crid=35VKKJ91LX2S6\&dib=eyJ2IjoiMSJ9.YNrMe3M1oX5xNzakhktRVgsU-usDsuSKqjI5bWJL4wn9WGGrYNin1-vb\_Yf0KxY-KLLTSBk2AgPN-uCkdGjCq1BYU\_Qc8vWF27L1tA1eW8HdDBJ2lWYwmLJHOuSKs7LSSl\_GesJiDh0IBKGEIibf6\_8nBdb6nNQtVo\_UYV406gFbgYwpVY49comNG5A8wdj3No6TWrZd12l7d6qq4m8xXKXWJziwdtw5XxXNraUyDLs.nn\_WdLecOc5bwBAS-IN39puy3cVnQHV62PLs5hOoEuU\&dib\_tag=se\&keywords=samsung\%2B9100\%2Bpro\&qid=1758064846\&sprefix=samsung\%2B9100\%2Caps\%2C161\&sr=8-1\&th=1}{\ttfamily\footnotesize 887276904894} & 1 & 779.99 & 779.99 \\
\footnotesize AMD Ryzen™ 7 9700X 8-Core & \href{https://www.amazon.com/AMD-9700X-16-Thread-Unlocked-Processor/dp/B0D6NMDNNX}{\ttfamily\footnotesize 730143315593} & 1 & 329.00 & 329.00 \\
\footnotesize MSI MPG B850I Edge TI WiFi Motherboard & \href{https://www.amazon.com/MSI-MPG-B850I-Edge-WiFi/dp/B0FC9CZTXL}{\ttfamily\footnotesize 824142431061} & 1 & 286.04 & 286.04 \\
\footnotesize G.SKILL Flare X5 Series DDR5 RAM & \href{https://www.amazon.com/G-Skill-288-Pin-CL36-36-36-89-Channel-F5-5600J3636D32GX2-FX5/dp/B0BJNKNX6M/ref=sr\_1\_4?crid=2TSWJ6K874FAP\&dib=eyJ2IjoiMSJ9.o3nmviq0A3db7zRbBSp-pfxStEzU7ncC1MIo8uHW31Kn6UpGd\_FFVb6ccsLCwk\_7T\_ILOGE3VG8UfPN\_eyfCsqRta-ZMM7Lt7Xz8M\_ruTCahtUrJkxPCeTUtwV1aFeHEl9vGJKuZKcW\_-0vP7HxgdVqFFzfEtW82x70CJXzxCfuAhB4ZWca5qwnOQnW5XJSKcpaYIfMXKTHtt98WUxGlP4oRRHDio0HfYRhfonXtDj0.ddJc27jSNEPOE3rALm0A-JsOlZ20hSD1\_USq4B7edyM\&dib\_tag=se\&keywords=ddr5+5600\&qid=1756553444\&refinements=p\_n\_feature\_eight\_browse-bin\%3A16955285011\%2Cp\_n\_feature\_nine\_browse-bin\%3A9559992011\&rnid=673240011\&sprefix=ddr5+5600+\%2Caps\%2C100\&sr=8-4}{\ttfamily\footnotesize B0BJNKNX6M} & 1 & 780.00 & 780.00 \\
\footnotesize Thermalright AXP90-X53 Black Low Profile & \href{https://www.amazon.com/Thermalright-AXP90-X53-Profile-Heatsink-TL-9015B/dp/B0BB66YT44/ref=sr\_1\_1?crid=1D2WUZ2Y23CEF\&dib=eyJ2IjoiMSJ9.m90MQM9Jti22wbCyjdNUvIjgKEjktT1H8p7BgGqOu0Mp67xt44mVzEBhiJdSgntTnwB2K0dWHb48rr3up0sjj-8noSjSVOHK49DkyKnu3\_MNxNTUhUXTFMuv1qb5lcvsFP1xj0P7SqzjHD9szr7bd107MnSftwVolPJig\_23ah7uUqhDycQXdOLrWCc0TxTYg9eDnwuUQfCjvPdnriUJZjWVGQbZfbRErHNJpdATU0sV6XGVK7uN6GBFZkgJf\_20x8KJP7O4JPyHk1e5Yp5i28WIE2VhZBjUWsAs8j4M4\_w.iCCrVq4MVBHvUbnR72hRiFkVdT6UD3xSMC9C64Qc\_P0\&dib\_tag=se\&keywords=AXP90-X53\%2BBLACK\&qid=1756553961\&s=electronics\&sprefix=axp90-x53\%2Bblack\%2Celectronics\%2C99\&sr=1-1\&th=1}{\ttfamily\footnotesize B0BB66YT44} & 1 & 21.19 & 21.19 \\
\footnotesize Thermal Paste & \href{https://www.amazon.com/ARCTIC-MX-4-2019-Performance-Durability/dp/B07L9BDY3T/ref=sr\_1\_3?crid=HOH2EINJSJF8\&dib=eyJ2IjoiMSJ9.zEwikJKQpTWbolJ63\_gpP3SGtcTv9kcKz6WDyzBTebeHjmq0DvF8gqh2ds40MsvKzTvMFa-LoND4\_zz\_hpSVBxe1VCvHcwxZKFxJw8OPfi8NtJk5jMxViodi8Jo\_GmfifA\_HeOfLJQcKry-r9gO9wG0F-WZh5VL\_534GDAlvB\_WBaXl\_8ifnnG-ndvS\_YNN0XehYaikkNKtgaWUlPbVDKXLScFLdi3kb5VUlpsPrjWE.\_juJssHE6lkGU9qP0plj5JWrppNFFMD6CBQAF2HuBS4\&dib\_tag=se\&keywords=thermal+paste\&qid=1756553998\&sprefix=thermal+past\%2Caps\%2C133\&sr=8-3}{\ttfamily\footnotesize B07L9BDY3T} & 1 & 6.39 & 6.39 \\
\footnotesize NOYITO 12mm Chassis Switch Metal Button Switch & \href{https://www.amazon.com/NOYITO-Chassis-Extension-Suitable-Computer/dp/B07PPDNPW9/ref=sr\_1\_2?crid=1VT2862WG3S2S\&dib=eyJ2IjoiMSJ9.6WoenQ67aUdpuojnYCfzTgQTrrwYynu7cYHKLVtmqoCzzltQKbScopY2-VvBMc9w\_9Uw1BIRiYor9d6F4Q-d1\_JeTD88denSmYh0f15n71XpTltvfcGeT8r1\_XKyZBZ8eFnSzwbz9c52kqZJxfVdZMlWN4-f2GYVJ24WXl8hDHQRo7RjSgNl2nyTrX6QM3WVAGzrMej-7YyBtoQXMri334hsWAljavp3WBtBB-PJFD4.fyNolZvv1kTeJrs-kOcTgs53bpwnNNpdT4Gmq0eZtzA\&dib\_tag=se\&keywords=pc\%2Bpower\%2Bbutton\%2B12mm\&qid=1763674652\&sprefix=pc\%2Bpower\%2Bbutton\%2B12mm\%2Caps\%2C137\&sr=8-2\&th=1}{\ttfamily\footnotesize 12mm-NOYITO} & 1 & 6.99 & 6.99 \\
\footnotesize Deal4GO NFA765A QCNFA765 2400Mbps & \href{https://www.amazon.com/QCNFA765-2400Mbps-802-11AX-Wireless-NFA765/dp/B0CJ4N5QSL}{\ttfamily\footnotesize 659459623768} & 1 & 20.99 & 20.99 \\
\footnotesize Amazon Basics RJ45 Cat 6 Ethernet Patch 15 ft & \href{https://www.amazon.com/Amazon-Basics-Ethernet-High-Speed-Snagless/dp/B089MGH8W3/ref=sr\_1\_3?crid=DFBJOLEOMKZ\&dib=eyJ2IjoiMSJ9.UqGZgUktqrNKx\_5SSwA5-i96jCUrUxZZjCkk4rbzqrjNrQQoYd1G5Ie37r6RuzC7t91PAkz1md5vllzkBFyVNmGBqrB6jVThZstmj6YMeyMJskaQFlHCTRpklSX1jth6trbTSWynIK0MyoMqpEI4HDdrW5k7VwGAR5dORoJuA4D6RnTg8lH6LgPOXNDg7Tnu5\_hOi4w76xZwTyxWb4a4kgbJ-SZvaCnVNTB3-4crqSY.sm5xiJFostaqeGh6RSaE7AlIvc1c4gjNc7uVbr-daFw\&dib\_tag=se\&keywords=ethernet\%2Bcable\%2B12\%2Bft\&qid=1756828288\&sprefix=ethernet\%2Bcable\%2B12\%2Bft\%2Caps\%2C130\&sr=8-3\&th=1}{\ttfamily\footnotesize 192835003018} & 6 & 6.39 & 38.34 \\
\addlinespace[2pt]\multicolumn{4}{r}{\textit{Compute subtotal}} & 3168.92 \\

\textbf{\textsc{Time Synchronization}} \\
\cmidrule(lr){1-5}
\footnotesize Teensy 4.1 ARM Cortex-M7 Processor & \href{https://www.amazon.com/PJRC-Cortex-M7-Processor-iMXRT1062-Without/dp/B088JY7P2H/ref=sr\_1\_3?crid=8W74BHFFELQI\&dib=eyJ2IjoiMSJ9.HPR4mTzfREjzA9Xhmpg4bF8u2IJc2idy77Hj0hpc6AoDTFNwV7DUfQUS1zabuIewTbLkFkVv9rdHt-RJecVXttPIvQaiEvS6xGPjongOpbeR9lv6ZIQZ5BDTf4r1E\_fkxR20\_6YWxL1b0b\_jE3-4lZ0sB5Hz1UeXYkcRXyS8quxtK4fLybEYLZjkPTnEvyDvuVJ0YcRrUPZkA7NtbVlSwRceQFRedORD8v5Vanc0OnY.UDOtwMHQS51gsTo8jzYiTfvBKFCPm\_G6MfjPs5fzZlA\&dib\_tag=se\&keywords=Teensy+4.1\&qid=1756638922\&refinements=p\_85\%3A2470955011\&rnid=2470954011\&rps=1\&sprefix=teensy+4.1\%2Caps\%2C187\&sr=8-3}{\ttfamily\footnotesize 732387205807} & 1 & 38.21 & 38.21 \\ 
\footnotesize OSH Park Custom PCB Set of 3 & {\ttfamily\footnotesize } & 1 & 37.60 & 37.60 \\
\footnotesize DS3231 AT24C32 IIC RTC Module Clock & \href{https://www.amazon.com/AT24C32-Replace-Arduino-Batteries-Included/dp/B07Q7NZTQS/ref=sr\_1\_1?crid=3H5K2EM98COPC\&dib=eyJ2IjoiMSJ9.Av0ZT44mgzkEZLgrGYpmsSAtCGIkxuMXQLAxgn1qwYAciZrGkWQl3TovOrt-PTkJIcHLMtQ\_OxNPRG6lhaWMOcrDfeyh1pIzwlLkvxD2xn1xzmKHZikPSswdAQ3VlmEu-h3JsAG4\_NJscMrpyHoQGZSf23piySspBORp\_VQQCoHDCAw6v0D74-x0iShCi4KWeprtuj8LB84ywzwuoLMI15rEgFWLPN9NiYdbikrWq-YSiUAScuMakWYa9yegFWw570T-x58J8KjTuEcoFYm6I9\_SKmN\_zKxwLvOZXFRqDBU.thwMcw7dkVlKqOhtDtW30sLcgsqWnC1ee40Ot2Ngyho\&dib\_tag=se\&keywords=rtc\%2Bclock\&qid=1763674045\&sprefix=rtc\%2Bclock\%2Caps\%2C124\&sr=8-1\&th=1}{\ttfamily\footnotesize DS3231} & 1 & 5.59 & 5.59 \\
\footnotesize Hirose HR10 GPIO Cable, 6-pin, 4.5 m & \href{https://www.teledynevisionsolutions.com/products/hirose-hr10-6-pin-circular-connector/?model=ACC-01-3010\&vertical=machine\%20vision\&segment=iis}{\ttfamily\footnotesize ACC-01-3010} & 2 & 43.90 & 87.80 \\
\footnotesize MCIGICM 10pcs Male Header Pin, 40 Pin Header & \href{https://www.amazon.com/MCIGICM-Header-2-45mm-Arduino-Connector/dp/B07PKKY8BX/ref=sr\_1\_3?crid=3B08RA4EKS0Z0\&dib=eyJ2IjoiMSJ9.9jN-S3CtDPJRAShATIrrQ3fezlKfzCW5Qv0-pdq-LAYzimUeWnPFfmsomxT\_HRjnCo8qfSMax32WXFGkRAlGIH4\_AXlFj9D8WDX8hsMFfCUW4pURh2xFXJu8xlxBFQ04ZvLtYbrzxETdFRLfgBmG5X3-IIyDH4HOnQog7Yoq3jLa\_qJpSCEUm2tgl7zaepGVaMrPWsSo7WuJAYOi8BYA3HQW0JymGb8Rjae6u933zAo.tEoDCqWVoWum8IMiVlwTh-0HaAIKgq717Rz3OSusDdU\&dib\_tag=se\&keywords=MCIGICM+10pcs+Male+Header+Pin\%2C+40+Pin+Header\&qid=1756639629\&sprefix=mcigicm+10pcs+male+header+pin\%2C+40+pin+header\%2Caps\%2C105\&sr=8-3}{\ttfamily\footnotesize B07PKKY8BX} & 1 & 4.99 & 4.99 \\
\footnotesize MICRO-FIT 3.0 R-S 10 Circuit 300mm & \href{https://www.digikey.com/en/products/detail/molex/2147561102/12180449?utm_source=oemsecrets\&utm_medium=aggregator\&utm_campaign=buynow}{\ttfamily\footnotesize 2147561102} & 1 & 1.81 & 1.81 \\
\footnotesize Conn Header Vert & \href{https://www.digikey.com/en/products/detail/molex/0430451012/252506?s=N4IgTCBcDaICwGYAMcCsBaAjEzEC6AvkA}{\ttfamily\footnotesize 0430451012} & 1 & 1.52 & 1.52 \\
\footnotesize 24AWG Hookup Wire & \href{https://www.amazon.com/Jameco-ValuePro-Stranded-Twisted-Hook-Up/dp/B0CQMSWVYY/ref=sr_1_6_pp?crid=4NLOR3VNW4P1&dib=eyJ2IjoiMSJ9.SmN2wfootw0K_nSV1yHJdoVe-yr3vvErL9oFmQQbToNaU-wfiGLHsuM_NDeJJHerrqa4Zn-xggzZdbZnnlzt3AL4tNPcYNbWp5nTdhjW5ffJyjtyZxqHwhBaOeM247KHx7Jn9Bje7_NBwBqAU1p8RF6v1FHkdYgEUKPJi-JH1AhKbY2pNiunzRGRd7TDYFecxQIqS6iL32e8RIcf3q5q7MlJLylMI1umVSI_YwJgRNp4Diw9DEzJmZV0rFAt0eYBf-lTBWuaGvLtzLHZUaUUL3qfDw3OJ87LchYJafrVQu4.97qRGbQdEqe2sr9-DzuTVk--AbbAzEW98ne0A-Csj88&dib_tag=se&keywords=hookup+wire&qid=1775990853&sprefix=hookup+wir%2Caps%2C162&sr=8-6}{\ttfamily\footnotesize 00840406205971} & 1 & 16.19 & 16.19 \\
\footnotesize 1072F12-12-2-1-S-00100 & \href{https://www.digikey.com/en/products/detail/cnc-tech/1072F12-12-2-1-S-00100/11203907}{\ttfamily\footnotesize 	
1175-1072F12-} & 1 & 23.03 & 23.03 \\
\addlinespace[2pt]\multicolumn{4}{r}{\textit{Time Synchronization subtotal}} & 216.74 \\

\textbf{\textsc{Storage}} \\
\cmidrule(lr){1-5}
\footnotesize Western Digital 22TB WD Red Pro & \href{https://www.amazon.com/Western-Digital-22TB-Internal-Drive/dp/B0B5W1CQ8W/ref=sr\_1\_1?crid=3QKPL4V9YPW32\&dib=eyJ2IjoiMSJ9.vLMwR1MhpWOs76T61NmV71bNdyCLDV9bcVKJCvMJhxMMTDG-QL1bcxHKJX8gPiPTndi04P7Ft87mxJJVZunf9GnSqiZbhcagzVM\_QvMnm6VdPPeQL4hZ-xeXLQz65BWmLdlFb3aWjL7gfVboz9s1RzXNiJX\_yyoHoUm0TxeKz3yTmDO0fjLEL4YTh8dDr1FhGNnsAnTEuSqzvxBOR7CRPlZoyixd8F03KD1qnA8fdCg.MrCnLgp1gegMSbByIsE5YyfKxFB8OG2w-Z3AB3UPnX4\&dib\_tag=se\&keywords=wd\%2Bred\%2B22tb\&qid=1756828142\&sprefix=wd\%2Bred\%2B22tb\%2Caps\%2C123\&sr=8-1\&th=1}{\ttfamily\footnotesize 718037893501} & 4 & 434.99 & 1739.96 \\
\footnotesize Synology DiskStation 12 Bay DS2422+ & \href{https://www.amazon.com/Synology-DiskStation-DS2422-Memory-Diskless/dp/B09JNRPK2Z/ref=sr\_1\_4?crid=1F8PXPUC4PWOX\&dib=eyJ2IjoiMSJ9.hJpfwIUW3FzOaEn86KFAfDwoKnvCk9BUD5wqT2i1WGpXLQ6h4SOGnQf29Jf2T3SyCUwJPw0WX08707-SakCbwEAgKPnsj62j\_jT3vzV4UsLn4IXuNodAAwejupE2-rO3LDyaCW10z9bBy2lDLfz5RktWkZOcZlhOgymGBZN0ZuNocB2Jin5zMlw5mt2hjEhrGGfZr7xye43P7jfo7mGLpbFNQa3--I38PCUcUPRf85A.CkRGFyslgZ1ZDBElu6kRSRBav3Ggd2Xqp5ULWFWZC88\&dib\_tag=se\&keywords=12+bay+nas\&qid=1756827033\&sprefix=12+bay+nas\%2Caps\%2C128\&sr=8-4\&ufe=app\_do\%3Aamzn1.fos.5998aa40-ec6f-4947-a68f-cd087fee0848}{\ttfamily\footnotesize 846504003983} & 1 & 1999.99 & 1999.99 \\
\footnotesize Synology E10M20-T1 & \href{https://www.amazon.com/Synology-Ethernet-Adapter-E10M20-T1-RJ-45/dp/B089RHK7CF}{\ttfamily\footnotesize 846504003488} & 1 & 229.99 & 229.99 \\
\footnotesize OWC 64GB (2X32GB) DDR4 RAM & \href{https://www.amazon.com/OWC-Compatible-Synology-Rackstation-PC4-21300/dp/B0CJW8K6BP/ref=sr\_1\_2?crid=1ZZSJE0D5CFCS\&dib=eyJ2IjoiMSJ9.ettKUVWZS2msen6HeDEtR\_yD\_jPKYeFjwzPy38HrA2hgAjwSvUZNOWnK7splv1Ek2jPmbLMzPZZoPEp6sYFvgoF3ia22Lp6z2K6gImrQeN\_9oHG436pba9Dw5xFLvwvX193iv9jiuYJhVYLj8sMgiHMB-5yxXr1rUuqQZI3nILayVz7TA2SEnxSxFNreZfPej0EiM3ChTu5tXkm9RHzU\_RwVFOu3\_24eANv6b9BIEgI.nnfBISdTH5S-q-1gApE99A4aw-Y2-CQYYQOebZmYiOw\&dib\_tag=se\&keywords=DDR4\%2BECC\%2BSODIMM\%2B64gb\&qid=1758065806\&sprefix=ddr4\%2Becc\%2Bsodimm\%2B64gb\%2Caps\%2C136\&sr=8-2\&th=1}{\ttfamily\footnotesize 810149202223} & 1 & 646.73 & 646.73 \\
\addlinespace[2pt]\multicolumn{4}{r}{\textit{Storage subtotal}} & 4616.67 \\

\textbf{\textsc{Power}} \\
\cmidrule(lr){1-5}
\footnotesize PicoPSU-160-XT, 160w & \href{https://www.mini-box.com/picoPSU-160-XT}{\ttfamily\footnotesize B005TWE6B8} & 1 & 46.50 & 46.50 \\
\footnotesize  29.4V 5A Li ion Battery Charger & \href{https://www.amazon.com/Tnvodejo-Multiple-Compatible-Superfly-PowerFast/dp/B0BN1CM86B/ref=sr\_1\_3?crid=2K0BKUW339L33\&dib=eyJ2IjoiMSJ9.qU0gFYl0bSHR-lTE9wS3w64RLO5RkWPZbD9jW-t7ctiiKiWbT0bx8B5pQjrnjuhP8O4V3n6QEjHe\_me2OWwifj-8vVw-GDen4z3XtCRY50ohaBhZ2FCBHynVg93MzQHKwAMF6kcIagRMb6BeicOVVKLDJ-AZPZri4rUQru9eS6IlmvdNElEbrHsWjVOZQNO4e4onW2-pPSWW9542lwfrOx5mS2eyrLYe\_C\_CKnJtbEU.zz5HxIKrbxlJV83ynXT5LzejwACsjHfsJRXB5Zgec8E\&dib\_tag=se\&keywords=29.4V\%2B5A\%2BLi\%2Bion\%2BBattery\%2BCharger\&qid=1768922455\&sprefix=\%2Caps\%2C279\&sr=8-3\&th=1}{\ttfamily\footnotesize 786750183001} & 1 & 29.99 & 29.99 \\
\footnotesize Yunsailing 2 Pcs High Precision Watt Meter Power & \href{https://www.amazon.com/Precision-Consumption-Performance-Backlight-Discharge/dp/B0B5KVLQ71/ref=sr\_1\_4?crid=2PTJ5V76BDA3T\&dib=eyJ2IjoiMSJ9.PB2KrQtxp2aQVqESDEIkL4qwMUnT5Lx9yblBYEGcs06vKrVxDvYBrRX7j19O4OBZzKe5iWRa12CkYeostMw0hC-7F9yVqnGBfM4UACdh8osHd5azWgfgiRnOwECm84RQAY4yihdYQbEYnedZ2my7aUIU8TSyymMUYcWdrgmTWzakQVbaNfoxOUOrzUd-sARlUXy3olcdYyU4H87SkYkMuDvMxG8of-FoOSORGKo95buPv4vHDL18HUBIkmYrd2EjPzmzt2HCzPWHfZqKdMrn1V89jydyBYmjMIEacP0nR6k.pPAo4gGVELUeJz-B-AMiA1\_p2G2VLvh7QPXed8zvlK0\&dib\_tag=se\&keywords=inline\%2Bpower\%2Bmonitor\&qid=1762131639\&sprefix=inpower\%2Bmonitor\%2Caps\%2C609\&sr=8-4\&th=1}{\ttfamily\footnotesize B0B5KVLQ71} & 1 & 19.99 & 19.99 \\
\footnotesize LM2596 DC-DC Buck Power Converter Module & \href{https://www.amazon.com/Converter-Adjustable-Regulator-Voltmeter-Display\%EF\%BC\%88Pack/dp/B0F1M24KG3/ref=sr\_1\_1\_sspa?crid=SAIVX9VNKW8T\&dib=eyJ2IjoiMSJ9.CivqPVN7ndLBSTp5EaI9Sciv4gW5j2fA2Lm8cKZJVhiRAnsgr3fjrYEp9iC-ROJvGZrv2CuuRxrsNIVXb3Mm9rRXmsGsXRDJojxW\_z7DSuKP5i\_-0k9KT27xeYh5xwrBNnV1MoA-fydGxlsKZkIUsIBUXNIBhpgj57X9ymQR7weAv5Y7EASJTZS3TzjR0PNr86kZpa6aOy9zIS\_7rJ\_embHfXlTQvRv9CTonZountOM.1tj6VZncbBrV3UJRkYrG3oAlENzE51hwd\_YmBScf6mo\&dib\_tag=se\&keywords=buck\%2Bconverter\&qid=1770319454\&sprefix=buck\%2Bconverte\%2Caps\%2C139\&sr=8-1-spons\&sp\_csd=d2lkZ2V0TmFtZT1zcF9hdGY\&th=1}{\ttfamily\footnotesize LM2596S} & 1 & 9.99 & 9.99 \\
\footnotesize  12V 5A DC Power Pigtail Barrel Plug Connector & \href{https://www.amazon.com/43x2pcs-Connectors-Security-Lighting-MILAPEAK/dp/B072BXB2Y8/ref=sr\_1\_8?crid=1Q315GOUOCIR9\&dib=eyJ2IjoiMSJ9.CURG1Uln\_-RZKd33HnWKRM5q3Jwqo80m5pNuE5DB7I6F453uM4SEIfjP7-FJYHUGXi9gz4-kLEJiYEUJw-Ov3ooi3GRcTjFkkzMjANNgBPhC12h3G3rAESq2-vXL5CbRoXBhGUQUD2zkhqEjZAMwcmsa4uwqZlXk4VzNVvsMAooZWl9wDmAeO8vhCLRL8Xz9hDGu3icSaj-Tk1duk-ldPtWgN5v-hXqteoNPNUT7JB8.vR4LVLJswx0BjdLBNAPXlPbopDJK\_7O7Hf5huOj-inE\&dib\_tag=se\&keywords=2.1+5.5mm+plug\&qid=1761768593\&sprefix=2.1+5.5mm\%2Caps\%2C120\&sr=8-8}{\ttfamily\footnotesize 872629092625} & 1 & 9.49 & 9.49 \\
\footnotesize 5pcs XT30 Plug XT-30 Male Connectors with 10cm & \href{https://www.amazon.com/Connectors-Slicone-Battery-Turnigy-Zippy\%EF\%BC\%88BDHI-18\%EF\%BC\%89/dp/B07M9VK764/ref=sr\_1\_7?crid=2IZWGXM2AI7UN\&dib=eyJ2IjoiMSJ9.oevBDVPAPvvNZ30LcR-BO20Hg\_eEUdKWl9kCAh2FX\_NcXiSUbMya8fZqfAdljm0CyV40reG7irOWGMRmRLPWrlby0Qu\_As6sUR7650X-eYTKnbo9NHFDvvbbx7HDx3VbHPDcyo2jAAjQSiLCwJRJIWYNL5on9kwTGqOaKitQBfecFMqsEK7ZU77r4Ph2vg8eITE4E8T8L8YkbFrW2uVN4aMInxvhhSKeq7PLr9LP0XFjKYOR-j9fQu4fI2wq3r5ECqX\_yQgM4o\_eJbWC9wsN1tjrUZyUjo4UIUBmPA\_C8aA.ppuOT9vtUczdssUE3gQiRwmcOjZXsQGqGwE4F3zLa8U\&dib\_tag=se\&keywords=male+XT30\&qid=1770316888\&sprefix=male+xt30\%2Caps\%2C139\&sr=8-7}{\ttfamily\footnotesize 736900640472} & 1 & 8.58 & 8.58 \\
\footnotesize 24V 12Ah Battery  & \href{https://www.amazon.com/RANSYRI-Battery-Compatible-Lightweight-Replacement/dp/B0F32HLJKZ/ref=sr\_1\_12?dib=eyJ2IjoiMSJ9.c1dQdk8JUV55oL-vujROzWynRbDpFyDLDzXSHI19PP-cjDGXGxtEEiJbHrRHg4ci-8rwihe5UbKIi\_TNl4WqIDvEYyCmB5C\_scuEYT48ccK\_enxj2UYcx-cC90GJ0Sqds1kT22ScyXlQsO0jmRaFXJ6378WSMYha6L88mE2kpqUZtfE9mtuPt36q0MMYZhzmUKo6ENWpEQI6IV3WUK9hU3Rz5WVjaxrLQEOuFmQ1y4H58Fm\_\_YJQP6WulXP96IaJ2qFLEQNoRQF5NXIusHRUhVBPNdhahzRYNnPrUfXFoXk.EJAQeD8-o7aBGwbK7DkdFyb1YPUDHAvmIlYVqfgab48\&dib\_tag=se\&keywords=24v+15ah+battery\&qid=1758124061\&sr=8-12}{\ttfamily\footnotesize 24120} & 1 & 89.99 & 89.99 \\
\footnotesize Victron Energy Orion-Tr DC to DC Converter  & \href{https://www.amazon.com/dp/B01M0HTT8D?ref=ppx_yo2ov_dt_b_fed_asin_title&th=1}{\ttfamily\footnotesize ORI241220200} & 1 & 46.63 & 46.63 \\
\footnotesize 16 Gauge Wire & \href{https://www.amazon.com/Voltage-Automotive-Primary-Security-Electrical/dp/B0CZ73SQP8/ref=sr\_1\_6?crid=33F1K5RJO5AGS\&dib=eyJ2IjoiMSJ9.tgm34ob3B7WC821lpVmxf\_ZVTwdLf5GJfid3xUIZ4tUp2LxW2PbGuDzTOmxIfCQfg726wXm2eYzCom28xURmN2kAHzMVAu0mzir4uwq7j24fHkHtgi2\_tfEIHmQKwrcuhURR83bH4blaNqwR7C9LglsOjPkbbTp5rcxyC0lTpa\_PCGbu5KRkaNN2tG4fFow4IVVB2XIf2IZA8SAEwbU7TEwEU8YmH5Z-DX8f5hSohX8ZE5C8YWCct37xO4AAt3\_40SY2a7LGX95HgxZZSMhjokfwOTgwPGKhJpw-sv5rHHU.zsRzoXG6qUaKAz3n40F1AiPzYyjwLib1NZ9GivQkjWc\&dib\_tag=se\&keywords=16\%2Bawg\%2Bwire\&qid=1758124437\&sprefix=16\%2Bawg\%2Bwir\%2Caps\%2C129\&sr=8-6\&th=1}{\ttfamily\footnotesize B0CZ73SQP8} & 1 & 8.99 & 8.99 \\
\footnotesize 18 Gauge/18 AWG Copper Clad Aluminum 2 & \href{https://www.amazon.com/PUDKLE-18-Conductor-Electrical-Automotive/dp/B0FFGJHPNB/ref=sr\_1\_7?crid=ZJ25A3PX7KD1\&dib=eyJ2IjoiMSJ9.QCz-PGSj0Z3PQ1lFCfjPHgV9BktXg3SR3GF3AQ3CxrNVLeHfdehY4ft0bsLU620u3SaBmx2dDsrokh6KZt1uaT26nKcpN99Y9DEcLvsZi5Jytf\_3NEBr\_GIiZyClZcPA7mxuY9To3Vrzgt0gBsskfmUdcSR2HEAkLp4ShY\_YBk\_tUnG09yxlxSGGUkBw8iPZHGZOGQZWE59rvUqhxNqOaN8BsuRJoFGpJL1xu8rjFod01ZR078tmLuV01TYGx-ar56aNWZdrx3Ku\_Ldpy182r92u8lm38WHvNLoLOvMMwsA.AZAG2hDEGMRWrylT6xFcr8rsy9crk5kSjpOTYigDQl4\&dib\_tag=se\&keywords=18\%2Bawg\%2Bwire\&qid=1756653176\&sprefix=18\%2Bawg\%2Caps\%2C129\&sr=8-7\&th=1}{\ttfamily\footnotesize B0FFGJHPNB} & 1 & 10.79 & 10.79 \\
\footnotesize Alex Tech 25ft - 1/2 inch Cord Protector Wire Tubing & \href{https://www.amazon.com/Alex-Tech-10ft-Protector-Sleeving/dp/B07FXF12HC/ref=sr\_1\_3?crid=1MAEM448XGWS7\&dib=eyJ2IjoiMSJ9.p8wGWvy-D5MMveSobdWdAmCUpPbnUziRm9oRVZ3UZJgULoKRzouMg4W0yTXJreeYyNMIEOSgIzU6yW-VnDDdkc4MI6AJ-kx\_iUrS5L4ZzJbYgEpJ13PXB5kef\_mJNm-vq92WCdEuYs1vMT7qxbLjrSuH6UfYi-k2ARZdo99U1q3k4d1YMDiVYDpTE0m3OVrBcwv81g0Whz-AFpM0hNHwJqTfw1PnKBaE5YS5wBCbmYHLRuAr-33X08f9fcgHmJif7E1Xr7nFGxH86SsaXRevk8T06JoSVPWZiy1Lj\_RIoJo.kcS2Or9r9JSJON6FzoVa0xGls6KbeGXISlLyTBXfBV8\&dib\_tag=se\&keywords=Protector\%2BWire\%2BLoom\&qid=1760306236\&s=electronics\&sprefix=protector\%2Bwire\%2Bloom\%2Celectronics\%2C117\&sr=1-3\&th=1}{\ttfamily\footnotesize 704256862602} & 1 & 14.99 & 14.99 \\
\footnotesize 12/2 Gauge 20 FT Red Black Hookup & \href{https://www.amazon.com/HiFind-Electrical-Extension-Flexible-Lighting/dp/B0D921S1G8/ref=sr\_1\_6?crid=CSLP9E4EHMYK\&dib=eyJ2IjoiMSJ9.0sAofHNjtrRg2h4YjGh4Vco5gay7tM0FQXcgo-tNGPL6TaTJbKrksAoOYeLPB3cAIoCahGw75SrLWZWqWhY2FZw35dbiNlAsOhW\_ily9GY9QVbeVme1fxdvLNjwvCg0\_hLttjIkgA\_Zj\_YNSLcE1rNUp0v4NI6tSj9wJQAsKJEJ1ORXGDJeUS5\_ep5vGvymPw-HTkZ4XX3UnS-AXGzHOwxMMqjzJilUV\_Vodjtq0aTF2ftq1IdBaGhH\_l7GxG\_aPcMJp3hV6EKQjk9LeCb8UPCM-a7OrxIfMJDr9F\_F323Y.bwMkBdKEJZ7K\_CYFnUW77n7jSiUUy5U9Kt23QYUwEVc\&dib\_tag=se\&keywords=12\%2Bawg\%2Bwire\%2B20ft\&qid=1756653643\&sprefix=12\%2Bawg\%2Bwire\%2B20f\%2Caps\%2C129\&sr=8-6\&th=1}{\ttfamily\footnotesize B0D921S1G8} & 1 & 9.99 & 9.99 \\
\footnotesize 10 Pairs 5.5x2.1mm DC Barrel Male Connector & \href{https://www.amazon.com/5-5x2-1mm-Connector-10Pairs-Pigtails-Security/dp/B0DPKTQFBV/ref=sr\_1\_11?crid=2QM86PNDT0RUS\&dib=eyJ2IjoiMSJ9.59CF0JSEwdzUtzgLS6NsULTnTWw2iRHH-XY0hSZ2F6CaXmdkPPUrlq6GAfp5ZskFIR6Gq51LSdtF7UNIrrhUTPp8MkUR9J73PjApe62NHbkHXma0WuSIOcvx1zqsaODDw8cHTyvyPsf6W7VhIfCFNpjXQY0s65h3xSRvQyJItuzX2v70o9bsVEA9iqBolewwlUEcsKAzkaVyGOCEvqj61UBMZglIHfzKNK\_4rHgRjlw.x99XE\_hkVv5A6BS40HGkBScSY1L3aN\_VSL42aiyIymw\&dib\_tag=se\&keywords=barrel\%2Bconnector\%2B16awg\&qid=1765321827\&sprefix=barrel\%2Bconnector\%2B16a\%2Caps\%2C558\&sr=8-11\&th=1}{\ttfamily\footnotesize B0DPKTQFBV} & 1 & 13.99 & 13.99 \\
\footnotesize Fancasee 10-Pack 18AWG DC Power Pigtail Cable & \href{https://www.amazon.com/Fancasee-10-Pack-Connector-Surveillance-Security/dp/B0F7RLHNJV/ref=sr\_1\_9?crid=O7YBAJ86GMAJ\&dib=eyJ2IjoiMSJ9.CMpM-XSwXogPllLOsl-UNpFrHHHho5BCSX5ygsx0pREBt2Cig7XpR6NYsoypnPJKb9qnpL-PDjOV8Yc-0H62AGsdoN8l0-4rWlJUBe\_Z8poxDunrv0ZWZwuvi9FYq4g8CJjXnRRAjHDNEANqdGaJ7hVXyhsCZufV5hjWL-7CUgxA0f\_fxzwq\_CbiHpSb2EoMsVwbKVOnownmlXvkIoZoNxupVL3bEZOFu1\_hvuyLR80.jGq7prtd0fVHl0XjA2piOBG1pK5bbV\_-N9CXlPushP8\&dib\_tag=se\&keywords=barrel+jack+power+connector+5.5+x+2.5\&qid=1756728435\&sprefix=barrel+jack+power+connector+5.5+x+2.5\%2Caps\%2C105\&sr=8-9}{\ttfamily\footnotesize B0F7RLHNJV} & 1 & 13.89 & 13.89 \\
\addlinespace[2pt]\multicolumn{4}{r}{\textit{Power subtotal}} & 333.80 \\

\textbf{\textsc{Interconnect}} \\
\cmidrule(lr){1-5}
\footnotesize Brainboxes 5Port Gigabit Ethernet & \href{https://www.amazon.com/Brainboxes-Gigabit-Ethernet-Switchtemperature-SW-015/dp/B08FMW62MB/ref=sr\_1\_3?crid=1VKVZY47W3BHW\&dib=eyJ2IjoiMSJ9.d\_qmg4Csii3CcEzNkLsJNMKOED65NUUO05ukSn3E0IQC0yqihwz3GLtHZzWBUILwtZG5aXFVmeHCBGKsmh7W8t9batRiC3lZDPZ-6WRpbFCpgzo0ImW8dmisdnhy6mranASUuSVwTlQZXCiSEHUk\_1LW\_e5KO6hBTpeULFz9tvvchKVFzyP6IOJ07zQ9EhtH237X83GfWSDy\_CkvwsQEz2mfczij6fdQnKaLeFE4RO4Ffh0FMxvGJD04y075350rDrkri945f6yo0\_fpfzPaRaMePLARZ5uGPGM\_MdLGB1o.zkGZKId5WEIsI4udjM1P0LQKjeWUDF1IWZfaMvnYXG0\&dib\_tag=se\&keywords=Brainboxes+Ethernet+Switch+sw-015\&qid=1760351170\&s=electronics\&sprefix=brainboxes+ethernet+switch+sw-01\%2Celectronics\%2C128\&sr=1-3}{\ttfamily\footnotesize B08FMW62MB} & 1 & 85.86 & 85.86 \\
\footnotesize DSD TECH SH-U09G USB to TTL Serial Cable & \href{https://www.amazon.com/DSD-TECH-SH-U09G-Serial-FT232RL/dp/B083HVM7VZ}{\ttfamily\footnotesize SH-U09G} & 1 & 13.99 & 13.99 \\
\footnotesize Acer USB Hub 4 Ports, Multiple USB 3.0 & \href{https://www.amazon.com/Multiple-Splitter-Laptop-Extender-Windows/dp/B0CN3F9Y1Z/ref=sr\_1\_6?crid=2KV6S6A2O6VHB\&dib=eyJ2IjoiMSJ9.kzz5HVMUN-\_3HE3Z7O3q7tjpqKoiA3F06IW1nTnWRR-p4nj83VjPxBu-4sGcnuWbq95kvPR9jnufnLPRq-tr5PIzHOiMQZp4ex3kyGwsz5XPG\_MN6cHJtEovVKsrnNRQz2RLJku9G-NiuRVQr5ZoVG0TE6WtzdABfxcmNta\_E6TIOyL7r3GRAqgENPHVbfBwzak1oXI5wPeSTqNIFjucgbQjvJrWAWMGDeF4klPxpnE.nmUF4OHp60U4r979d-8BVJFUZ0Yk8J4mCVTRM5\_8esw\&dib\_tag=se\&keywords=usb\%2Ba\%2Bport\%2Bextender\%2Bdata\&qid=1760351673\&sprefix=usb\%2Ba\%2Bport\%2Bextender\%2Bdata\%2Caps\%2C119\&sr=8-6\&th=1}{\ttfamily\footnotesize 195133227902} & 1 & 9.99 & 9.99 \\
\footnotesize PoE Texas DC-Powered PoE+ & \href{https://www.amazon.com/PoE-Texas-DC-Powered-Injector-Converts/dp/B0CRVQ4GZ1/ref=sr\_1\_2?crid=2B4ZZ3KV0HRB1\&dib=eyJ2IjoiMSJ9.yNsm7oBwlEDkZ5TMYTNjmMr0NYacKa52Cs\_uCuBXrrJQ9GcmfojNrDl7fevA4KzH8GQvesor-nfpGM3jM8D1S1\_zaOq2xgSofBTA6ADcEA7OaGAMxkTYkyyWk\_qi8X4K8RfJEzYUD41xSW-443KtkxRefRvKvHnpKNLdjiN4w7QfeZ3cys2NlvydAACmPHiMmXQaZ35326yMnDRMtpi7\_2bThU6wHt4cMuduXhjna5o.4g\_oCc\_RwLXZwU841vC9c9C2UpTmOjOXIY3z3shAnGE\&dib\_tag=se\&keywords=poe+injector+battery+24v\&qid=1761768480\&sprefix=poe+injector+battery+24v\%2Caps\%2C134\&sr=8-2}{\ttfamily\footnotesize 655360670907} & 1 & 21.99 & 21.99 \\
\footnotesize Ginsco 580 pcs 2:1 Heat Shrink & \href{https://www.amazon.com/Ginsco-580-pcs-Assorted-Sleeving/dp/B01MFA3OFA/ref=sr\_1\_3?crid=3T6GVBDONNIRD\&dib=eyJ2IjoiMSJ9.afAtOFc0AWv7IewHnY-Ssvj0\_hiD56x36qChq\_os\_H5I\_pO0FoCoHaQ1yiyiGepVfjWYOY30aqDVmfGdN5LlAqrXx2IXhcrjMaMzK8UlTezDQULVhf3hFBEy80GCZP\_G0zfh1PxvPljI\_mBPB86SOupDP3wG0hu7tanhE0q0BXT3atWsLIP1jpOfwpuyyqylZsqkRjgdb3M3pv7fMu6vm5FgJXnAKBbt86Ip8AuOhPU.mrOhBwQ0YOISSaI3gnSFtvFRMKh8WEVpyPdDNI3sphc\&dib\_tag=se\&keywords=shrink+tube\&qid=1756912031\&sprefix=shrink+tubei\%2Caps\%2C112\&sr=8-3}{\ttfamily\footnotesize B01MFA3OFA} & 1 & 6.99 & 6.99 \\
\footnotesize Micro USB Cable 1 Feet & \href{https://www.amazon.com/Charging-Android-Charger-Samsung-Devices/dp/B0BLL6QW4T/ref=sr\_1\_5?crid=1J5Q57DNUO6KO\&dib=eyJ2IjoiMSJ9.5NqoKgxz5s43vyXL-gn6mMFofuy-8ZrwGMEnjrJhXOMeRlBkxK1pAZnusyAp8a9yBGyA\_vSM93htIPKmTerIkKK0R53V2JvhFxDi\_2Ua-yFB\_2vnd38lrj\_Awsx9t0KSGTduC\_9dy-ADRXOZC-tSGEn1bpPIIkKjk0KxzZKQMiaarY2GkoE3G-q4R4RQkHxMjkynJ7uZ-AkuT4veaqY6cCjrQnmJOHlo5wrFGa3W5VI.7dM2TXQjJq4mDwgJt5VFmk6nBVEHAJqN-ZxmVXxljGU\&dib\_tag=se\&keywords=micro\%2Busb\%2B1ft\&qid=1762364515\&sprefix=micro\%2Busb\%2B1ft\%2Caps\%2C122\&sr=8-5\&th=1}{\ttfamily\footnotesize B0BLL6QW4T} & 1 & 4.69 & 4.69 \\
\footnotesize Short USB C to USB C Cable (1.5ft 2 Packs), & \href{https://www.amazon.com/Transfer-Charging-Compatibile-Samsung-MacBook/dp/B094V4RJGC/ref=sr\_1\_4?crid=2DVEIDQN7GO1O\&dib=eyJ2IjoiMSJ9.pz1QCOl2zoY-gxEH6WfNY\_om6aJkhkJ4xqPARuzWlWGq0ZwLgV1evrAB2copNZQqglDqQ2DQRPsF3Lbc\_k-r9IJ7gO6VEeV9gnoYEeRRxWhSsxuXo4Hj3MT-DmVm1y8vupkWv2F4jd3AJpRulu\_fax\_7ZI8R4ph2-Hqk0VE8j5Ryj5oIXFbH17es22kXSbMx27mEnLorAcnjTN6tVhBerKz4MfEGgfWqqer0iRi5IWI.G94-hI5nGSlr2W1Mm\_UElg7GaVcGrZtLoEkKPJyZ\_i8\&dib\_tag=se\&keywords=USB\%2BC\%2Bto\%2BC\%2Bcable\%2B2ft\%2Bdata\&qid=1762354570\&sprefix=usb\%2Bc\%2Bto\%2Bc\%2Bcable\%2B2ft\%2Bdata\%2Caps\%2C127\&sr=8-4\&th=1}{\ttfamily\footnotesize B094V4RJGC} & 1 & 9.99 & 9.99 \\
\footnotesize SUNGUY USB to USB C Cable 2FT & \href{https://www.amazon.com/SUNGUY-Charging-Transfer-Compatible-iPhone15/dp/B0CWGQGNS7/ref=sr\_1\_8?crid=ORCHPOY6WG94\&dib=eyJ2IjoiMSJ9.bNW108XdT9Ltv7Kwuq7mj1WMUFYcGaFTLNIYbFzLncXOk-SBakcKkT0g2SbDLUsasKuXN6559fhSx\_IO1V1gqkszZjyeTWCJwilXMjDiqLOcK-NtROywZ0t\_wKRcAOVU\_KdENRprQ5K3aYUhTK-aIVSdVwKeh1UeRlmx0fO7psW4CUfojSd3mf-PV\_JN2K2ENk8MNOP2itSOQ14\_DcGCt67IaDrZ2kDEtdkRyy\_V\_80.MC-HE0t-DeFDyjQdZ0wDIxV6UOVE0yGF6p9FdseX1BI\&dib\_tag=se\&keywords=USB\%2BA\%2Bto\%2BC\%2Bcable\%2B2ft\%2Bdata\&qid=1762354625\&sprefix=usb\%2Ba\%2Bto\%2Bc\%2Bcable\%2B2ft\%2Bdata\%2Caps\%2C128\&sr=8-8\&th=1}{\ttfamily\footnotesize B0CWGQGNS7} & 1 & 6.99 & 6.99 \\
\footnotesize SUNGUY Micro USB 3.0 Cable 1.5FT, 5Gbps USB A Male to Micro B & \href{https://www.amazon.com/SUNGUY-Compatible-Galaxy-Toshiba-Canvio/dp/B09HJNFGGB/ref=sr_1_1?crid=O9ZO46NMNX5F&dib=eyJ2IjoiMSJ9.CW_GaDnTkzlZ4_v3r70gHArknZ8cM3YcCBrgzkOcuQDV0YtbqDFg_6SwYZICr-ffJlC0EmVLFuYqEqiXKeAokll8TU9c8GOmc2aVjts3qeuaJVDIyFy87g3WqVWcd4Mo9yKNdokA8SqfVB_0LJNJ06fEXmiZd-Xpx_jw-P3drKFP9esOHv1r2nxi4z6HFPoEn1nWPb-dVNpZt-s2H0rLY_9uazAcXVnOvbPTcuWBWq8.ryOMF3GzuM-ZrM6pp1pRunugLwS4nsI7nXc01fhn5LE&dib_tag=se&keywords=USB%2B3.0%2BA%2BMale%2Bto%2BMicro%2BB%2BMale%2BCable%2Blocking&qid=1761287760&refinements=p_n_g-101014941094111%3A119746480011%2Cp_n_top_review_rated%3A207707406011&rnid=207707405011&sprefix=usb%2B3.0%2Ba%2Bmale%2Bto%2Bmicro%2Bb%2Bmale%2Bcable%2Blocki%2Caps%2C328&sr=8-1&xpid=g6M5L3lwz2AIu&th=1}{\ttfamily\footnotesize 606184479795} & 2 & 6.99 & 13.98 \\
\footnotesize Cat 6 Ethernet Cable 1.5 ft, Indoor/Outdoor High Speed & \href{https://www.amazon.com/JARNHNG-Ethernet-Internet-Computers-Televisions/dp/B0D6FZ24FP/ref=sr\_1\_3?crid=2P85WU27MV3JF\&dib=eyJ2IjoiMSJ9.onJ4gLiEEGdy9Oye-q0vtkl6mgQp\_XcinunEpcuFpOyRKrqDBb9nSAraUr2cvBlgIYwt0dO3Rkfu0xy0z12Hhqv7aGaye-RuwIy8QqwNt6GozKCI2VfSqarnGQFzEozGNUa19jXqulzLHcgkaAhR3lmOu0FEGgToNv8TfL0Hy0ThrcnXO3yhZeAVWVHCOK9hLJEqWMHvsVNk8CH-fIJm9\_a5Axt0RIu86BWyMecWnQ4.s5lSEsm783PRnXR3yL\_5uJKcFl\_ZthoMSqXI545k6gM\&dib\_tag=se\&keywords=1.5ft\%2Bethernet\%2Bcable\&qid=1762357539\&sprefix=1.5ft\%2Beth\%2Caps\%2C119\&sr=8-3\&th=1}{\ttfamily\footnotesize B0D6FZ24FP} & 3 & 1.99 & 5.97 \\
\addlinespace[2pt]\multicolumn{4}{r}{\textit{Interconnect subtotal}} & 180.44 \\

\textbf{\textsc{Mechanical}} \\
\cmidrule(lr){1-5}
\footnotesize 580 PCS M3 Black Nylon Standoff Kit with Spacers, & \href{https://www.amazon.com/Black-Nylon-Standoff-Spacers-Screws/dp/B0F297R23T/ref=sr\_1\_5?dib=eyJ2IjoiMSJ9.Nhq3VwxGOpdGGMrBOfRAY5gzu1ZU32xwLUiMfiuk\_zNUfoN8-TniE5\_Lf8XJwa7xfN\_zVX7qOgh0bZgjx7wgq71P0OpgQMpDuK1njw6\_TK-CcPaivcq5JHUs5Yyx8dyhJ3OkEbqIj5BFy7gZP255alz10w8-757sq9hEvI3LMU6tlze-WPc1tX9uPWSw\_GmM5YMOS1f3eBAQYM0vVpauDgfEL4Pvu8kOv--6xyrWPr4.IQUoMZopx4tOHiEN089RMVmbVoHyOoUAhLF3Dt\_m\_VM\&dib\_tag=se\&keywords=m3+standoff\&qid=1756743675\&sr=8-5}{\ttfamily\footnotesize B085H7GXHG} & 1 & 6.99 & 6.99 \\
\footnotesize Low-Strength Steel Square Nut & \href{https://www.mcmaster.com/96887A111/}{\ttfamily\footnotesize 96887A111} & 1 & 7.82 & 7.82 \\
\footnotesize 60 PCS 6 Inches Reusable Cable Ties & \href{https://www.amazon.com/Reusable-Newlan-Adjustable-Organizer-Management/dp/B081HH5X61/ref=sr\_1\_3?crid=17FTDN7DIHQ0Q\&dib=eyJ2IjoiMSJ9.UbumXohXNv4XTUtObG3wOZGzSEN9H7BszV7ZFifnE0X7PqDtyzfDLyu2XFEUlEpvgLl-gEovKjfQ5OEryHP8oB459poGrUILZgk95kzolp5ER9yAAF5rNuYiLis\_neLpeMRZ08jrD5WnuJYjK5GjY5l059T-tj25wFQQ\_QhIIlBqdql2ZM4OiptAdEaJ21YgmbwaEpZzhFtiAKyLLJP9m6L9gABygcQ2b6anScXGQe\_\_6aYqudPxYZCduU5zYvAIJ-G1EfwS4fWpkXUpIkOwmREQZq-zvh-B28x-ht2sATY.co7g\_XSXPUkhPBe0s3Ikt1p9c5vPSFnUuDtbChAnXUY\&dib\_tag=se\&keywords=Reusable\%2BCable\%2BTies\&qid=1760306252\&s=electronics\&sprefix=protector\%2Bwire\%2Bloom\%2Celectronics\%2C165\&sr=1-3\&th=1}{\ttfamily\footnotesize 701722438798} & 1 & 6.98 & 6.98 \\
\footnotesize 50 PCS M5 x 0.8 mm Thread, 8 mm Long Socket Head Screw & \href{https://www.mcmaster.com/products/91290a194/}{\ttfamily\footnotesize 91290A194} & 1 & 11.52 & 11.52 \\
\footnotesize 25 PCS M4 x 0.7 mm Thread, 8 mm Long Socket Head Screw & \href{https://www.mcmaster.com/products/91290a180/}{\ttfamily\footnotesize 91290A180} & 1 & 7.96 & 7.96 \\
\footnotesize 100 PCS M3 x 0.5 mm Thread, 10 mm Long Socket Head Screw & \href{https://www.mcmaster.com/products/91290a110/}{\ttfamily\footnotesize 91290A110} & 1 & 8.74 & 8.74 \\
\footnotesize 100 PCS M3 x 0.5 mm Hex Nuts & \href{https://www.mcmaster.com/products/90592a085/}{\ttfamily\footnotesize 90592A085} & 1 & 6.31 & 6.31 \\
\footnotesize 10 PCS M3 Female-Female Hex Standoff, 40 mm Long & \href{https://www.mcmaster.com/products/90592a085/}{\ttfamily\footnotesize 92510A330} & 1 & 9.88 & 9.88 \\
\footnotesize 4 PCS M3 Rubber Vibration Damper Standoff & \href{https://www.mcmaster.com/catalog/132/3736/92510A330}{\ttfamily\footnotesize B07WZY8N3K} & 1 & 6.99 & 6.99 \\
\footnotesize 98.4" Long 20 mm x 20 mm T-Slotted Framing Rail & \href{https://www.mcmaster.com/catalog/132/3736/92510A330}{\ttfamily\footnotesize 47065T101} & 1 & 48.37 & 48.37 \\
\footnotesize 24" x 24" x 0.125" Thick Multipurpose 6061 Aluminum Sheet & \href{https://www.mcmaster.com/products/89015k48/}{\ttfamily\footnotesize 89015K48} & 1 & 93.90 & 93.90 \\
\footnotesize 24" x 24" x 0.1875" Thick Multipurpose 6061 Aluminum Sheet & \href{https://www.mcmaster.com/products/89015k33/}{\ttfamily\footnotesize 89015K33} & 1 & 118.11 & 118.11 \\
\footnotesize 8" x 8" x 0.5" Thick Multipurpose 6061 Aluminum Plate & \href{https://www.mcmaster.com/products/9246k31/}{\ttfamily\footnotesize 9246K31} & 1 & 38.73 & 38.73 \\
\footnotesize 240mm x 240mm x 3mm Carbon Fiber & \href{amazon.com/FANCYWING-240X240X1-0MM-Laminate-Surface-Thickness/dp/B0BL77Q41G/ref=sr_1_2?crid=3MUNMHNKWKJ1&dib=eyJ2IjoiMSJ9.KkT0oJ2yajRW9lH4_i0gPg.BmoBSi716knDckofir3EsHaBJ9flz64oJOpSluHk6tM&dib_tag=se&keywords=B0BL77Q41G&nsdOptOutParam=true&qid=1779901160&sprefix=b0bl77q41g%2Caps%2C130&sr=8-2}{\ttfamily\footnotesize B0BL77Q41G} & 4 & 30.98 & 123.92 \\
\footnotesize 1 KG Bambu Lab ABS 1.75 mm Filament & \href{https://www.mcmaster.com/products/filaments/?s=b0cgr29r63}{\ttfamily\footnotesize B0CGR29R63} & 1 & 18.99 & 18.99 \\
\addlinespace[2pt]\multicolumn{4}{r}{\textit{Mechanical subtotal}} & 515.21 \\
\textbf{\textsc{Miscellaneous}}\\
\cmidrule(lr){1-5}
\footnotesize Kikerike Self Adhesive Foam Tape Weatherstrip 1In x 1/2In x 20Ft High & \href{https://www.amazon.com/Kikerike-Weatherstrip-Insulation-Stripping-Soundproofing/dp/B0FKMRZ817/ref=sr\_1\_18?crid=B5D7B7RIDD6O\&dib=eyJ2IjoiMSJ9.q88xJrNhWEMQt79CP36yY6kGMi9pF\_rKYL8WusOy9jtLkl7za0nNLQI61a4XDi8Sty7BN3Ln5z-3-WHEzsRsBJyJvQkBhLsWgA\_AI\_sd\_OkP4MGjw4MwcDGGZSRDwmh1-X7kkx\_7DyL6ecEUTq9LH\_AhYqDt4rYZkB9odLfW9InPSIXSI\_yRZPYvXkGp288VQMKvXI8eUUHPeAf7fOngOA.5KDKnXQFFJOTNcZmSTUZjLiwU2aQyIQo4OdG67sXPK8\&dib\_tag=se\&keywords=rubber\%2Bfoam\%2B1\%2F2\%22\&qid=1768922605\&sprefix=rubber\%2Bfoam\%2B1\%2F2\%2B\%2Caps\%2C149\&sr=8-18\&xpid=YRfPX7lUs5vMo\&th=1}{\ttfamily\footnotesize B0FKMRZ817} & 1 & 17.99 & 17.99 \\
\footnotesize 3M Double Sided Tape 0.5”×18ft & \href{https://www.amazon.com/Double-Mounting-Waterproof-Strong-Adhesive/dp/B0CQ1XKW3N/ref=sr\_1\_2?crid=3DZD8J2Z8EJOL\&dib=eyJ2IjoiMSJ9.WYATUye6UGO0VsvBdHd6GDUDcfYo\_xyxFfJfj\_CMXr\_0OA8i4hmtFBlLj4MLd7UaqXt6hrvQcXlyVgxT50CfHFm-dfz8-SVL-ERM2euH9QFMQV4AKWjsuqklzq-t6Z5MEVy1RwoQizHvbp86JrY5N3nOhBe5eqgarPhxJQjWvqwYV4FiwBHQV6Jfvqm7frNRl1\_mFLu4y6igRDPKE82ZUctibDv4CSkr34qW9lOTT9c.RMeGEb6vd9tWNKrnvnRNexxffUt9cn4O2UOXVYEyMnQ\&dib\_tag=se\&keywords=3m\%2Bvhb\&qid=1763674831\&sprefix=3m\%2Bvgb\%2Caps\%2C144\&sr=8-2\&th=1}{\ttfamily\footnotesize 5925} & 1 & 9.99 & 9.99 \\
\footnotesize Scotch Vinyl 700 Electrical Tape & \href{https://www.amazon.com/Scotch-Heat-Resistant-Listed-Certified-Electrical/dp/B001AXD0EY/ref=sr\_1\_3?crid=1XYXOZIS92WXF\&dib=eyJ2IjoiMSJ9.lNltt7cumfXsZCkJlMxFYZFrSxB7kOViFSdf9c29UQHJVGVcXiP9invgT7YKIFiqvGfPoKDu\_0sVvvUogvLfbXWQ1ROYAC6XPsfku2Cb2WXt-uZMysDNHGtNtdSef9nzvPmX2Aoq4uy6b4N7Nvd1xkHWQZ4BdcUENvEbH\_gI-mx1VYL1n0D2CN0Y\_lCrKJxCe8\_rwXL6Pk5UeNU8e7ombBTDOZCE9t9t0odQW2UWhZc.y44SkZUMB6AjJNM4trYnjnlM43GZKeb09h\_QIOHBOsY\&dib\_tag=se\&keywords=electrical\%2Btape\&qid=1763675633\&sprefix=electrical\%2Btap\%2Caps\%2C162\&sr=8-3\&th=1}{\ttfamily\footnotesize 5961-01-139-} & 1 & 2.98 & 2.98 \\
\addlinespace[2pt]\multicolumn{4}{r}{\textit{Miscellaneous subtotal}} & 30.96 \\
\bottomrule
\multicolumn{4}{r}{\bfseries Total} & \bfseries 29542.04 \\
\end{longtable}


\section{Details of OctoSense sensors and hardware}
\label{app:octo_detail}
\begin{figure}[h]
    \centering
    \includegraphics[width=\linewidth]{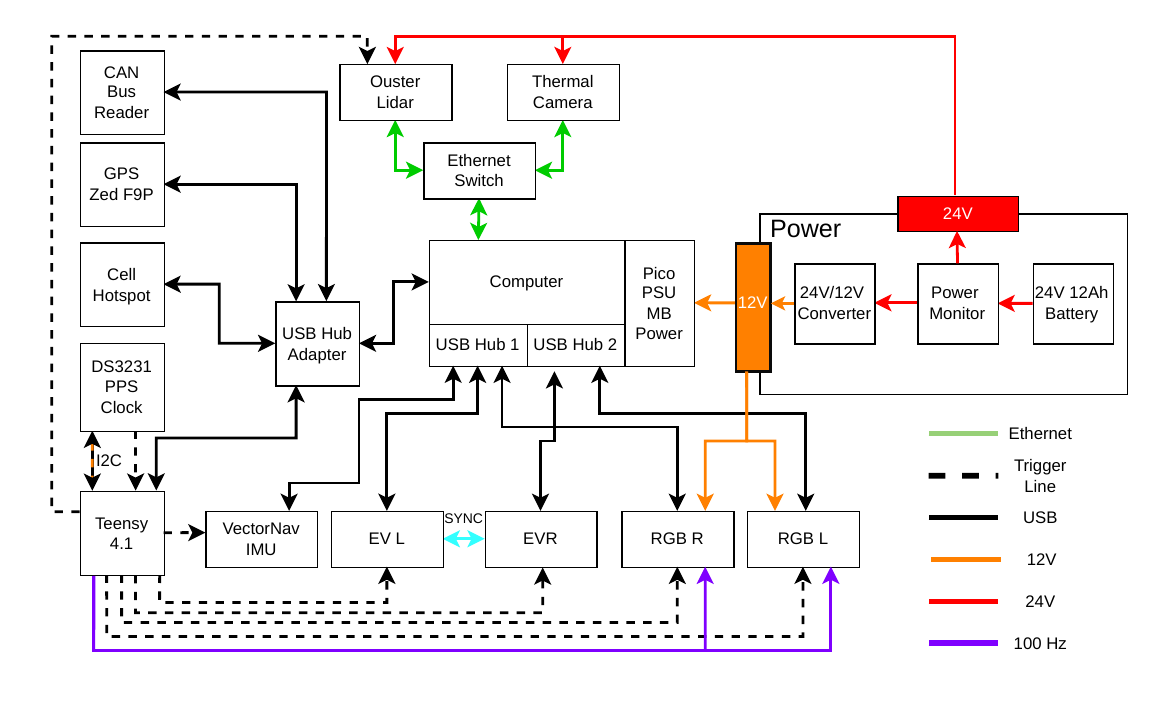}
    \caption{\textbf{OctoSense system diagram}: OctoSense integrates eight sensors (stereo RGB and event cameras, a thermal camera,  IMU, LiDAR, GPS, and platform proprioception), each often requiring a different combination of power, ground, trigger, time-sync, and data cabling. This results in over 30 cables across the entire system. The power system is fed by a 24 V battery that directly powers the thermal camera and LiDAR; a DC-DC converter steps this down to 12 V for the PicoPSU motherboard supply and the cameras. Trigger signals originate from a Teensy 4.1 via our custom PCB (\cref{fig:sync_pcb}) and a temperature compensated oscillator~(DS3231). Sensor data is transferred to the host computer over Ethernet or USB. }
    \label{fig:sys_diag}
\end{figure}

\begin{figure}
    \centering
    \includegraphics[width=\linewidth]{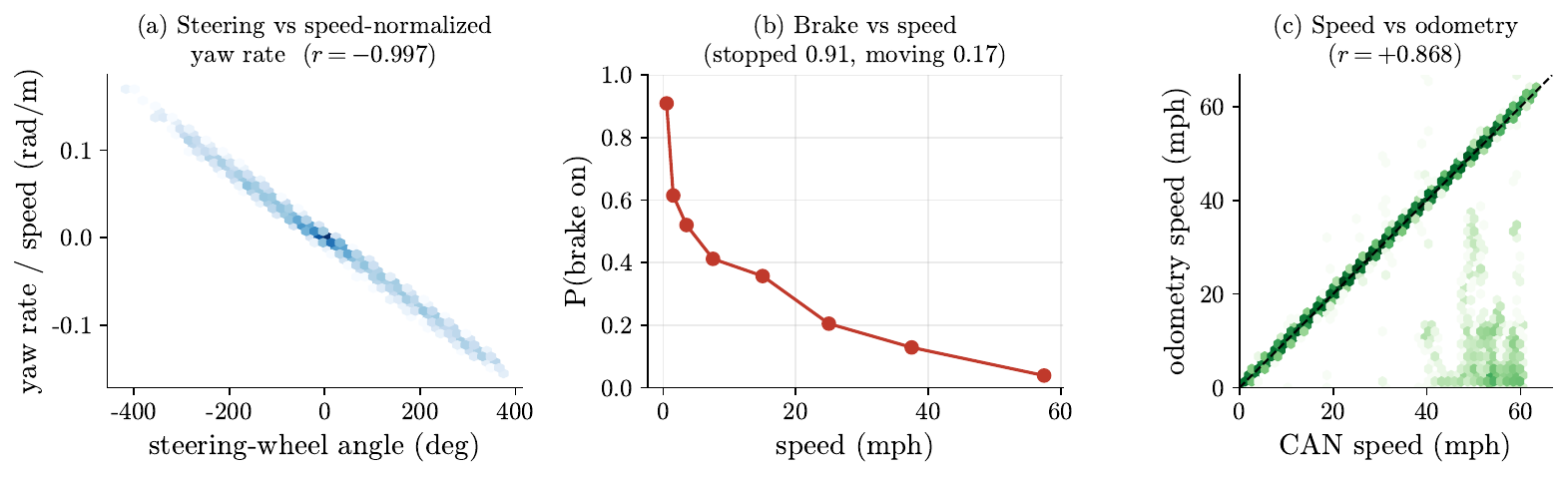}
    \caption{\textbf{Ground-truth consistency between CAN bus and LiDAR-inertial odometry} (a) Steering angle vs.\ odometry yaw-rate normalized by speed; 2D hexagonal
histogram follows the bicycle model~($r = -0.997$). (b) The brake-on signal is asserted 91\% of the time while stopped ($<$1~mph) and only 18\% while moving ($>$5~mph). (c) CAN speed vs.\ odometry-derived velocity; 2D hexagonal histogram~($r = 0.868$). The small number of histogram cells with low odometry speed comes from poor initial egomotion estimates from RKO-LIO.}
    \label{fig:can_gt}
\end{figure}

Each sensor in OctoSense requires configuration of various parameters and settings, which we discuss below.
\begin{itemize}
\item \textbf{RGB cameras} are set to a 100 Hz frame rate for capturing rapid motion and use an auto-exposure policy that targets a mean frame brightness of 60/255, first by increasing exposure up to 8 ms, and then by adjusting sensor gain. The code for this policy can be found \href{https://github.com/ros-drivers/flir_camera_driver/blob/humble-devel/spinnaker_synchronized_camera_driver/src/master_exposure_controller.cpp}{here}.

\item \textbf{Event camera} biases are fixed to tuned values across all data collection to ensure consistent event generation. Active IR from the LiDAR interferes with events caused by ambient intensity, so we added a 390–750 nm bandpass filter to each event camera to block the LiDAR's IR emission.  Our initial winter collections produced unusable recordings: temperatures below the SilkyEV's rated 0--50° C range corrupted event timestamps. We resolved this by wrapping each camera in 1" rubber foam insulation to isolate it from ambient conditions.

\item \textbf{Thermal camera} is set to a 50 Hz frame rate in high gain mode for outdoor environments, with digital detail enhancement to boost edge sharpness. The auto-exposure controller is set to \href{https://cdn.graftek.com/system/files/11822/original/83213-0102_Genicam_ICD_FLIR_Ax5.pdf}{Plateau Histogram} mode, which maps the 14-bit thermal data to 8 bits by clipping over-represented intensity bins before histogram equalization so that dominant background regions don't consume the available gray levels.

\item \textbf{LiDAR} runs at 10 Hz and outputs a range image, near infrared image, signal intensity image, reflectivity image, and its internal IMU signal in \texttt{RNG19\_RFL8\_SIG16\_NIR16} mode. We observed packet loss under high vibration of the LiDAR's data cable, which we mitigated by mounting the cable to a rigid platform and zip-tying it down onto rubber foam to dampen vibrations at the connector inlet.

\item \textbf{IMU}  is configured to run at 400 Hz and measures accelerations, angular velocities, magnetic fields, and barometric pressure over a UART interface at 921,600 baud. Vehicle vibration injects high-frequency noise into the IMU readings, degrading inertial estimates; we mitigated this by soft-mounting the IMU on rubber-damped standoffs.

\item \textbf{GPS} is configured to provide output at 5 Hz using a Waveshare patch antenna. During initial testing, mounting the antenna on the platform itself produced insufficient satellite lock due to the limited ground plane. We resolved this by magnetically mounting the antenna directly to the vehicle roof. RTK corrections are delivered via NTRIP, using \href{https://cors.dot.ny.gov/sbc/Account/Index?returnUrl=%2Fsbc}{NYSNet} for New York collections and \href{https://rtkdata.com/}{RTKData} for Pennsylvania. Cellular dongles initially used for these corrections drew enough power from the USB bus to cause bus crashes that corrupted sensor data across the entire platform. We resolved this by switching to a cellular hotspot with an internal battery and a USB-to-Ethernet interface, isolating the cellular power draw from the main USB bus. Note, RTK corrections are provided when available; limited cellular service or GPS lock can limit the data where corrections are present.

\item \textbf{CAN Bus:} OBD ports are commonly firewalled in modern vehicles, restricting CAN bus access. We therefore tap the bus directly from the \href{https://mx5things.blog/2017/01/26/hacking-mazda-forward-sensing-camera-fsc/}{vehicle's forward-sensing camera connector} via a custom splitter harness and feed signals into a CAN-bus~(DSD-C31G) adapter that exposes the bus as a Linux network socket. Each CAN message consists of a unique identifier and a data payload encoding the corresponding signal. Messages are recorded via \texttt{ROS2 SocketCAN} and decoded using the OpenDBC message definitions for the Mazda CX-5 (2017 release). Although our vehicle is a 2021 model, we confirmed the decoded signals are physically consistent with our LiDAR-inertial odometry: CAN speed matches the odometry velocity, steering angle tracks the odometry yaw rate via the kinematic bicycle model, and the brake signal is engaged almost exclusively at low speed (\cref{fig:can_gt}).

\item \textbf{Compute:} AMD Ryzen 9700X 8-core processor, 64GB DDR5 RAM, PCIe 5.0 Samsung 9100 NVMe SSD for high-throughput data acquisition.
\end{itemize}
\subsection{Lessons learned}

One of the most important lessons came early in our testing phase: many of the original electronics we tried to use failed to function properly in cold weather. Notably, we tried three different power supplies in our attempts to get the motherboard functioning: 1) \href{https://www.amazon.com/RGEEK-Switch-24pin-Supply-Computer/dp/B071P3HMNK/ref=sr_1_1?crid=3AS3MREKCEU0L&dib=eyJ2IjoiMSJ9.og2Zkw2vVMK1EPI2FSMQwJ-hlhxl8WoWcY7w6c7TvoYtPXxX5DektkeKEGz7H7TphqOvw2Ec-xMaq1tWh5IFzkrLbJqVf4D41HK6oH6S3w50PfMcbHX-JdMVcsf1EHFSE3IB6zcz6mNjHBom-G1f5eQKv2tVEH5lztxV_87KEmRZMQosEIkokqrrgdy3flaNpH7po0h7kFn4H044we9W_PKh1pkKlwlObaY3sKokC-k.v4j4Ss3TwGGfGlsUPHrwG0uu33LdyYaXV7mK8mVzPas&dib_tag=se&keywords=RGEEK+1106+DC+to+DC+ATX+PSU+12V+300W+Pico+ATX+Switch+PSU+24pin+M&nsdOptOutParam=true&qid=1778176990&sprefix=rgeek+1106+dc+to+dc+atx+psu+12v+300w+pico+atx+switch+psu+24pin+m%2Caps%2C132&sr=8-1}{RGEEK 1106 DC to DC ATX PSU 12V 300W Pico ATX Switch PSU 24pin M}, 2) \href{https://www.amazon.com/DC-ATX-Supply-Module-Switch-PS002d/dp/B0CHS14Q6B/ref=sr_1_2?crid=2AWVSV9K1WNEW&dib=eyJ2IjoiMSJ9.Bpmvrhspw3bh42KhHHr4gkm66XX3KXw0REnLx6rE6O1uhC4akcYBF9HktmRycNd707SqQEfUmCbqUD_eLcuuPoheOYvJ1hlhOX4OFvVOU4PnF7c8zEDIdRSHhnzPQPEdcUdJt2DfULND82PfN-xCLiXSCnGyntFEOn8rd_TRr6OuMEkrVAoiEbqoQiX6VTmnZbCnyElYIfvgkmXC9u-CSfkxd6VALoCZ5iAUORQ_wEA.3s-oUk2ZJfMYTeRGzS2-XmWoe8F2HeYrHfzxKENeQcM&dib_tag=se&keywords=picopsu&qid=1778177026&sprefix=picopsu%2Caps%2C139&sr=8-2}{DC 12V 300W 24Pin ATX Connect with Motherboard}, and 3) \href{https://www.amazon.com/Mini-Box-picoPSU-160-XT-Power-Mini-ITX-Supply/dp/B005TWE6B8/ref=sr_1_1_pp?crid=2AWVSV9K1WNEW&dib=eyJ2IjoiMSJ9.Bpmvrhspw3bh42KhHHr4gkm66XX3KXw0REnLx6rE6O1uhC4akcYBF9HktmRycNd707SqQEfUmCbqUD_eLcuuPoheOYvJ1hlhOX4OFvVOU4PnF7c8zEDIdRSHhnzPQPEdcUdJt2DfULND82PfN-xCLiXSCnGyntFEOn8rd_TRr6OuMEkrVAoiEbqoQiX6VTmnZbCnyElYIfvgkmXC9u-CSfkxd6VALoCZ5iAUORQ_wEA.3s-oUk2ZJfMYTeRGzS2-XmWoe8F2HeYrHfzxKENeQcM&dib_tag=se&keywords=picopsu&qid=1778177026&sprefix=picopsu%2Caps%2C139&sr=8-1}{Mini-Box picoPSU-160-XT High Power}.
The first two power supplies operated for about fifteen minutes before failing during cold weather operation. We traced the issue to their electrolytic capacitors, whose equivalent series resistance increases at low temperatures, reducing the supplied voltage and causing the motherboard to shut off. Our third option, the Mini-Box picoPSU, uses polymer capacitors instead, eliminating this problem and enabling stable operation.

The second component that gave us a lot of trouble was the 24V-to-12V converter used to power the picoPSU and cameras. We tried: 1) \href{https://www.amazon.com/Nilight-Converter-Regulator-Waterproof-Transformer/dp/B0CYQ8X29T/ref=sr_1_3?crid=2XJVY189I0LRU&dib=eyJ2IjoiMSJ9.NHCw_sJwFXSXkoc8fuwNUz9A9PpXZ_yPBbDK6EukU-7EnLkJuEeA4VzEGeZFFBMfYI0eXqPXFNujx2qm_eJi2r2NF9evjQtgyyxCR_7lnDQLkGpx7YQ_CRnMS0XbuTexMxLMOZefEkNL3hfXr78TbKg5NSrHUdMAGWUAV_haSkpN95iZJuAOF0G5EzQNPm3QNk8rL6-0uoB0JIIH5N4yRvtNAvsgNBveaeMa_xNdA-U.3JAgtAPYjvV3puv1bu-G_hFanq0wv_lHp4BQBp7EC-w&dib_tag=se&keywords=DC%2BTransformer%3A%2BNilight%2B24V%2Bto%2B12V%2BVoltage%2BConverter%2B240W%2B20A%2BRegulator%2BStep%2BDown%2Bto%2B12VDC&nsdOptOutParam=true&qid=1778177511&sprefix=dc%2Btransformer%2Bnilight%2B24v%2Bto%2B12v%2Bvoltage%2Bconverter%2B240w%2B20a%2Bregulator%2Bstep%2Bdown%2Bto%2B12vdc%2Caps%2C185&sr=8-3&th=1}{DC Transformer: Nilight 24V to 12V Voltage Converter 240W 20A Regulator Step Down to 12VDC} and 2) \href{https://www.amazon.com/dp/B01M0HTT8D?ref=ppx_yo2ov_dt_b_fed_asin_title&th=1}{Victron Energy Orion-Tr DC to DC Converter - 24/12-Volt}. The first option would fail after thirty minutes of cold weather operation, and thus we switched to the second option that supported our full mission lifetime.

Finally, another component that caused us trouble was the \href{https://www.amazon.com/waveshare-SIM7600G-H-DONGLE-Adapter-Communication/dp/B08CSB596W/ref=sr_1_2?crid=3TFG7JB760LSW&dib=eyJ2IjoiMSJ9.L0AApUpVdJrYY_pzYm5V75jaxyabqN1H_HP7vZxldc-nbnRTft25u1enDGyclEqv61Fke7fBSNyLAdDnN1hF6P60tvu8bMckFT_-n9Y_WqlkW5uT2upKEcSRMmGjALjXjTT8EM8JZ1ISgyBHGFaGAR3UJO8_UT8Vya3B5mQJjAzvzFqMOlHGW1b-PxoYFpz5VvNnj85x7GZqYO7fDEYsiEuiZpHU4Xx_rM6UDNed8w0.pWGh8P2c3XTKYasvK8N7fKWhnVcAKnqDWW3HnLWlujc&dib_tag=se&keywords=IM7600G-H+4G+DONGLE&qid=1778177927&sprefix=im7600g-h+4g+dongle+%2Caps%2C110&sr=8-2}{waveshare SIM7600G-H 4G DONGLE LTE USB} used to receive NTRIP correction data. When plugged in, it would overdraw current from the USB bus, causing a brownout that crashed connected USB devices, halted their drivers, and corrupted our data collection sequence. Our first attempt at a solution was a powered USB hub \href{https://www.amazon.com/dp/B07G7GP15C?ref=ppx_yo2ov_dt_b_fed_asin_title}{Coolgear 3.2 Gen 1 USB Hub} but still this too proved unstable. We then switched to a \href{https://www.amazon.com/dp/B079FNC379?ref=ppx_yo2ov_dt_b_fed_asin_title&th=1}{KuWFi 4G LTE Unlocked Wi-Fi Hotspot Device}, which has an internal battery and exposes a network interface to the host over USB. This eliminated the power draw issue entirely, since the device charges steadily at standard USB current rather than drawing peak cellular-radio current from the bus. A recurring challenge with many of these components was the lack of complete specification sheets, which meant we had to discover most of these failure modes through experimentation. We document them here for the benefit of future users of such hardware.

Another challenge surfaced during winter testing of our time synchronization system. After collecting over five hours of data, we ran temporal synchronization and discovered that most of the timestamps from our event camera were corrupted. Some background on this issue: the SilkyEV uses the \href{https://docs.prophesee.ai/stable/data/encoding_formats/evt3.html}{EVT3.0 encoding}, which represents each microsecond timestamp as a 24-bit value split into a 12-bit time-high field (the 12 most significant bits) and a 12-bit time-low field (the 12 least significant bits). Each time-high value represents a 4.096 ms step, while time-low ticks in 1 $\mu$s increments within that step, giving a full time-high cycle of 16.78 s before it wraps back to zero. The decoder tracks the timestamp by combining the most recent time-high with the incoming time-low values, and when time-high wraps, the decoder must detect the overflow and maintain a higher-order counter to keep timestamps monotonic. In our corrupted data, we observed time-high overflowing multiple times within a 20-second window, far more frequent than the one overflow per 16.78 s that the format specifies. To cross-validate, we asked a collaborator to collect data in a room-temperature environment and then place the camera in a freezer for 15 minutes before collecting data again. They reproduced our failure mode exactly, confirming that our issue was induced by cold. We first tried active warming with a \href{https://www.amazon.com/dp/B0CM6DTTKP?ref=ppx_yo2ov_dt_b_fed_asin_title&th=1}{ Film Heater Plate Adhesive Pad 12V 7W}, but this solution did not provide enough heat to keep the cameras in their operating range. We then switched to a passive solution, wrapping the cameras in 1" rubber foam insulation, which provided sufficient thermal shielding.

\begin{figure}[h]
    \centering
    \includegraphics[width=\linewidth]{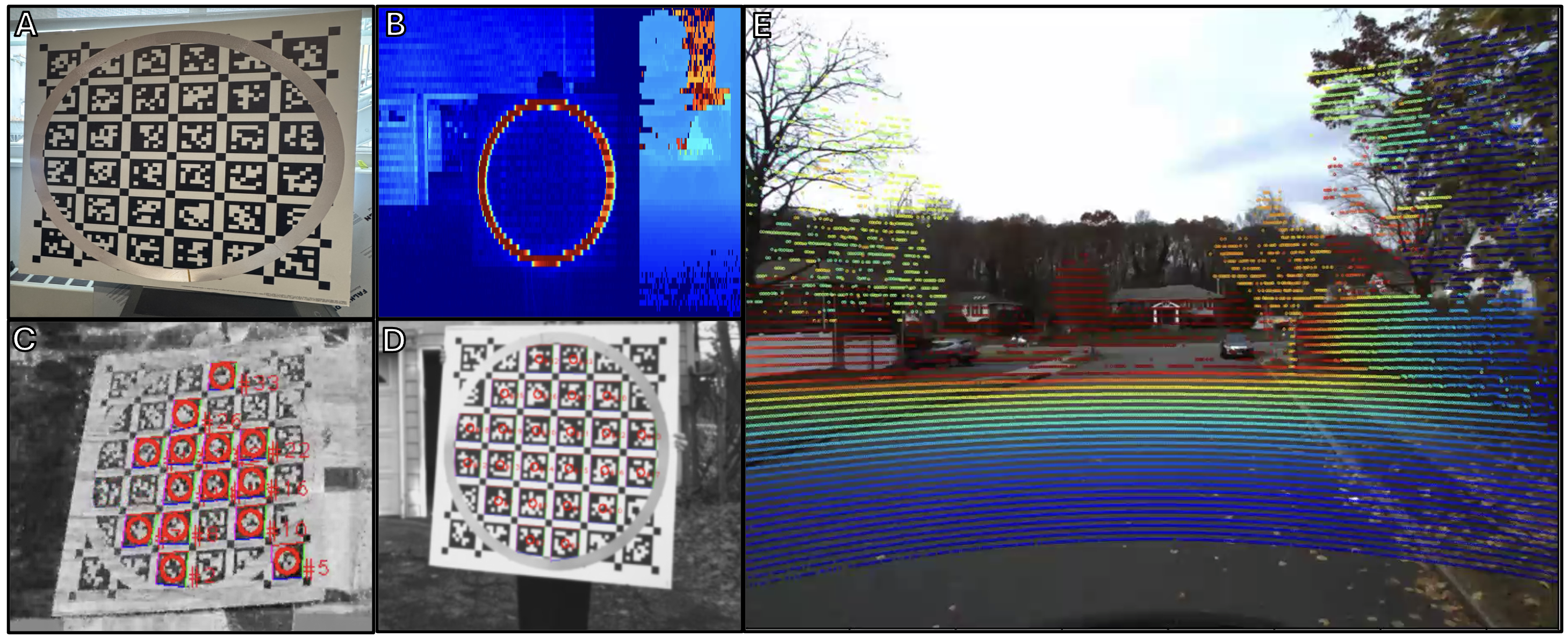}
    \caption{\textbf{Calibration techniques used in OctoSense.} We detect unique features from our calibration target across modalities to solve for sensor extrinsics. (A) A 6×6 AprilGrid with 11.2 cm marker size augmented with a retro-reflective circle. (B) The retro-reflective circle enables easy detection via simple thresholding on the LiDAR reflectivity image. (C) Reconstructed event images using the IIR filters of \cite{pfrommer2022frequencycam}, with corners detected by an AprilTag detector. (D) RGB images with detected AprilTag points. We use all of these correspondences to solve for the sensor extrinsics. (E) Reprojection of the LiDAR depth into the RGB image, showing tight spatial alignment and confirming a good calibration.}
    \label{fig:cal}
\end{figure}

\subsection{Mechanical design of OctoSense}
\begin{figure}
    \centering
    \begin{subfigure}{0.4\linewidth}
        \centering
        \includegraphics[width=\linewidth]{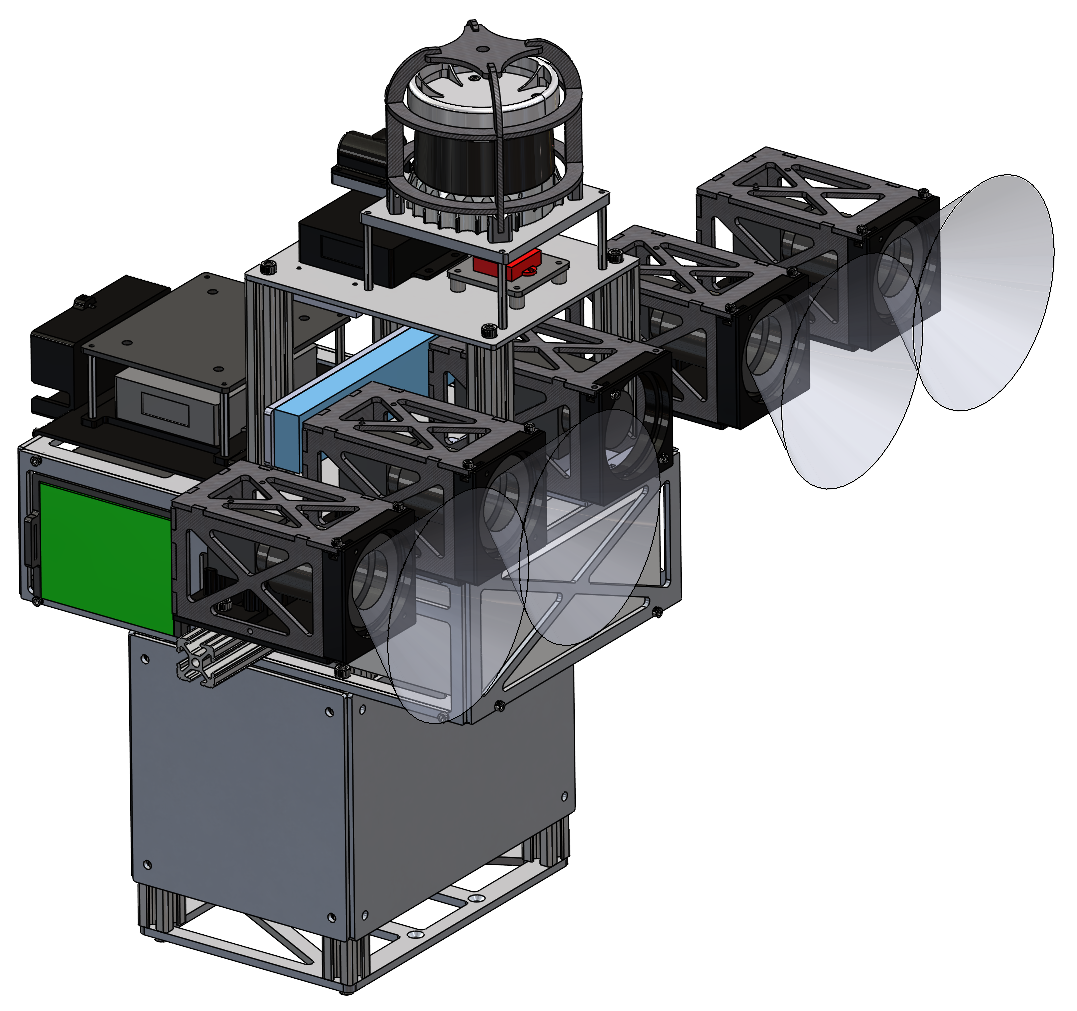}
    \end{subfigure}
    \hspace*{1em}
    \begin{subfigure}{0.49\linewidth}
        \centering
        \includegraphics[width=\linewidth]{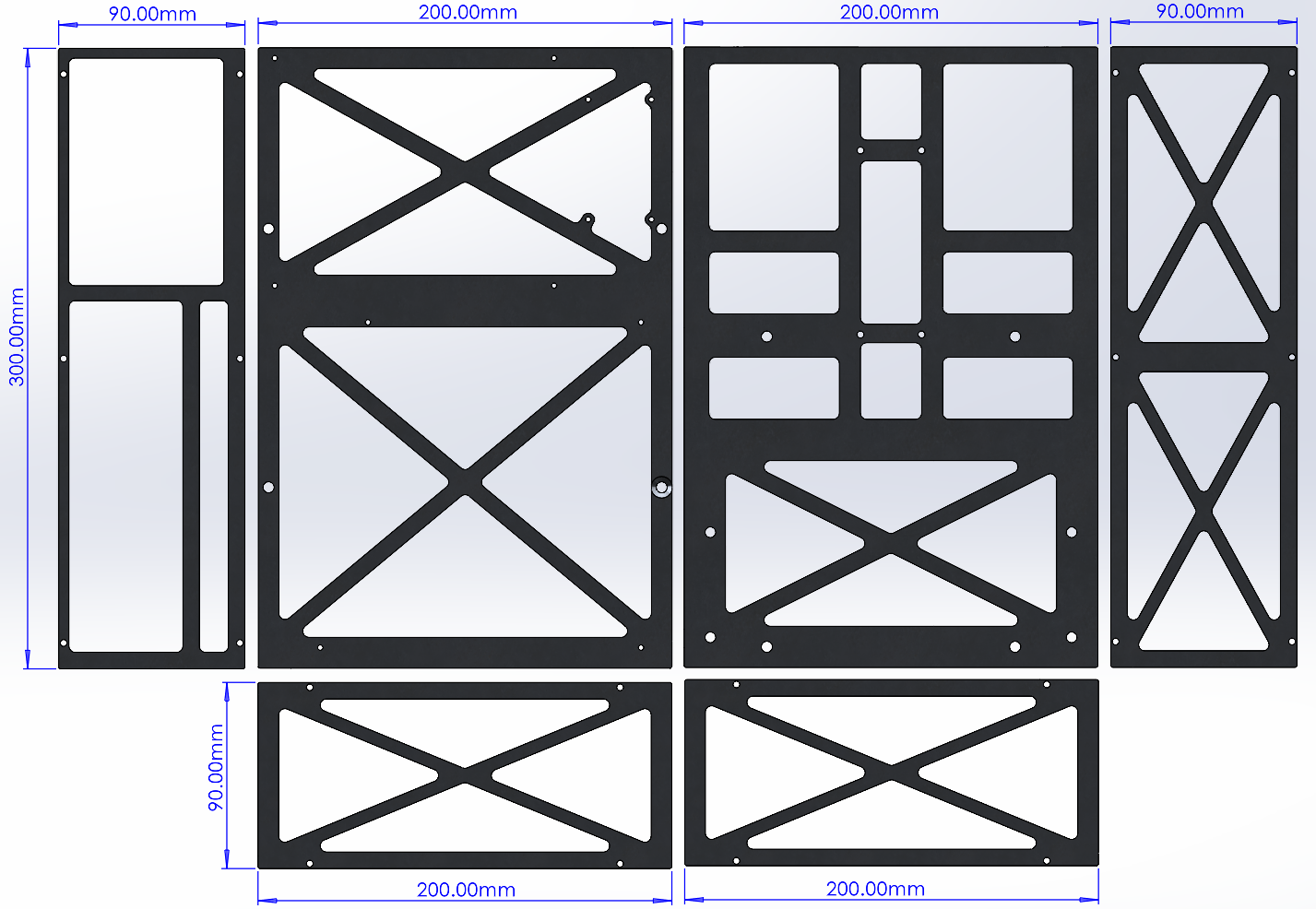}
    \end{subfigure}
    \caption{\textbf{Left:} A CAD rendering of OctoSense with five cameras: stereo event, stereo RGB, and thermal camera mounted on an aluminum rail for configurable spatial placement, each in a carbon fiber cage for protection. A similar cage is used for LiDAR. Our system was made for modular deployment across a variety of robot platforms. \textbf{Right:} Mechanical drawings for the inner cage surrounding the motherboard.}
    \label{fig:octosense_CAD}
\end{figure}

OctoSense was designed with a primary focus on mechanical robustness and reliability for deployment across platforms operating in outdoor environments. Similar to modular robotic systems, the design emphasizes structural rigidity, modular integration, maintainability, and rapid reconfiguration across experiments. Lightweight carbon fiber enclosures were developed for both the camera assemblies and the LiDAR to maximize impact resistance while minimizing overall system weight. Aluminum structural members with weight-reducing cutouts served as the primary support structure, providing increased stiffness while limiting additional mass that could otherwise raise the center of mass and introduce vibration during operation. The camera assemblies were mounted using an aluminum extrusion, enabling straightforward sensor adjustment, replacement, and system reconfiguration across platforms when needed. Thin aluminum reinforcement plates were incorporated to reduce vibration-induced flexing caused by the elevated sensor configuration, while custom vibration dampers and mechanically secured connections with foam-supported cable bracing were included to further improve operational reliability during extended outdoor deployment.
The system was mounted to the vehicle's sunroof using \href{https://www.seasucker.com/collections/vacuum-mounts/products/6-seasucker-black}{SeaSucker} vacuum cup mounts.  

\begin{figure}
    \centering
    \includegraphics[width=\linewidth]{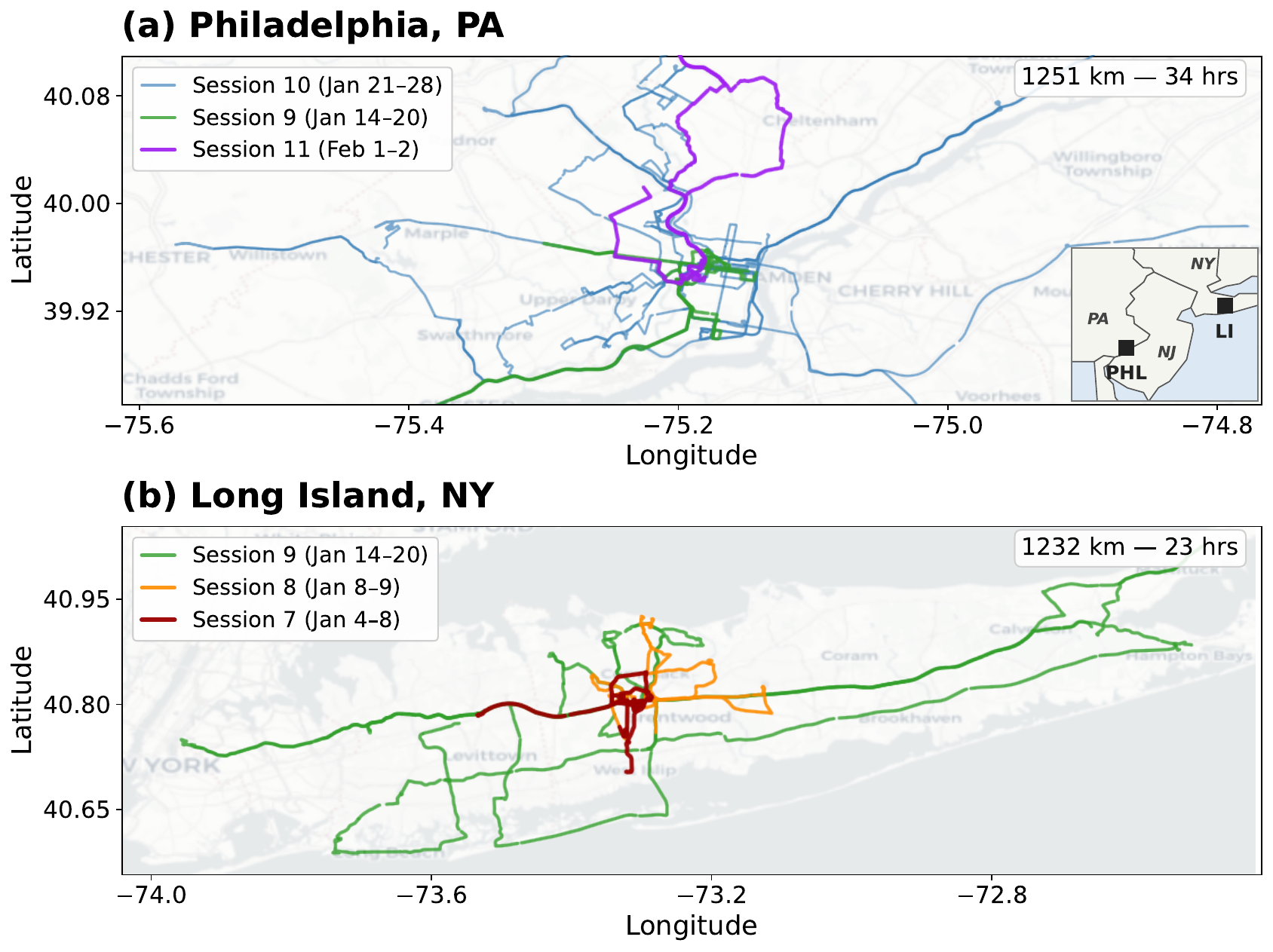}
    \caption{\textbf{OctoSense Dataset GPS Routes} Routes for each sequence on Long Island and in Philadelphia. These cover only intervals with valid GPS lock and therefore total less than the full 59 hours of recorded data. }
    \label{fig:gps_routes}
\end{figure}


\section{Software Improvements}
\label{sec:software}

Throughout this investigation we have used many open-source drivers and utilities. While using these tools, some have needed improvements or configuration changes to solve various bugs or enable new features. We list below the various repositories we either created or modified:

\medskip \noindent \faGithub~\textbf{\texttt{ouster\_http\_node}}\\
\href{https://github.com/anthonytec2/ouster_http_node}{\nolinkurl{github.com/anthonytec2/ouster_http_node}} \\
\ul{Bug:} For time synchronization across our sensors, we need to know when the LiDAR has achieved hardware synchronization lock from our timing pulse. This lets us uniquely identify the first valid pulse and align the LiDAR with the rest of the sensors. \\
\ul{Fix:} We built a ROS2 node that wraps the Ouster HTTP API, polls the sensor for its time-synchronization status, and republishes it as a ROS2 topic so it can be recorded into a rosbag alongside the sensor data.

\medskip \noindent \faGithub~\textbf{\texttt{rko\_lio}} \\
\href{https://github.com/anthonytec2/rko_lio/commits/master/}{\nolinkurl{github.com/anthonytec2/rko_lio}} \\
\ul{Bug:} Upstream RKO-LIO publishes only a downsampled version of the deskewed point cloud and discards per-point channels (reflectivity, signal, near-infrared).\\
\ul{Fix:} We retain the full-resolution scan, apply deskewing to it directly, and preserve all original per-point channels in the published cloud.

\medskip \noindent \faGithub~\textbf{\texttt{kalibr}} \\
\href{https://github.com/anthonytec2/kalibr/commit/d2b8369995b485b275c0c83056a4c832a3c9e778}{\nolinkurl{github.com/anthonytec2/kalibr}} \\
\ul{Bug:} Kalibr's AprilTag detector assumes a black border width of 2 by default, which prevented the tags used in this investigation (border width 1) from being detected.  \\
\ul{Fix:} Changed the default black border width to 1, enabling reliable detection of our AprilTags.

\medskip \noindent \faGithub~\textbf{\texttt{flir\_camera\_driver}} (branch: \texttt{line\_status})  \\ 
\href{https://github.com/anthonytec2/flir_camera_driver/commits/line_status}{\nolinkurl{github.com/anthonytec2/flir_camera_driver}} \\
\ul{Missing feature:} For time synchronization of our FLIR cameras, we need to know during which frame the pulse-per-second (PPS) signal is active. The FLIR cameras expose a GPIO line status register that reports the level of an input pin, to which we route the PPS signal. \\
\ul{Fix:} Exposed the line status as part of each frame's published metadata, allowing the PPS state to be recorded alongside the image data.

\medskip \noindent \faGithub~\textbf{\texttt{ffmpeg\_image\_transport\_tools}} (branch: \texttt{transcoder\_pts}) \\  \href{https://github.com/anthonytec2/ffmpeg_image_transport_tools/commits/transcoder_pts/}{\nolinkurl{github.com/anthonytec2/ffmpeg_image_transport_tools}} \\
\ul{Missing feature:} Our recorded rosbags contain hardware-encoded FFmpeg image streams, but the mapping between recorded
 frames and their decoded timestamps was broken. After transcoding a bag to a video file, there was no reliable way to tell
which output frame corresponded to which original capture time, making downstream time synchronization impossible.\\
\ul{Fix:} Added a new \texttt{bag\_to\_video} executable that decodes \texttt{ffmpeg\_image\_transport} messages from a bag and re-encodes them via libav, tracking each frame's presentation timestamp (PTS) through the decode/encode pipeline and emitting aligned timestamp sidecars that map every encoded frame back to its original bag capture time. The tool exposes  a zero-copy raw \texttt{AVFrame} path between decoder and encoder to skip cv\_bridge conversion.

\medskip \noindent  \faGithub~\textbf{\texttt{ffmpeg\_encoder\_decoder}} (branch: \texttt{pts\_arr}) \\
\href{https://github.com/anthonytec2/ffmpeg_encoder_decoder/commit/3f3d9cf1982835cd26c238d07c7f5dd467d92a9f}{\nolinkurl{github.com/anthonytec2/ffmpeg_encoder_decoder}} \\
\ul{Missing feature:} The decoder and encoder APIs only operated on ROS \texttt{sensor\_msgs/Image} messages through
cv\_bridge, which forced a pixel-format conversion and full-frame copy on every frame, even when the decoder output already
matched the encoder input, and provided no way to tell which PTS had actually made it through a
decode pass, so the bag-time $\leftrightarrow$ PTS mapping could not be recovered on the other side of a transcode.\\
\ul{Fix:}
Added a raw \texttt{AVFrame} pathway on both sides of the library. The decoder's new \texttt{initializeRaw()} delivers
decoded \texttt{AVFrame}s directly to the caller, and a companion \texttt{getAndClearDecodedPTS()} accessor returns the list
of PTS values that were successfully decoded since the last call, making the input$\leftrightarrow$output frame mapping
explicit. The encoder's matching \texttt{encodeAVFrame()} accepts a raw \texttt{AVFrame}, uses \texttt{av\_image\_copy} when
source and destination pixel formats agree, and otherwise caches an \texttt{sws\_scale} context so the conversion is set up
only once per stream.

\medskip \noindent  \faGithub~\textbf{\texttt{simple\_image\_recon\_lib}} (branch: \texttt{python\_bindings}) \\
\href{https://github.com/anthonytec2/simple_image_recon_lib/tree/python_bindings}{\nolinkurl{github.com/anthonytec2/simple_image_recon_lib}} \\
\ul{Missing feature:} For camera calibration, we need grayscale-like frames reconstructed from event-camera data in order to detect AprilTags. The upstream reconstruction code is implemented in C++, but the rest of our processing pipeline is in Python.\\
\ul{Fix:} We added Python bindings to the C++ reconstruction library, allowing batches of events to be passed in and reconstructed frames returned directly for use in camera calibration.

\section{Synchronizing different sensor clocks}
\label{app:kalman_filt}

We model the map from each sensor's clock to the main clock as a linear system with state, say, $x = [b, a]^\top$, where $b$ is the clock offset (bias) and $a$ is the clock skew (relative tick rate).  Suppose that this state evolves as a random walk and is observed through the affine clock map:
\begin{align}
    x_k &= x_{k-1} + w_k\\
    z_k &= H_k x_k + v_k \text{ with } H_k = [1,\ t^s_k]
\end{align}
where $w_k \sim N(0, Q \Delta t_k)$ and $v_k \sim N(0, R)$ are Gaussian noise in the dynamics and observation models respectively. The quantity $z_k$ is the main-clock pulse timestamp and $t_k^s$ is the device clock timestamp corresponding to the pulse per second (PPS) event of $z_k$. We set $Q = \text{diag}(\sigma_b^2,\ \sigma_a^2)$ and therefore $Q \Delta t_k$ is the variance accrued over the interval $\Delta t_k$. Initial state is bootstrapped from the first pulse as $a_0 = 1$, $b_0 = z_0 - t^s_0$. A similar estimation framework can be seen in \cite{Hamilton2008ACESAC}.

\begin{table}[!htpb]
\centering
\caption{Per-sensor Kalman filter noise parameters. $R$ is the measurement variance, calibrated to each sensor's edge-detection jitter. $Q = \text{diag}(\sigma_b^2, \sigma_a^2)$ is the per-second process noise on offset and skew. The initial covariance of the state is $P_0 = \text{diag}(1.0,\ 10^{-6})$.}
\vspace*{1ex}
\label{tab:kf-params}
\small
\begin{tabular}{lrrrl}
\toprule
Sensor & $\sqrt{R}$ (ms) & $\sigma_b^2$ & $\sigma_a^2$ & Rationale \\
\midrule
IMU            & 1    & $10^{-8}$  & $10^{-10}$ & 400\,Hz edge quantization\\
FLIR cameras   & 5    & $10^{-10}$ & $10^{-13}$ & 100\,Hz  edge quantization  \\
Event cameras  & 0.01 & $10^{-10}$ & $10^{-13}$ & $\mu$s-resolution trigger events \\
System clock   & 25   & $10^{-6}$  & $10^{-9}$  & OS scheduling jitter \\
\bottomrule
\end{tabular}
\end{table}

\begin{figure}[!htpb]
    \centering
    \begin{subfigure}[c]{0.65\linewidth}
        \centering
        \includegraphics[width=\linewidth]{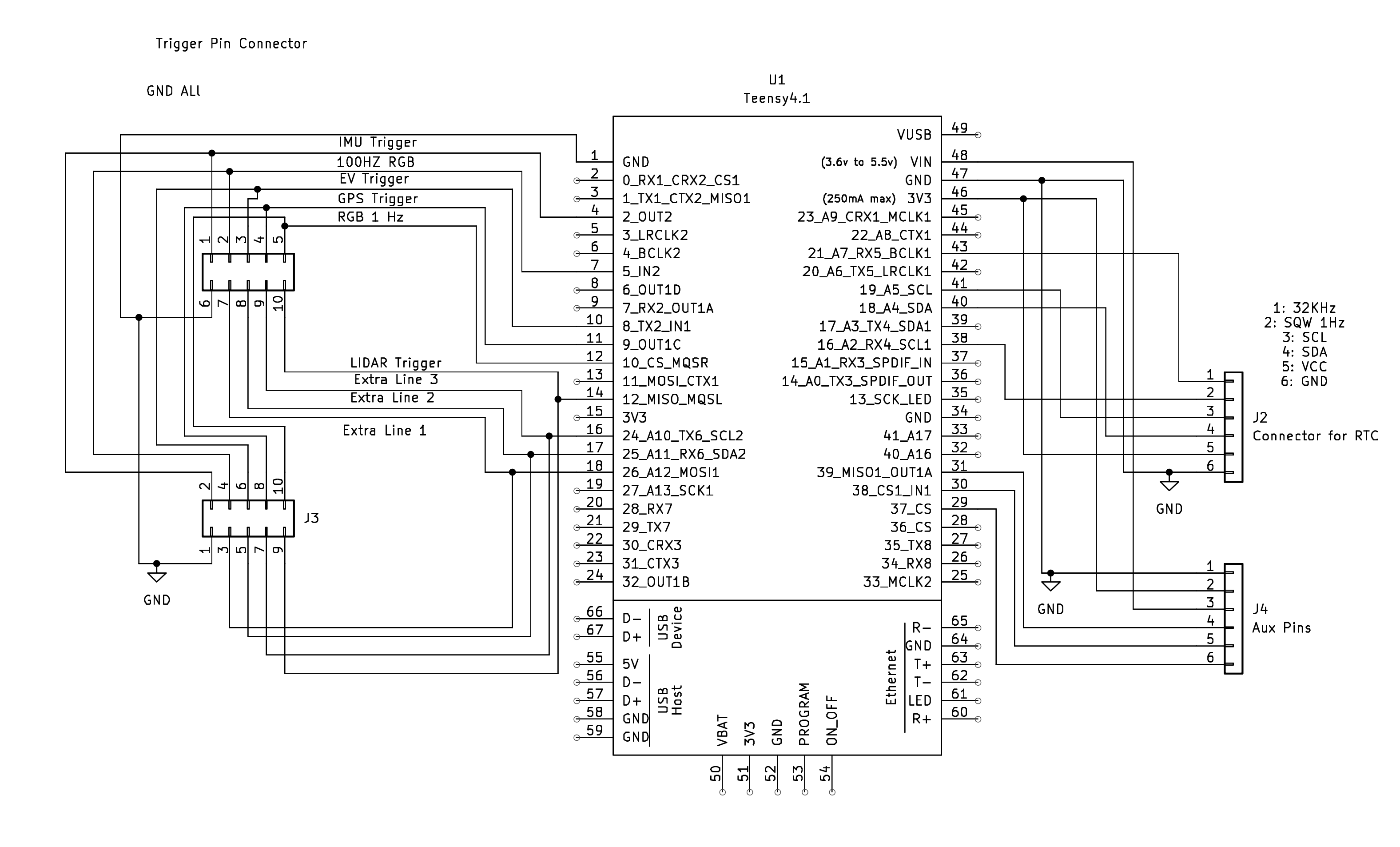}
        \label{fig:sync_schematic}
    \end{subfigure}
    \hfill
    \begin{subfigure}[c]{0.34\linewidth}
        \centering
        \includegraphics[width=\linewidth]{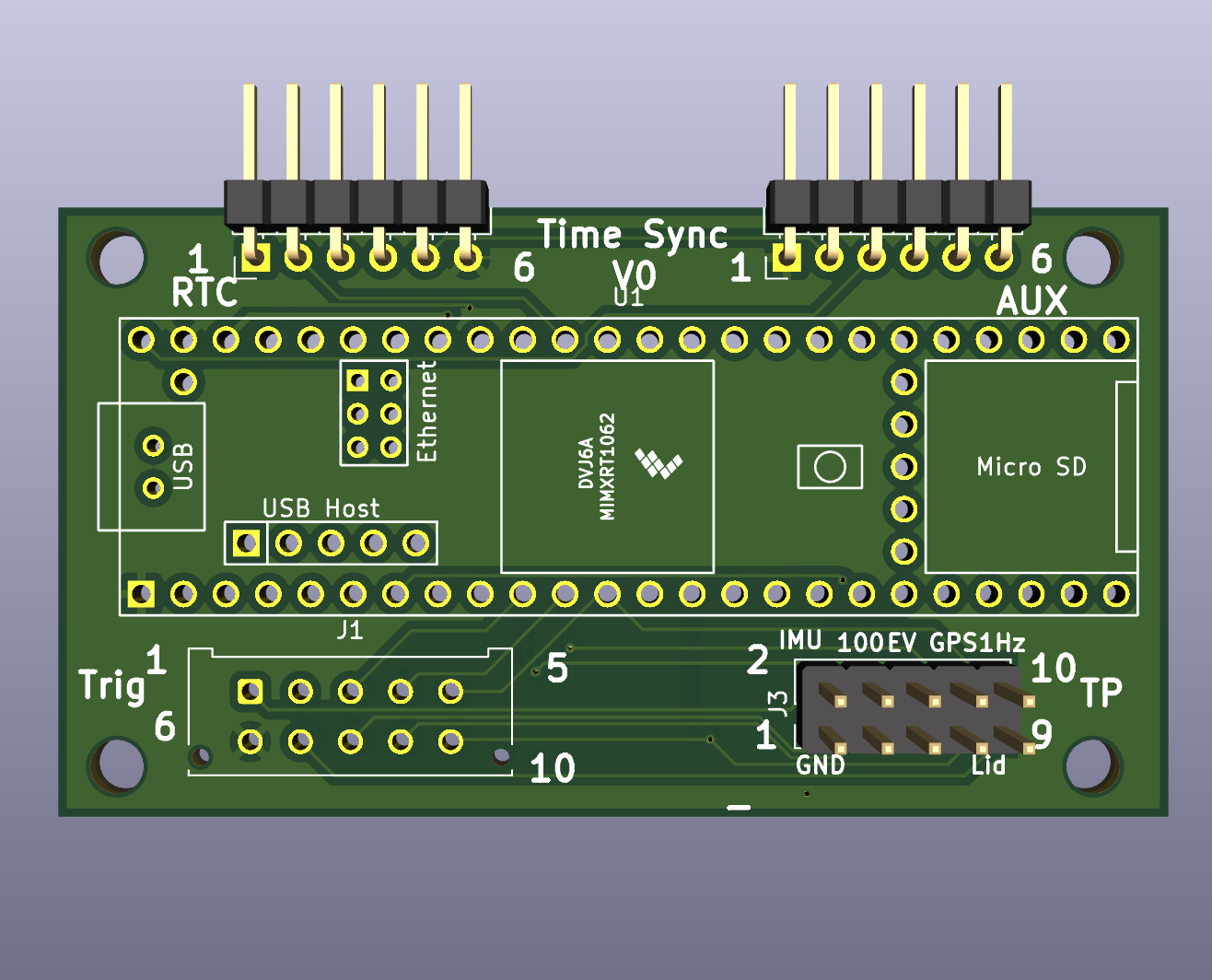}
        \label{fig:sync_pcb}
    \end{subfigure}
    \caption{\textbf{Custom time synchronization board}: We design a custom PCB integrating a Teensy 4.1 microcontroller, a DS3231 temperature-compensated crystal oscillator, and a header exposing all trigger output signals, providing a unified platform for distributing sensor triggers. We chose the Teensy for its 600 MHz clock, which allows precise, low-jitter trigger generation and the DS3231 for its low 2 ppm drift. \textbf{Left:} Schematic for our integration of these parts. \textbf{Right:} Rendering of the PCB created in KiCad from this schematic.}
    \label{fig:sync_board}
\end{figure}

\begin{figure}[!htpb]
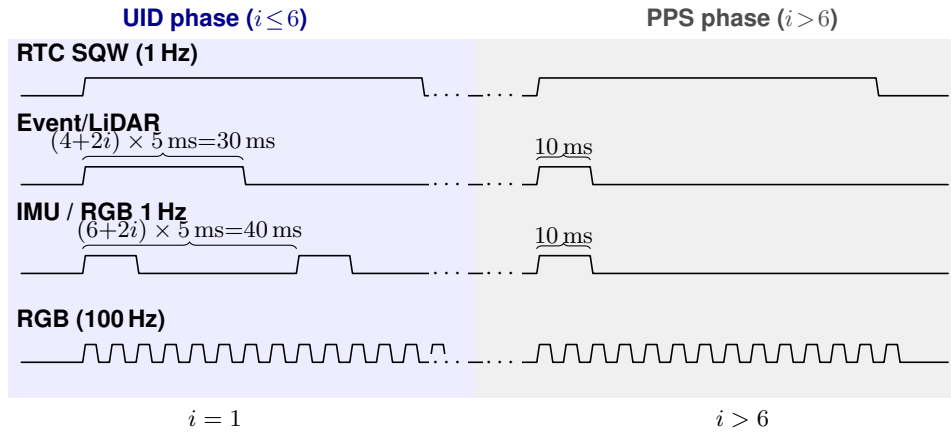

  \centering
  \begin{tikztimingtable}[
      xscale=1.2,
      yscale=0.80,
      semithick,
      timing/rowdist=5,
      timing/name/.style={inner sep=0pt, outer sep=0pt},
  ]
    & 2L 0.3L 12.7H 2L  2L 0.3L 12.7H 2.7L                               \\
    & 2L 0.3L 6H 8.7L  2L 0.3L 2H 10.7L 2.7L                             \\
    & 2L 0.3L 2H 6L 2H 4.7L  2L 0.3L 2H 10.7L 2.7L                       \\
    & 2.3L 14{0.5H 0.5L} 0.7L  2.3L 14{0.5H 0.5L} 1.4L                   \\
    \extracode
      \begin{background}
        \fill[blue!7]  (-0.5, 3.2) rectangle (17, -17);
        \fill[gray!12] (17,   3.2) rectangle (35, -17);
      \end{background}
 
      \node[anchor=south west, font=\small\bfseries, inner sep=1pt]
        at (-0.3, 1.5) {RTC SQW (1\,Hz)};
 
      \node[anchor=south west, font=\small\bfseries, inner sep=1pt]
        at (-0.3, -2.2) {Event/LiDAR};
      \draw[decorate, decoration={brace, amplitude=2.5pt}, thin]
        (2.3, -3.6) -- (8.3, -3.6);
      \node[anchor=south, font=\footnotesize, inner sep=1.5pt] at (5.3, -3.55)
        {$(4{+}2i) \times 5\,\mathrm{ms}{=}30\,\mathrm{ms}$};
 
      \node[anchor=south west, font=\small\bfseries, inner sep=1pt]
        at (-0.3, -7.2) {IMU / RGB 1\,Hz};
      \draw[decorate, decoration={brace, amplitude=2.5pt}, thin]
        (2.3, -8.6) -- (10.3, -8.6);
      \node[anchor=south, font=\footnotesize, inner sep=1.5pt] at (6.3, -8.55)
        {$(6{+}2i) \times 5\,\mathrm{ms}{=}40\,\mathrm{ms}$};
 
      \node[anchor=south west, font=\small\bfseries, inner sep=1pt]
        at (-0.3, -13.5) {RGB (100\,Hz)};
 
      \draw[decorate, decoration={brace, amplitude=2pt}, thin]
        (19.3, -3.6) -- (21.3, -3.6);
      \node[anchor=south, font=\footnotesize, inner sep=1.5pt] at (20.3, -3.55)
        {$10\,\mathrm{ms}$};
 
      \draw[decorate, decoration={brace, amplitude=2pt}, thin]
        (19.3, -8.6) -- (21.3, -8.6);
      \node[anchor=south, font=\footnotesize, inner sep=1.5pt] at (20.3, -8.55)
        {$10\,\mathrm{ms}$};
 
      \node[anchor=south, font=\small\bfseries, blue!55!black] at (7.25, 3.1)
        {UID phase ($i\!\le\!6$)};
      \node[anchor=south, font=\small\bfseries, black!70] at (27, 3.1)
        {PPS phase ($i\!>\!6$)};
 
      \node[anchor=north, font=\footnotesize] at (7.25, -17.2) {$i=1$};
      \node[anchor=north, font=\footnotesize] at (27,   -17.2) {$i>6$};
 
      \node[font=\footnotesize, fill=blue!7, inner sep=1pt]  at (16,  0)   {$\cdots$};
      \node[font=\footnotesize, fill=blue!7, inner sep=1pt]  at (16, -5)   {$\cdots$};
      \node[font=\footnotesize, fill=blue!7, inner sep=1pt]  at (16, -10)  {$\cdots$};
      \node[font=\footnotesize, fill=blue!7, inner sep=1pt]  at (16, -15)  {$\cdots$};
      \node[font=\footnotesize, fill=gray!12, inner sep=1pt] at (18,  0)   {$\cdots$};
      \node[font=\footnotesize, fill=gray!12, inner sep=1pt] at (18, -5)   {$\cdots$};
      \node[font=\footnotesize, fill=gray!12, inner sep=1pt] at (18, -10)  {$\cdots$};
      \node[font=\footnotesize, fill=gray!12, inner sep=1pt] at (18, -15)  {$\cdots$};
  \end{tikztimingtable}%
  \caption{%
    \textbf{Hardware time-synchronization signals.} The cycle index $i$ starts
    at 1 with the first PPS edge after \texttt{START}. All four
    signals rise simultaneously on each PPS rising edge; the 100\,Hz
    \texttt{RGB} clock is phase-locked because its hardware timer
    is reset on every PPS edge. During the UID phase
    ($i \leq 6$) variable-width pulses encode the index $i$:
    \texttt{Event} clock emits a single pulse of width
    $(4 + 2i) \times 5$ ms, and \texttt{IMU}/\texttt{RGB 1 Hz} clocks emit
    two 10 ms pulses with rising-edge of period $(6 + 2i) \times 5$ ms.
    During the PPS phase ($i > 6$), all channels emit a single
    10 ms pulse per second.
  }
  \label{fig:pps-sync}
\end{figure}

\paragraph{Implementation details}
We had to make some alterations to the above idea in our setting. Since the observation matrix $H_k$ depends upon the timestamp, the filter above as written suffers from numerical conditioning issues if we use Unix time for $t_k^s$. Therefore, we center the device clock at its first sample, $\tilde{t}^s_k = t^s_k - t^s_0$, and recover the original mapping afterward. 

In our data, we observed malformed synchronization sequences containing spurious pulses between PPS events. We reject such malformed updates by gating each measurement on its Mahalanobis
distance as follows. Suppose $\hat x_{k \mid k-1}$ and $P_{k \mid k-1}$ are the predicted mean and covariance of the estimated state at step $k$ before incorporating the observation $z_k$. Suppose $y_k = z_k - H_k \hat x_{k \mid k-1}$ is the innovation, and $S_k = H_k P_{k \mid k-1} H_k^\top + R$ is its covariance. If $d_k = |y_k| / \sqrt{S_k}$ is larger than a threshold $\tau = 5$ (after the first 30 pulses, to allow the filter to converge), we drop the observation update.
We run a Rauch-Tung-Striebel~\cite{Rauch1965OnTM} smoother to refine the filtering estimates. \cref{tab:kf-params} describes more parameters for different sensors.

Originally, both the GPS and thermal camera used external triggers, but we found GPS trigger recording unreliable and external triggering of the thermal camera caused image tearing.  Therefore, we record these two sensors using system time and fit a transformation to the global clock with the Kalman filter, using the trigger signals from the IMU data packets.  All our sensor triggers are sent from our custom time synchronization board (\cref{fig:sync_board}).


\section{Searching the OctoSense dataset using natural language}
\label{sec:caption}
Finding interesting sequences for downstream tasks is a core challenge in dataset scaling. In driving data, for example, we often care less about average-case performance than about long tail events such as pedestrians in the roadway, animals, unusual road conditions and so on.  This is precisely why Waymo~\cite{Xu2025WODE2EWO} curates long tailed cases to evaluate model performance under rare scenarios. 

\begin{figure}[!b]
    \centering
    \includegraphics[width=\linewidth]{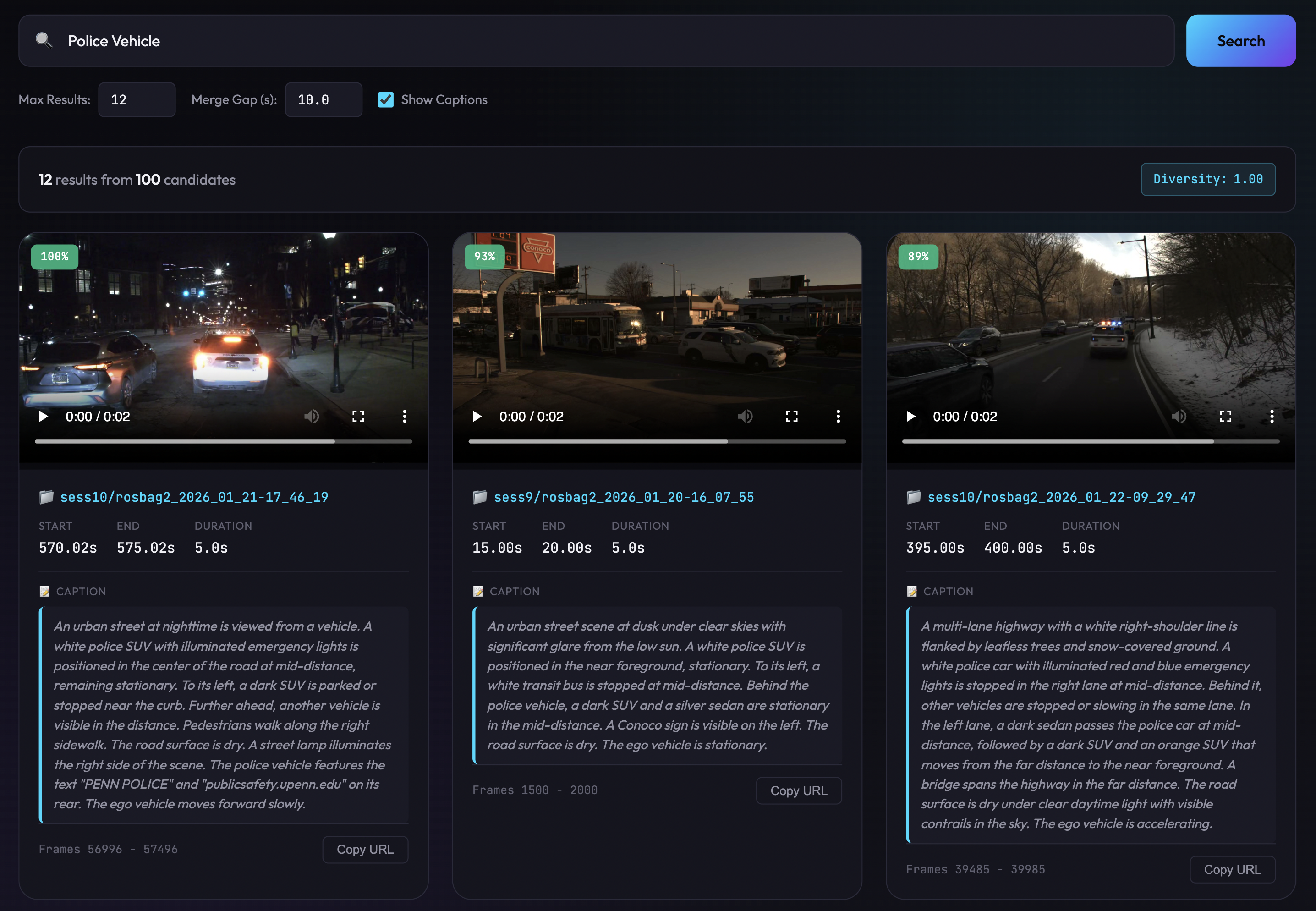}
\caption{\textbf{OctoSense Video Semantic Search} Our dataset tools allow you to search over our dataset with natural language. This example shows a Query: \emph{police vehicle}, above shows the results of the query, where we do find police vehicles in all result. }
    \label{fig:sem_search}
\end{figure}

To support this use case in our dataset, we built a semantic search pipeline that lets users query for specific events in natural language. We caption all image data at 5-second intervals using Gemma 4~\cite{gemma4_2026} with the system and captioning prompts shown below. Each captioning prompt is also enriched with a short ego-motion telemetry sentence (speed, turning, distance traveled, and local time-of-day) derived from the window's trajectory. The resulting captions are embedded with Qwen3-Embedding-8B~\cite{Zhang2025Qwen3EA}, and we build a FAISS~\cite{douze2024faiss} index over the full set of embeddings.  At query time, a user's natural language query is embedded with the same model and matched against the index. An example query is shown in \cref{fig:sem_search}.

\paragraph{System Prompt}\mbox{}\\
\begin{lstlisting}[style=promptstyle]
You are a driving scene annotator for a dashcam retrieval system. Captions are embedded for semantic search, so every sentence must describe something positively present in the image, searchable content only. Mentioning absent things pollutes the search index, so skip any category that does not apply and move on without comment. Describe only what you can identify with confidence: only call something a traffic light if you clearly see a signal housing with an illuminated red, green, or yellow lens; otherwise do not mention it. Only state a vehicle's color if lighting makes it unambiguous; otherwise describe the vehicle without color. Read sign text only if legible; otherwise call it an unreadable sign.

Example of correct style for a sparse scene:
'A two-lane rural road with a double yellow centerline curves right through dense trees. A silver sedan travels in the oncoming lane at mid-distance. The road surface is dry under overcast daytime light.'
Note the example simply omits lights, signs, and pedestrians rather than remarking on them.
\end{lstlisting}

\paragraph{Caption Prompt (Multi-Frame)}\mbox{}\\
\begin{lstlisting}[style=promptstyle]
You are given 3 dashcam frames from the same driving sequence, spaced ~2.5 seconds apart. {ego_motion_line} Write one paragraph of continuous prose describing the scene and what happens across the sequence, as a single continuous scene -- never reference frame numbers. Begin directly with the scene itself. Cover, in order, whichever of the following are actually present: (1) Road type -- choose the best fit: highway, on-ramp/off-ramp, urban street, urban intersection, residential street, rural two-lane road, dirt or gravel road, off-road trail, parking lot, parking garage, bridge, tunnel, T-junction, 4-way stop, roundabout -- plus lane count and markings. (2) Each visible vehicle: type, color if clearly discernible, lane position, distance (near, mid-distance, far), and how it moves across the sequence (e.g. a white SUV ahead pulls away, an oncoming truck passes). (3) Any illuminated traffic lights (red, green, yellow), including changes. (4) Any signs, text read verbatim. (5) Road surface: dry, wet, light snow cover, heavy snow cover, ice, slush, debris, or construction zone with cones or barriers. (6) Weather and lighting: clear, overcast, rain, light snow falling, heavy snow, fog, daytime, dusk, dawn, nighttime, glare. (7) Any emergency vehicles or unusual hazards. (8) Any pedestrians or cyclists, with position and action. (9) End with one clause on ego motion, phrased from the telemetry above if provided, otherwise inferred from how the scene changes (stationary, moving forward slowly, moving forward, accelerating, decelerating, turning left, turning right). Keep the paragraph under 150 words.
\end{lstlisting}

\section{Details of the architecture and hyper-parameters for the tokenizer and MAE}
\label{app:hparams}

\subsection{Event camera representation}
\label{app:ev_cam}

Event-cameras produce a set of events: $\{(u, p, t)\}_{i \in \integers}$, where $u \in \integers^2$ is the spatial location $p \in \{-1, +1\}$ is the sign of the intensity change since the last event and $t$ is the timestamp. We construct a frame-based representation $Z \in \mathbb{R}^{H\times W \times C}$, for sensor height $H$, width $W$ and $C$ temporal channels that encodes event activity at multiple temporal scales. The $c^{\text{th}}$ channel of the representation at pixel $u$, denoted by $Z_{uc}$, evolves according to an ordinary differential equation (ODE):
\[
\dv{}{t} Z_{uc} =\frac{-Z_{uc}}{\tau}+\alpha \sum_i  p_i \delta(t-t_i)
\]
for $t \in [t_i, t_{i+1})$ where $\tau$ and $\alpha$ are the time constant and the gain, and $\delta(\cdot)$ is a Dirac delta marking the arrival of an event. Solving this ODE yields a causal exponential sum over the past events: $Z_{uc}(t) = \alpha \sum_{t_i \leq t, u_i=u} p_i \exp(-\frac{t-t_i}{\tau})$.

We implement this filter as a streaming update. Each pixel stores the timestamp of its last update $t_i$. When a new event arrives at time $t$, we (1) decay the existing state by $\exp(-(t-t_l)/\tau)$, (2) add the contribution $\alpha p_i$, and (3) set $t_l \gets t$. The filter state can be read out asynchronously at any query time by applying a single decay step from each pixel's $t_l$ to the query time. This representation is equivalent to the time surfaces of \citet{Lagorce2017HOTSAH} and the continuous version of the discretized IIR filter of \citet{Bisulco2020FastMU}.

\subsection{IMU tokenization architecture}
\label{app:imu_arch}
Measurements from the 3-axis accelerometer and 3-axis gyroscope within a 1.6 s window (640 samples at 400 Hz) are processed into discrete tokens. We first break this window into chunks of 40 samples each.  Within each chunk, we apply a 3-layer 1D convolutional network with kernel size 5, increasing dilation (1,2,4), and channel counts (6, 64, 128). We then collapse each chunk into a single token via attention pooling with a learned query, followed by RMS normalization. This yields sixteen such ``tokens'', one per chunk, across the window. We pass these through a 4-layer transformer (4 heads, MLP ratio 2, RMSNorm) with 1D rotary position embedding across the chunks. Each chunk is projected into a 5-dimensional FSQ latent with levels [8,8,8,8,5], yielding a code-book of 20,480 entries and one discrete token for each chunk. The decoder mirrors the encoder comprising a 4-layer cross-token transformer with RoPE, followed by a per-chunk path that reconstructs each chunk token. On top of this, we apply an additional decoder which applies a 1D sinusoidal positional encoding and a 3-layer dilated 1D convolution stack (dilation 4,2,1) to reconstruct the 6-dimensional time series of the original signal.

\subsection{Decoder to train the image-based tokenizer}

\begin{figure}[h]
    \centering
    \includegraphics[width=\linewidth]{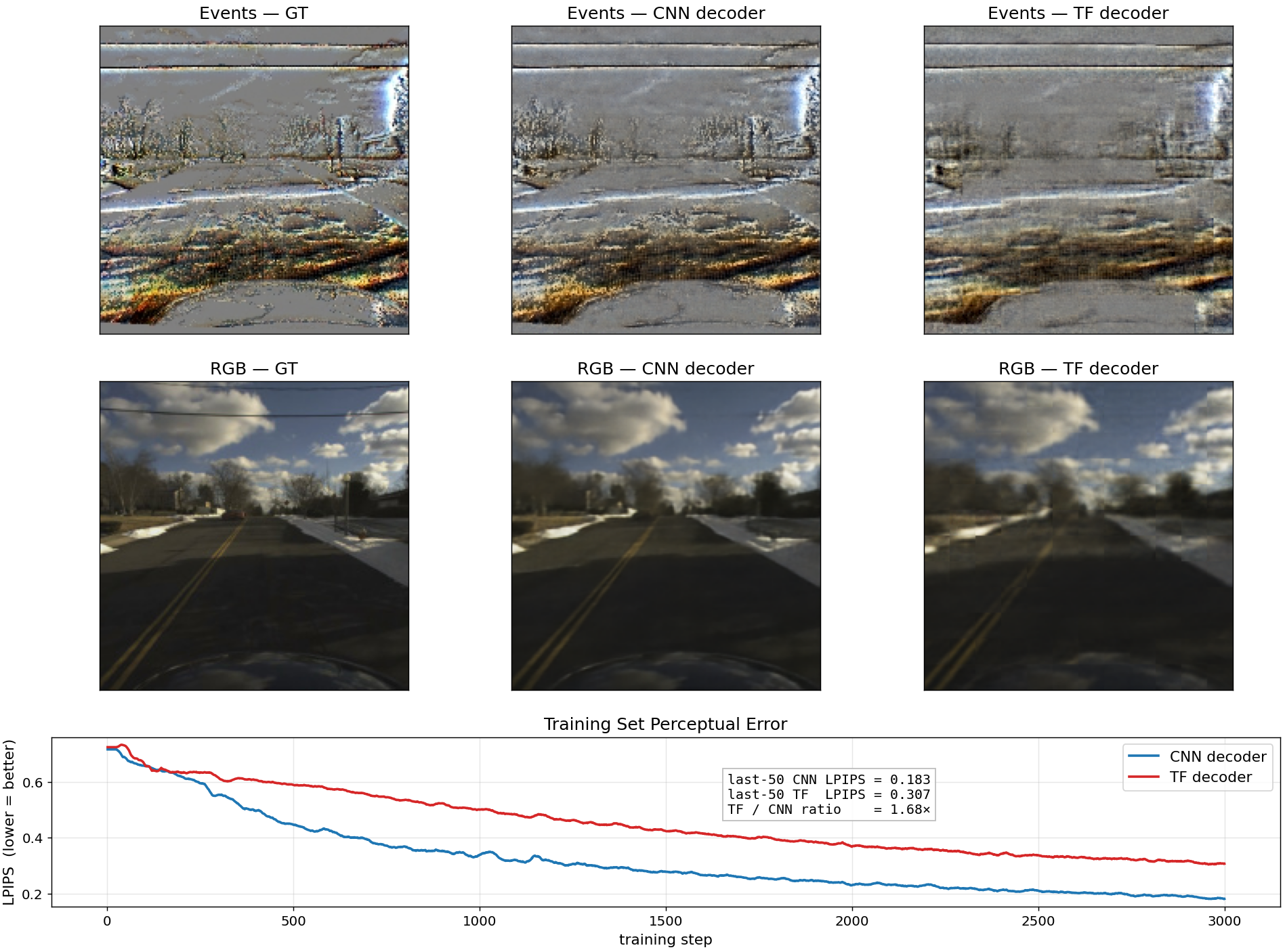}
    \caption{\textbf{Architecture for tokenizing Event/RGB camera data}
    During our initial investigations of our sensor tokenization architecture, we found that a transformer decoder with a linear readout led to blurred predictions compared to using a convolutional~(CNN) decoder. Across the columns we show the ground truth, the CNN-decoder prediction, and the Transformer-decoder prediction. At the bottom we show for an equal training budget the CNN decoder reaches lower error; we therefore adopt a CNN decoder for all our frame-based tokenizers. }
    \label{fig:convtoken}
\end{figure}

When building our frame-based sensor tokenizers, we noticed that our initial designs using a transformer decoder and a linear head led to blurry reconstructions~\cref{fig:convtoken}. We therefore use a convolutional decoder that maps the grid of latent codes to the reconstructed image through three upsampling stages. Each stage is a 3$\times$3 convolution to expand the channel dimension to that particular stage's upsampling factor, and these channels are reshaped into their respective spatial dimensions. Each upsampling stage is wrapped in a residual block consisting of GroupNorm-GELU-3$\times$3 conv applied twice, with a residual connection, followed by a dilated residual block to enlarge the receptive field. The channel width is halved at each stage, and a final 3$\times$3 convolution projects to the output channels.  We found that compared to a transformer decoder, our convolutional decoder yields better perceptual quality for the same number of training steps. Therefore, we adopt a convolutional decoder for all our frame-based output modalities.

\subsection{The multi-modal late-fusion MAE architecture}
The late-fusion architecture takes as input 8 frames each from the left event camera, left RGB camera, and LiDAR, along with 640 samples of IMU data. The first step of our architecture is encoding each of our inputs into tokens for a Transformer. For RGB/events/LiDAR, we apply 2D convolutions with kernel size and stride equal to the patch size. As for the IMU, we use the tokenization procedure described in \cref{app:imu_arch}.  All attention layers in the encoder and decoder use a RoPE on four axes: $(t, u_1, u_2, s)$, where t is the position within the sequence, $(u_1, u_2) $ are the patch coordinates and $s\in \{1,...,4\}$ is the modality index. Before encoding, we mask tokens at the granularity of spatiotemporal tubes, each tube being the set of tokens that share a patch location $(u_1, u_2)$ and modality $s$ across all timesteps. We mask a fraction of the tubes and pass only the visible tokens through the encoder; for all modalities, a 3-layer spatial-encoder transformer is applied to the visible tokens within a timestep per modality independently. Then, within each unmasked tube, we apply a 2-layer transformer encoder along the temporal axis, independently per tube, to exploit temporal information. Finally, we apply a 3-layer transformer encoder over the entire token sequence. Our decoder is an 8-layer transformer over the entire token sequence. We use a set of linear layers for each modality that map to the FSQ bottleneck of the tokenizer where all values are bounded to [-1,1]. Full details for the architecture's hyper-parameters are in \cref{tab:hparams_late_fusion}.

\begin{figure}
    \centering
    \includegraphics[width=\linewidth]{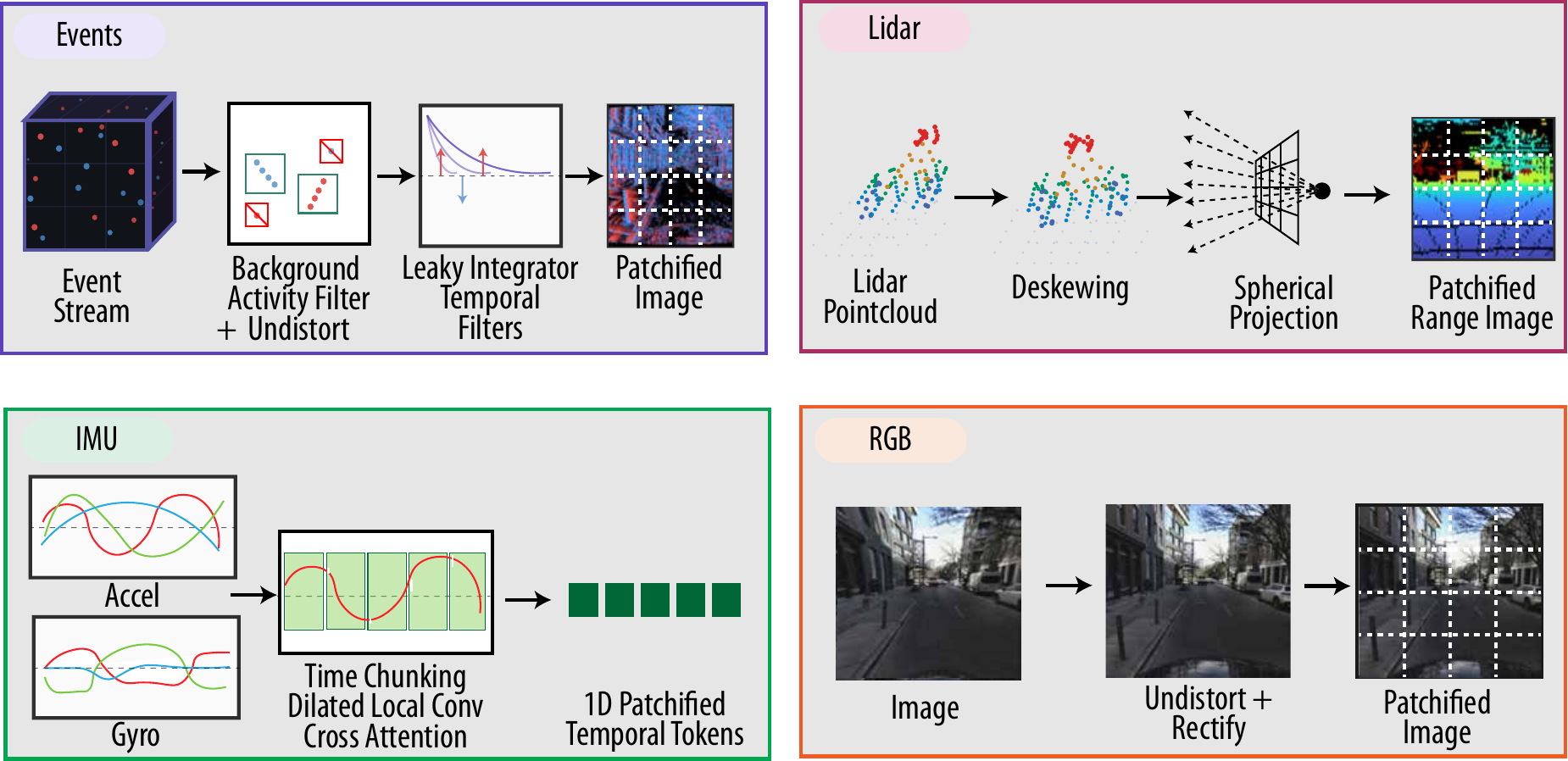}
    \caption{\textbf{Sensor Representations} Each modality is converted to a tokenizable form: events via a leaky integrator and patch-based tokenization; LiDAR via deskewed range images and patch-based tokenization; image via undistortion, rectification and patch-based tokenization; and IMU via a convolutional-cross-attention encoder \cref{app:imu_arch}. }
    \label{fig:method}
\end{figure}

\begin{figure}
\ContinuedFloat
    \centering
    \includegraphics[width=\linewidth]{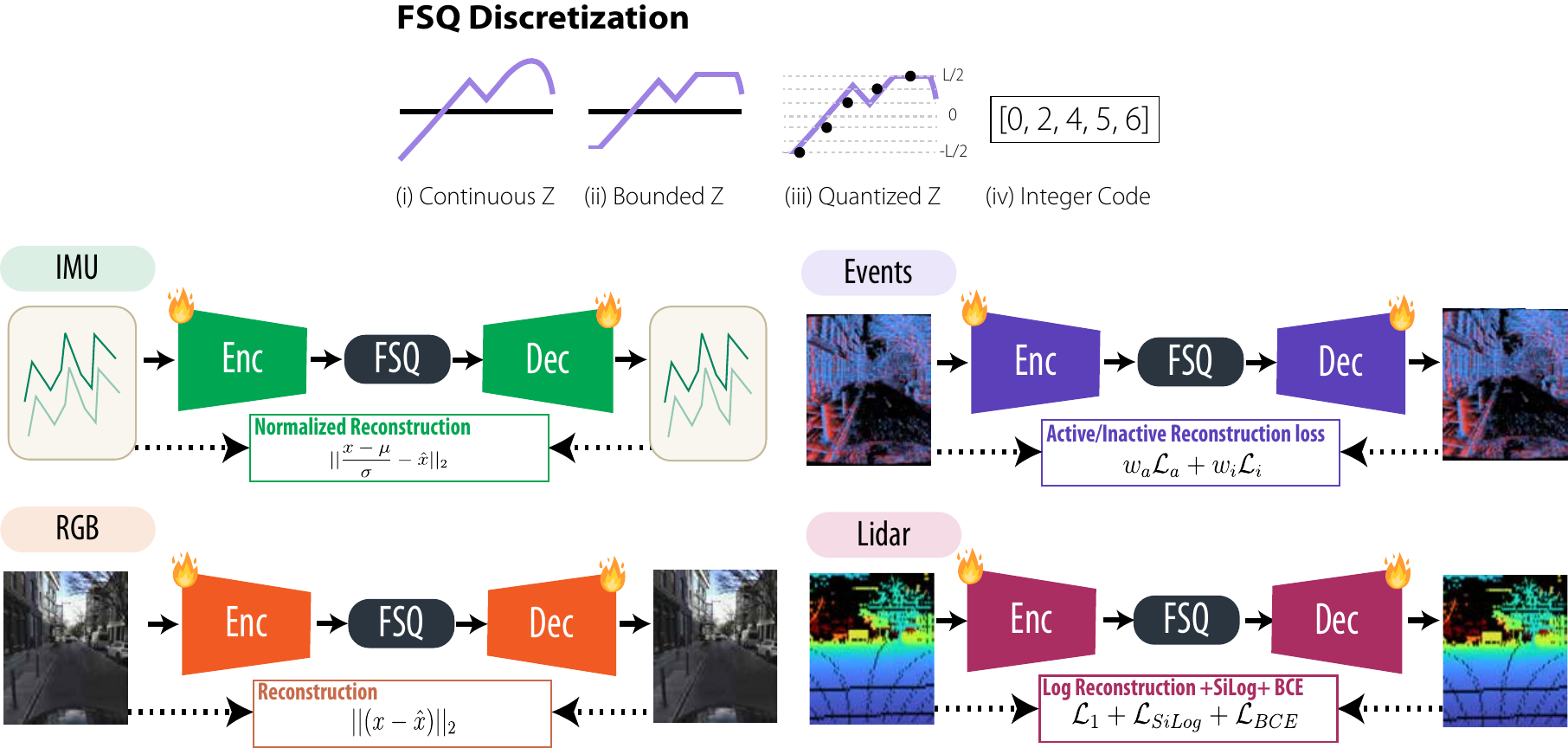}
    \caption{\textbf{Sensor Tokenized Targets} To balance learning progress across modalities, we predict frozen Finite Scalar Quantization (FSQ) codes in a bounded range, with a per-modality encoder/decoder trained beforehand.}
\end{figure}

\begin{figure}
\ContinuedFloat
    \centering
    \includegraphics[width=\linewidth]{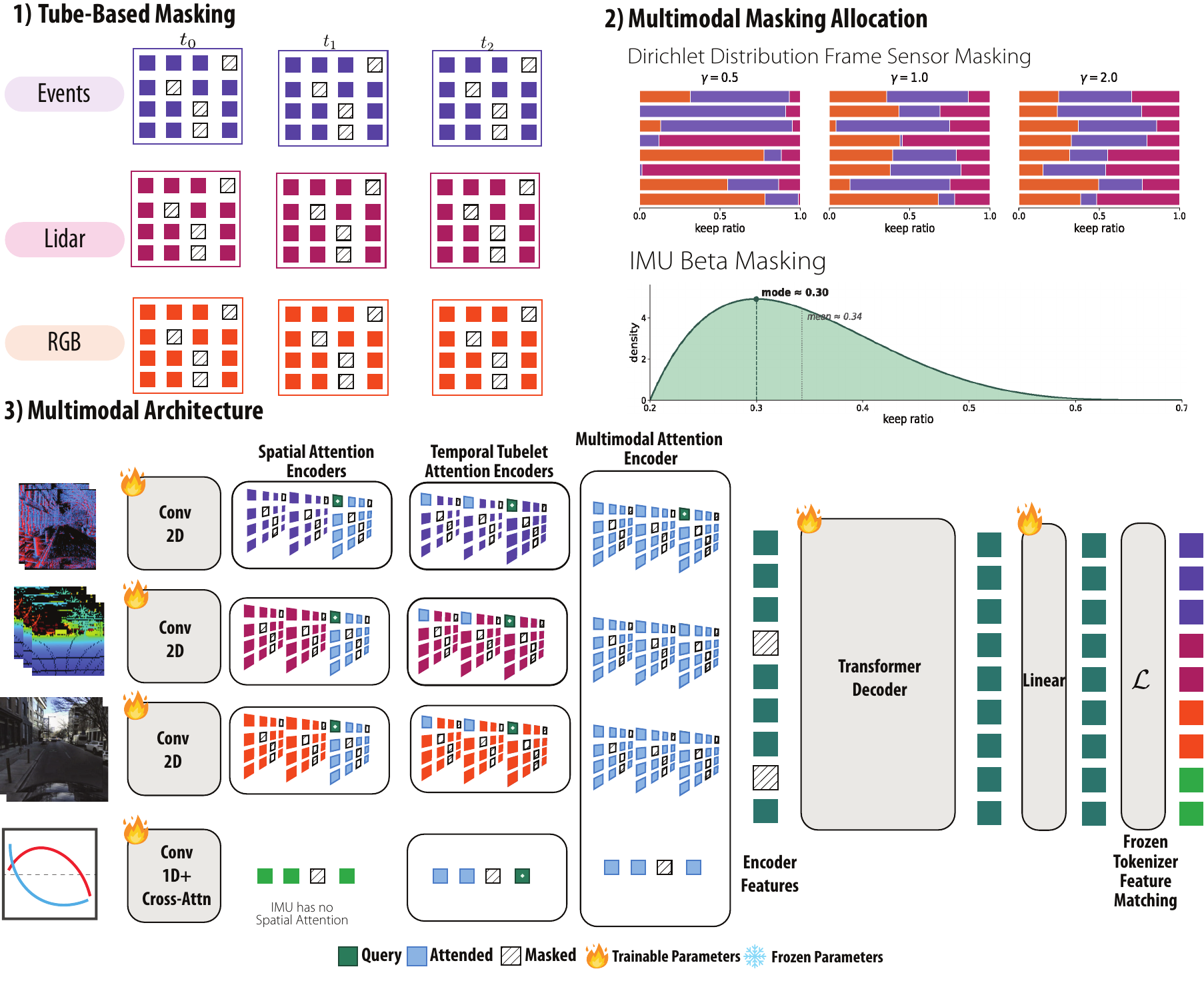}
    \caption{\textbf{Late-fusion MAE} Per-modality tokens are masked in spatiotemporal tubes with masking ratios per modality sampled from a Dirichlet distribution. The encoder applies attention in three stages: (i) \emph{spatial}, within each frame, (ii) \emph{temporal}, along per-patch tubes across time, (iii) \emph{multimodal}, joint attention over all tokens.  Mask tokens are inserted after encoding, and a shared transformer decoder with per-modality linear heads predicts the FSQ latent from (B); the loss is $\ell_1$ against the frozen targets.}
\end{figure}

\begin{figure}
\ContinuedFloat
    \centering
    \includegraphics[width=\linewidth]{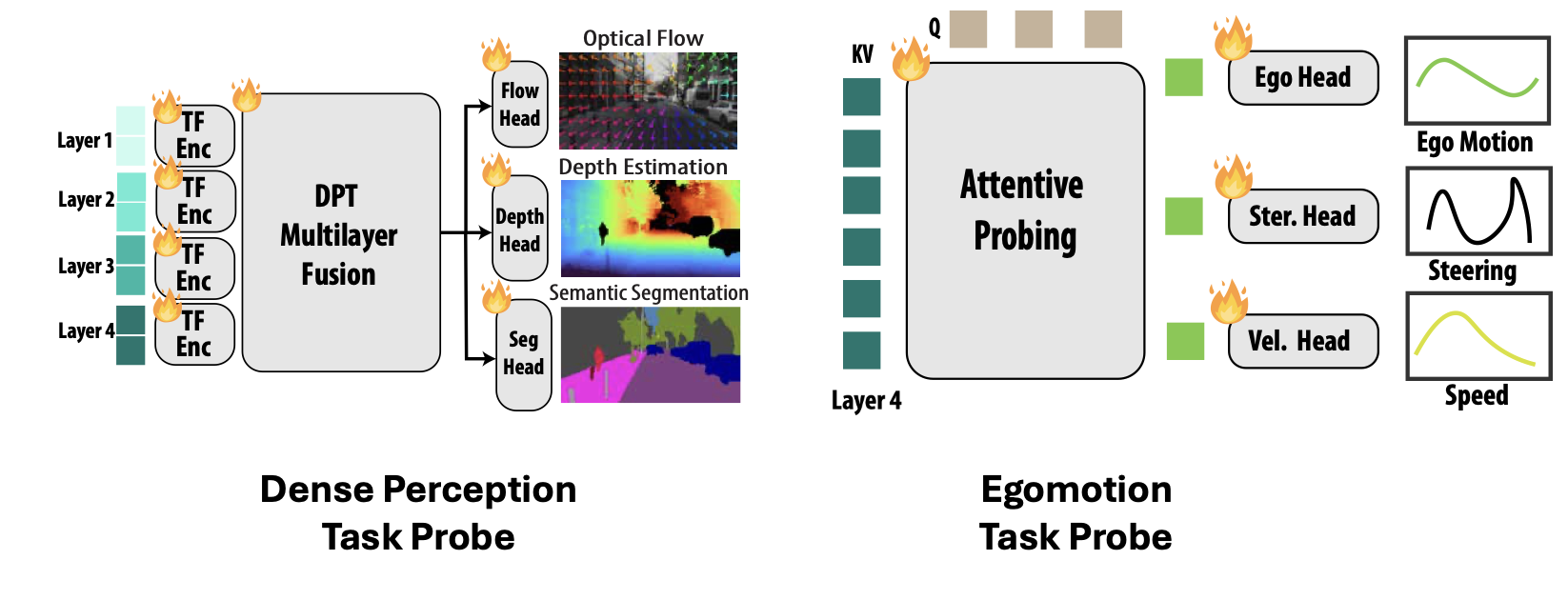}
    \caption{\textbf{Downstream probes} On the frozen representation we train a Dense Prediction Transformer to predict optical flow, depth and segmentation, and a separate attentive egomotion probe for relative pose, steering, speed and angular/linear velocity. }
\end{figure}

\begin{table}[h]
  \centering
  \small
  \setlength{\tabcolsep}{4pt}
  \caption{Training hyper-parameters for our four sensor tokenizers consumed by the late-fusion MAE.
  \textsuperscript{a} LiDAR is cropped to
  a 90$^\circ$ forward-looking view to obtain input of shape 64 $\times$512.
  \textsuperscript{b}IMU is a 1.6 sec window at 400 Hz with 6 channels (3-axis accelerometer + gyroscope) discretized into 16 chunks of 40 samples each.
  The notation TF-N stands for a transformer of N layers; Conv for residual convolutional blocks and 2-D pixel-shuffle for spatial upsampling of RGB/events/LiDAR. The IMU uses 1-D dilated convolutions.
  Tokenizers are trained with exponential moving averaging (EMA) of its parameters.
  ConvAttn is the architecture described in \cref{app:imu_arch}.}
  \vspace*{1ex}
  \label{tab:hparams_sensor_tokenizers}
  \renewcommand{\arraystretch}{1.2}
  \begin{adjustbox}{width=0.85\linewidth}
  \begin{tabular}{lcccc}
  \toprule
  & \textbf{RGB} & \textbf{Events} & \textbf{LiDAR} & \textbf{IMU} \\
  Input shape          & $3{\times}192{\times}192$ & $3{\times}192{\times}192$ & $2\textsuperscript{a}{\times}64{\times}512$ &
  $6{\times}640$\textsuperscript{b} \\
  Patch / chunks   & $16{\times}16$ & $12{\times}12$ & $8{\times}16$ & 40 \\
  Tokens per window    & 144 & 256 & 256 & 16 \\
  \midrule
  Encoder              & TF-6, $d{=}384$ & TF-6, $d{=}384$ & TF-6, $d{=}384$ & ConvAttn+TF-4, $d{=}128$ \\
  Decoder              & TF-6 + Conv & TF-6 + Conv & TF-6 + Conv & TF-4+Conv \\
  FSQ levels           & [8,8,8,8,5] & [8,8,8,8,5] & [8,8,8,8,5] & [8,8,8,8,5] \\
  TF Heads / MLP ratio    & 8 / 2.0 & 8 / 2.0 & 8 / 2.0 & 4 / 2.0 \\
  FSQ codebook size        & 20{,}480 & 20{,}480 & 20{,}480 & 20{,}480 \\
  \midrule
  Recon.\ loss         & L1 & L1 + inactive & L1 + SiLog + validity & L1 \\
  Optimizer            & \multicolumn{4}{c}{Muon ($2$-D weights) + AdamW (biases / norms), \texttt{match\_rms\_adamw} LR
  rescale} \\
  Learning rate        & \multicolumn{4}{c}{$10^{-3}$, OneCycleLR, $\mathrm{pct\_start}{=}0.1$} \\
  Precision            & \multicolumn{4}{c}{bf16-mixed, gradient clip $0.5$, EMA $0.999$} \\
  \midrule
  GPUs                 & $4{\times}$B200 & $4{\times}$B200 & $4{\times}$B200 & $1{\times}$B200 \\
  Batch size           & 192 & 192 & 192 & 2048 \\
  Epochs               & 60 & 60 & 60 & 60 \\
  \bottomrule
  \end{tabular}
  \end{adjustbox}
  \end{table}

\begin{table}
    \centering
    \small
    \caption{\textbf{Training hyper-parameters for the late-fusion MAE} which contains three stages: modality-specific spatial encoder $\rightarrow$
    modality-specific temporal encoder $\rightarrow$ cross-modal fusion. Frozen FSQ tokenizers
    (\cref{tab:hparams_sensor_tokenizers}) supply the reconstruction targets.
    \textsuperscript{a} LiDAR is cropped to
  a 90$^\circ$ forward-looking view to obtain input of shape 64 $\times$512.
   \textsuperscript{b}IMU is a 1.6 sec window at 400 Hz with 6 channels (3-axis accelerometer + gyroscope) discretized into 16 chunks of 40 samples each.
  Events are denoised with a 3$\times$3 spatial and 1 ms temporal window, filtered with time constants $\tau=$ [8, 18, 44] ms and impulse gains $\alpha=$[0.0247, 0.01105, 0.00454] before gamma correction with coefficients [0.4,0.5,0.6] and per-channel normalization with coefficients [0.5289, 0.3912, 0.2587].}
    \label{tab:hparams_late_fusion}
    \renewcommand{\arraystretch}{1.25}
    \begin{adjustbox}{width=0.9\linewidth}
    \begin{tabular}{lcccc}
    \toprule
     & \textbf{RGB} & \textbf{Events} & \textbf{LiDAR} & \textbf{IMU} \\
    \midrule
    Input shape          & $8{\times}3{\times}192{\times}192$ & $8{\times}3{\times}192{\times}192$ &
                            $8{\times}1{\times}64{\times}512$\textsuperscript{a} & $6{\times}640$\textsuperscript{b} \\
    Patch / atom         & $16{\times}16$ & $12{\times}12$ & $8{\times}16$ & 40 samples \\
    Tokens per frame     & $144$ & $256$ & $256$ &  \\
    Tokens per window    & $1{,}152$ & $2{,}048$ & $2{,}048$ & $16$ \\
    Front-end            & Conv2d & Conv2d & Conv2d & \hyperref[app:imu_arch]{ConvAttn} \\
    FSQ target           & \multicolumn{4}{c}{frozen FSQ, levels $[8{,}8{,}8{,}8{,}5]$, $20{,}480$ codes/token} \\
    Loss / weight        & $\ell_1$ / $1.0$ & $\ell_1$ / $1.0$ & $\ell_1$ / $1.0$ & $\ell_1$ / $1.0$ \\
    \midrule
    Spatial encoder      & TF-3, $d{=}576$ & TF-3, $d{=}576$ & TF-3, $d{=}576$ & --- \\
    Temporal encoder     & TF-2, $d{=}576$ & TF-2, $d{=}576$ & TF-2, $d{=}576$ & TF-2, $d{=}576$ \\
    Cross-modal encoder  & \multicolumn{4}{c}{TF-3, $d{=}576$, TF Heads${=}9$, MLP ratio $4.0$} \\
    Decoder              & \multicolumn{4}{c}{TF-8, $d{=}384$, TF Heads${=}12$ MLP ratio $4.0$} \\
    Positional encoding  & \multicolumn{4}{c}{4D RoPE over (patch-x, patch-y, time, modality)} \\
    \midrule
    Masking              & \multicolumn{4}{c}{Dirichlet masking policy, mean masking ratio $(r=0.8) \,+$ spatiotemporal tubelet masking} \\
    IMU keep             & \multicolumn{4}{c}{$0.2 + 0.5\cdot\mathrm{Beta}(2,5)$ → $[0.2, 0.7]$} \\
    \midrule
    Optimizer            & \multicolumn{4}{c}{Muon (2-D weights) $+$ AdamW (biases / norms), \texttt{match\_rms\_adamw} LR
  rescale} \\
    Learning rate        & \multicolumn{4}{c}{$1{\times}10^{-3}$, OneCycleLR, $\mathrm{pct\_start}{=}0.1$} \\
    Precision            & \multicolumn{4}{c}{bf16-mixed, gradient clip $1.0$} \\
    Batch (per GPU / eff.) & \multicolumn{4}{c}{$128 \,/\, 512$} \\
    Sequence             & \multicolumn{4}{c}{$T{=}8$ frames, $\Delta t{=}0.2$\,s, stride $0.4$\,s} \\
    Epochs               & \multicolumn{4}{c}{$150$} \\
    \midrule
    Hardware             & \multicolumn{4}{c}{$4\times$ NVIDIA B200, DDP} \\
    Seed                 & \multicolumn{4}{c}{$42$} \\
    \bottomrule
    \end{tabular}
    \end{adjustbox}
  \end{table}

   \begin{table}[h]
    \centering
    \caption{\textbf{Architectures used for evaluation on different tasks.} Each encoder is frozen and evaluated with a probe on either its last layer features for egomotion tasks or multiple intermediate features for dense prediction tasks. The four left columns are our masked autoencoder variants, first two using only RGB, second two using multi-modal data. \emph{4-mod} denotes the four input modalities (RGB, events, LiDAR, IMU).  As compared to ViT-B our late-fusion architecture does not have four decoder layers, so we use layer indices [1,2,3,3] for DPT probing.
    \textsuperscript{a} DINOv2/v3 both feature multi-resolution training, 224 is one of these resolutions, and that is the one we use.}
    \vspace*{1ex}
    \label{tab:probe_encoders}
    \renewcommand{\arraystretch}{1.35}
    \begin{adjustbox}{width=\linewidth}
    \begin{tabular}{lcccccccccc}
    \toprule
    & \makecell{RGB-only\\image MAE} & \makecell{RGB-only\\video MAE} & \makecell{Early-fusion\\MAE} &
  \makecell{Late-fusion\\MAE} & DINOv2 & DINOv3 & V-JEPA 2.1 & SigLIP 2 & PE \\
    \midrule
    Backbone               & ViT-B & ViT-B & ViT-B & 3-stage fusion & ViT-B & ViT-B & ViT-B & ViT-B & ViT-B \\
    Params (encoder)       & $85.6$\,M & $85.6$\,M & $86.2$\,M & $81.6$\,M & $86.2$\,M & $85.7$\,M & $86.8$\,M & $85.8$\,M &
  $85.8$\,M \\
    Depth (blocks)         & $12$ & $12$ & $12$ & $3{+}2{+}3$ & $12$ & $12$ & $12$ & $12$ & $12$ \\
    Hidden dim             & $768$ & $768$ & $768$ & $576$ & $768$ & $768$ & $768$ & $768$ & $768$ \\
    Patch size             & $16$ & $16$ & $16$ & $-$ & $14$ & $16$ & $16$ & $16$ & $16$ \\
    Native input           & $192^2$ & $192^2$ & $192^2$ & $192^2$ & $224^2$ \textsuperscript{a} & $224^2$ \textsuperscript{a} & $384^2$ & $256^2$ & $224^2$ \\
    Frames / token         & $1$ & $1$ & $1$ & $1$ & $1$ & $1$ & $2$ & $1$ & $1$ \\
    RGB tokens / frame     & $144$ & $144$ & $144$ & $144$ & $256$ & $196$ & $576$ & $256$ & $196$ \\
    Modalities ingested    & RGB & RGB & 4-mod & 4-mod & RGB & RGB & RGB & RGB & RGB \\
    \midrule
    Hook target            & last block & last block & last block & crossmodal stage & last block & last block & last block &
  last block & last block \\
    Joint-DPT layers       & $[3,6,9,12]$ & $[3,6,9,12]$ & $[3,6,9,12]$ & $[1,2,3,3]$ & $[3,6,9,12]$ & $[3,6,9,12]$ &
  $[3,6,9,12]$ & $[3,6,9,12]$ & $[3,6,9,12]$ \\
    Ego attentive layer    & $\{12\}$ & $\{12\}$ & $\{12\}$ & $\{3\}$ & $\{12\}$ & $\{12\}$ & $\{12\}$ & $\{12\}$ & $\{12\}$ \\
    \bottomrule
    \end{tabular}
    \end{adjustbox}
  \end{table}

   \begin{table}[!t]
    \centering
    \caption{\textbf{Hyper-parameters for training task evaluation probes} Probes are used with frozen encoders in \cref{tab:probe_encoders}. An attentive probe reads the features of the last layer to predict ego-motion and a dense prediction transformer (DPT) takes as input a multi-scale pyramid of features at intermediate layers. Same probe architectures and training recipe are used for all encoders. Details on training losses and training procedures can be found in \cref{app:eval}. As part of preparing the evaluation data, we remove any window overlapping a pose transient (linear acceleration above $1.5g$) within 2s of a bag boundary; these arise from RKO-LIO odometry estimation \href{https://github.com/PRBonn/rko_lio/issues/139}{issues}. }
    \label{tab:probe_arch}
    \renewcommand{\arraystretch}{1.25}
    \begin{adjustbox}{width=\linewidth}
    \begin{tabular}{lcc}
    \toprule
     & Dense Perception Tasks Probe & Attentive Ego-motion  Probe \\
    \midrule
    Reader                 &  Dense Prediction Transformer~(DPT) & $2{\times}$ cross-attn block \\
    Layers from encoder    & $4$ (multi-scale pyramid) & $1$ (last block) \\
    Hidden width           & $64$ ch (\texttt{dpt\_channels}) & matches encoder hidden \\
    Heads / queries        & --- & $8$ heads, learned queries (3 readouts) \\
    Task targets & Depth, Optical flow, Cityscapes-19 seg & rel.\ pose $(R, t)$, body-frame $(v,
  \omega)$, CAN speed $+$ steer \\
    Losses                 & SiLog $+$ Charbonnier $+$ Cross Entropy & Chordal $+$ Huber \\
    \midrule
    Optimizer              & \multicolumn{2}{c}{AdamW} \\
    Learning rate          & \multicolumn{2}{c}{$10^{-3}$} \\
    LR schedule            & linear warm-up ($10\%$) $\to$ cosine & OneCycle (cosine, $10\%$ warm-up) \\
    Precision              & \multicolumn{2}{c}{bf16-mixed, gradient clip $1.0$} \\
    Batch (per GPU / eff.) & $32 \,/\, 128$ & $128 \,/\, 512$ \\
    Epochs                 & \multicolumn{2}{c}{$60$} \\
    \midrule
    Hardware               & \multicolumn{2}{c}{$4\times$ NVIDIA B200, DDP} \\
    Eval Filter     & \multicolumn{2}{c}{$\mathrm{gt\_speed} \geq 5$\,mph,\ $|a| < 1.5\,g$ in first/last 2\,s} \\
    Seed                   & \multicolumn{2}{c}{$42$} \\
    \bottomrule
    \end{tabular}
    \end{adjustbox}
  \end{table}

\clearpage
\section{Evaluation Protocol}
\label{app:eval}

\subsection{Evaluation of existing pretrained models}
All probes operate on features of the frozen encoder. The five vision-only baselines are fed only the left RGB stream of our 8-frame window, bilinearly resampled to each encoder's native resolution from 192 $\times$ 192 resolution and normalized with either ImageNet statistics (DINOv2, DINOv3, V-JEPA 2.1) or $[-1,1]$ scaling (SigLIP 2, Perception Encoder) to match each model's preprocessing. Per-frame 2D ViTs (DINOv2/v3, SigLIP 2, Perception Encoder) run independently on each frame and their per-frame token grids are concatenated in the order of these frames (register and CLS tokens are removed). V-JEPA 2.1 is a spatiotemporal model which takes frames in pairs of two; we repeat the output tokens of V-JEPA 2.1 to recover per-frame tokens for single-frame decoding on dense tasks.

\begin{figure}[h]
    \centering
    \begin{minipage}[c]{0.38\linewidth}
        \centering
        \includegraphics[width=\linewidth]{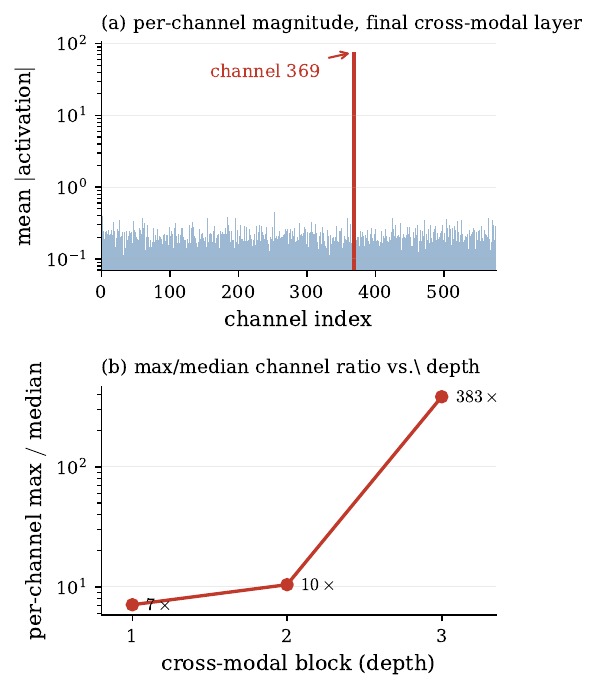}
    \end{minipage}%
    \begin{minipage}[c]{0.61\linewidth}
        \centering
        \scriptsize
        \setlength{\tabcolsep}{3pt}

        \textbf{(a) Dense Prediction Tasks Errors}\\[2pt]
        \begin{tabular}{llrrr}
        \toprule
        Encoder & Norm & Depth\,(m)\,$\downarrow$ & Flow\,(px)\,$\downarrow$ & Seg\,(mIoU)\,$\uparrow$\\
        \midrule
        \multirow{3}{*}{\shortstack[l]{Late-Fusion\\MAE}}
        & \textbf{LayerNorm}\,$\checkmark$ & 4.73 & 1.97 & 0.411\\
        & BatchNorm & 4.81 & 2.06 & 0.404\\
        & Hybrid (BN+LN) & 4.71 & 2.05 & 0.388\\
        \midrule
        \multirow{3}{*}{DINOv2}
        & \textbf{LayerNorm}\,$\checkmark$ & 6.82 & 19.46 & 0.410\\
        & BatchNorm & 6.84 & 19.28 & 0.422\\
        & Hybrid (BN+LN) & 6.92 & 20.05 & 0.426\\
        \bottomrule
        \end{tabular}

        \vspace{1.0ex}

        \textbf{(b) Egomotion Tasks Errors}\\[2pt]
        \resizebox{0.95\linewidth}{!}{%
        \begin{tabular}{llrrrr}
        \toprule
        Encoder & KV norm & Trans\,(m)\,$\downarrow$ & Rot\,($^\circ$)\,$\downarrow$ & Lin.vel\,(m/s)\,$\downarrow$ & Speed\,(mph)\,$\downarrow$\\
        \midrule
        \multirow{3}{*}{\shortstack[l]{Late-Fusion\\MAE}}
    & \textbf{raw}\,$\checkmark$ & 0.06 & 0.24 & 0.36 & 0.78\\
    & BatchNorm & 0.05 & 0.17 & 0.32 & 0.75\\
    & Enc.\ final-norm & 0.07 & 0.29 & 0.43 & 1.06\\
\midrule
  \multirow{3}{*}{DINOv2}
    & \textbf{raw}\,$\checkmark$ & 0.96 & 0.80 & 4.92 & 5.84\\
    & BatchNorm & 0.93 & 0.65 & 4.63 & 5.85\\
    & Enc.\ final-norm & 0.95 & 0.67 & 4.69 & 5.93\\
        \bottomrule
        \end{tabular}}
    \end{minipage}

    \caption{\textbf{Late-fusion MAE outlier and feature normalization choices.}
    \textbf{Left, top:} per-channel mean activation magnitude for a single sample in the final late-fusion layer
    reveals a high-norm outlier (channel 369).
     \textbf{Left, bottom:} the max/median channel ratio grows sharply with encoder depth, highlighting this issue becomes larger in later layers.
     \textbf{Right:} probe results are robust to the feature-normalization choice.
    We read the frozen encoder at its pre-final-norm layer and normalize on the probe side. Across normalizers the differences are small for both a multimodal encoder and an RGB baseline (DINOv2), so we adopt the choices ($\checkmark$): per-token layer-norm
    for the dense DPT and raw (unnormalized) keys-values for the
    attentive ego-motion head. These results are provided for OctoSense \texttt{test\_day}, motion-filtered $\geq$5\,mph.}
    \label{fig:hmae_outlier}
\end{figure}

\subsection{Task prediction heads}

\paragraph{Dense prediction transformer}
We use the DPT head for all our frozen encoders. Features are acquired from four uniformly spaced encoder layers and we pass our features through per-layer LayerNorm before the DPT. Additionally, we use a uniform 64-channel pyramid and GroupNorm for efficient and stable training on our DPT. A single transformer block per input layer equipped with an additive 4D positional encoding$(t, u_1,u_2,s)$ mixes information over the encoder's full token sequences before we slice out the per-frame left-RGB tokens. This lets per-frame vision-only encoders exploit temporal context and lets the multi-modal tokens route LiDAR/event/IMU information to the RGB token slots for the decoder. The four feature maps are then projected to 64 channels and reassembled into a feature pyramid (4 ${\times}$ and 2${\times}$ transpose-conv up, identity, and stride-2 conv down), fused across early layers with skip connections through pre-activation residual conv units and bilinearly upsampled to the ground-truth resolution. Each task head is a 1${\times}$1 convolution, i.e., a per-pixel linear projection, over this shared feature map. The depth and segmentation heads read each frame's features independently, while the flow head reads the channel-wise concatenation of feature maps of consecutive frames to regress the flow field (note that it produces 7 flow fields).

\paragraph{Ego-motion head}
Three independent attention-based readouts cross-attend using learnable query tokens (two cross-attention blocks each) to the last layer of the encoders: velocity(8 queries are mapped to per-frame $[v_x, v_y, v_z, \omega_x, \omega_y, \omega_z]$ in the rectified left-camera frame), relative pose (7 queries are mapped to $\mathrm{SE}(3)$ relative pose between successive image pairs), and CAN bus outputs (8 queries are mapped to per-frame speed and steering position).

\paragraph{Feature outliers}
Consistent with \citet{simeoni2025dinov3}, we observe outlier features with a large norm in our pretrained models. Because we adopt a pre-normalization transformer architecture~\cite{Xiong2020OnLN}, the output features are unnormalized and vulnerable to outliers. \cref{fig:hmae_outlier} shows outliers in channel 369, which has a magnitude that is 383$\times$ the median channel magnitude. We evaluated several normalization schemes for our probes. For egomotion we compared raw encoder features, batch-normalized features and features passed through the encoder's final norm layer. For the DPT probe, we compared layer-normalized features, batch-normalized features, and a hybrid scheme of applying batch norm to the three early layers and applying  layer norm to the last. For the DPT probe, variations across the schemes were small, so we adopted Layer Normalization. For egomotion, batch normalization was marginally better than raw features, but the differences are still small compared to our improvement over baselines, so we adopted raw features for simplicity. While layer normalization can in principle be overwhelmed by the outlier channel, we did not observe any deterioration in our empirical results.

\subsection{Loss functions for downstream tasks}

\noindent \ul{Depth:} Square of the scale-invariant logarithmic loss given by $\ell(\l) = \frac{1}{n} \sum_i d_i^2-\frac{\lambda}{n^2} (\sum_i d_i)^2$ with $d_i=\log \hat z_i - \log z_i$ at the $i^{\text{th}}$ pixel among a total $n$ pixels with $\lambda=0.15$ (head regresses $\log z_i$ directly). This loss is imposed only for pixels with $z \in [3, 200]$ m. For evaluation, we report the standard SiLog loss (i.e., after taking the square root of our loss) with the conventional $\l = 0.5$.

\noindent \ul{Flow:} Charbonnier loss given by $\ell = \sqrt{x^2+\varepsilon^2}-\varepsilon$ with $\varepsilon=10^{-3}$ on the end-point error $x$ between the predicted and the ground-truth optical flow.

\noindent \ul{Segmentation:} Cross-entropy loss across all classes on valid output pixels.

\noindent \ul{Rotation:} The relative-pose head emits a 3 $\times$ 3 matrix $\hat R$ which is fitted with the square of the chordal distance, i.e., squared Frobenius norm $\norm{\hat R -R}_\text{F}^2 $\cite{Geist2024LearningW3,Levinson2020AnAO}. During inference, we project the prediction $\hat R$ onto $\mathrm{SO}(3)$ under the Frobenius norm. This is done by computing the SVD $\hat R = U \Sigma  V^\top$ and setting $\hat R \equiv U \text{diag}(1,1,\text{det}(UV^\top))V^\top$. The $\text{det}(UV^\top)$ term flips the sign of the last component when needed so that $\text{det}(\hat R) = +1$.

\noindent \ul{Scalar Targets:} Velocity, translation, and the CAN prediction  use a Huber loss
  $$
    \ell(x) =
    \begin{cases}
      \tfrac{1}{2}x^2 & |x| \le 1\\[2pt]
      |x| - \tfrac{1}{2} & |x| > 1,
    \end{cases}
  $$
  where $x$ is the per-channel residual. Before applying the loss, we rescale each target by fixed constants: $\text{vel}=[5,5,25,0.3,0.5,0.3]$, $\text{trans}=[0.5,0.5,3]$. CAN speed is regressed in $\log(1+x)$ space and steering is z-scored using statistics computed over the dataset (standard deviation is clamped at $5^\circ$).

All $\mathrm{SE}(3)$ and $\mathrm{SO}(3)$ operations, relative pose targets, SVD onto $\mathrm{SO}(3)$ and the rotation-error metrics are computed in FP32. This is required for two reasons: torch.linalg.svd has no bf16 CUDA kernel and the targets are derived from absolute world-frame poses with translations spanning hundreds of meters. Using bf16 can inject errors of several meters in relative translation.

\subsection{Evaluation metrics}

Metrics are computed across all valid pixels when the vehicle is traveling above 5 mph.

\noindent \ul{Segmentation:} Over the 19 classes of Cityscapes we form a global confusion matrix and report  (i) mean IoU where the per-class Jaccard index is $\text{IoU}_c=\frac{\mathrm{TP_c}}{\mathrm{TP_{c}}+\mathrm{FP}_c+\mathrm{FN}_c}$  where TP stands for true positives, FP for false positives, FN for false negative, and (ii) accuracy $\sum_c \mathrm{TP}_c/N$ which is the fraction of valid pixels that are labeled correctly.

\noindent \ul{Depth:} The depth head predicts logarithmic depth $\log \hat z_i$ at a pixel $i$ with depth between [3, 200] m. Let $d_i = \log \hat z_i - \log z_i$ over $n$ pixels. We calculate the absolute relative error, the SiLog loss, root mean square error (RMSE) and inlier rate with $\delta<1.25$:
\begin{align*}
  \text{abs.\,rel.} &= \f{1}{n} \sum_i \frac{\abs{\hat z_i - z_i}}{z_i}\\
  \text{RMSE} &= \rbr{\f{1}{n}  \sum_i \rbr{\hat z_i - z_i}^2}^{1/2}\\
  \text{SiLog} &= \rbr{\f{1}{n} \sum_i d_i^2- \f{1}{2 n^2} (\sum_i d_i)^2}^{1/2}\\
  \delta < 1.25 &= \frac{1}{n} \abs{\text{max} \rbr{ \f{\hat z_i}{z_i}, \f{z_i}{\hat z_i}  } < 1.25 }.
  \end{align*}

\noindent \ul{Optical Flow:} The ground truth is the per-pixel 2D displacement due to ego-motion $f_u \in \reals^2$ for a pixel $u \in \integers^2$. Over valid pixels, we report (i) average end-point error which is the mean euclidean discrepancy $e_u = \norm{\hat f_u - f_u}_2$ between the prediction and the ground truth, and (ii) the outlier rate which is the fraction of pixels whose error exceeds 3 pixels and 5\% of the ground-truth flow magnitude:
\begin{align*}
    \text{EPE} &= \frac{1}{n} \sum_i e_i\\
    \text{outlier} &= \frac{1}{n} \abs{\cbr{u: e_u > 3 \text{ and } \f{e_u}{\norm{f_u}} > 0.05 }}.
\end{align*}

\noindent \ul{Ego-motion:} The attentive probe regresses the instantaneous linear and angular velocities $(\mathbf v,\boldsymbol\omega)$, the relative inter-frame pose in the rectified left-camera frame and two CAN-bus outputs (vehicle speed and steering position). We report the RMSE for all of these. Rotation error is the average of the geodesic angle between the predicted and ground-truth rotation in degrees:
\[
    \arccos \rbr{ \f{\text{tr}(\hat R^\top R) -1}{2}}.
\]
Speed (mph), steering ($^\circ$), linear velocity ($m/s$) and angular velocity ($rad/s$) statistics are computed directly in their native units.

\subsection{Evaluation on the M3ED dataset}

The M3ED dataset with the same Ouster OS1-64 LiDAR, event cameras (a different Prophesee EVK4 one than ours), RGB cameras (different 1/4" AR0144 cameras than ours), IMU and ground-truth odometry is a good candidate to evaluate the cross-dataset generalization of our approach.
We evaluate optical flow, depth estimation and ego-motion prediction on M3ED. We use the same procedure as in OctoSense to generate the ground-truth. M3ED provides segmentation data only in the left event-camera frame which cannot be easily transferred to the RGB camera for evaluation. We therefore do not perform any segmentation evaluation for M3ED.

Due to sensor differences, we make a few modifications to our input representations:
(i) LiDAR data is not deskewed but is still cropped to a  similar field of view as that of OctoSense,
(ii) M3ED's IMU Z-axis points in the vehicle forward direction so we rotate the IMU measurements to be consistent with the OctoSense frame,
(iii) event-camera in M3ED is not stereo-rectified so we undistort, center-crop, and downsample events (192 $\times$ 192) and apply the same leaky-integrator filters with the same parameters,
(iv) the RGB stream is undistorted but not rectified (no stereo pair is available), center-cropped and downsampled (192 $\times$ 192).
We split the 16 car sequences in M3ED with ground truth into 13 training sequences and 3 test sequences (\texttt{car\_urban\_day\_city\_hall}, \texttt{car\_urban\_night\_rittenhouse}, \texttt{car\_forest\_into\_ponds\_short}). We use the same 1.4 sec non-overlapping windows to construct samples.  We evaluate our pretrained models by training ego-motion and dense probes for each task using our pretrained late-fusion model on the OctoSense dataset. The dense probe omits the segmentation loss and our ego-motion probe only uses two cross attention readout modules. All probe hyper-parameters, losses, metrics~(reported without motion filtering) match those used for the late-fusion MAE on OctoSense.


\section{Analysis of the computational cost of the late-fusion MAE}

A perception backbone deployed on a moving robot must keep pace with its sensor streams (200 ms per step in our current setup) while leaving budget for downstream planning. Faster future sensor rates will demand correspondingly faster inference. We designed our late-fusion MAE explicitly for this constraint: (i) per-modality spatial encoders are applied independently to each frame-based sensor stream, (ii) per-modality temporal tubelet encoders aggregate information across the time window, and (iii) a final cross-modal encoder attends to these tokens.
We compare the early-fusion MAE (85.0M,  $W{=}768$, $D{=}12$) with our late-fusion MAE (83.7M, $W{=}576$) with 3, 2 and 4 blocks for the spatial, temporal, and cross-modal stages, respectively, at roughly similar parameter counts. The 1.5\% parameter gap reflects that the three-stage decomposition does not admit an exact match.
 
\paragraph{Streaming inference by caching spatial features} Sensor data arrives incrementally over time. Spatial activations for the past frames can therefore be cached.
An early fusion architecture would run all $L$ blocks over all $N$ tokens each step because self-attention couples every past token to the new tokens. A late-fusion encoder in contrast runs the spatial encoders only on the new tokens. Empirically this caching reduces per-step encoder time by $\sim$17\% on a Jetson Orin.

\paragraph{Implementation on edge devices}
Self-attention has $\mathcal{O}(N^2d)$ per-block cost~\cite{Vaswani2017AttentionIA}, where $N$ is the number of tokens and $d$ is the number of features. An early-fusion encoder pays this cost in every one of its $L$ blocks at the full multi-modal token budget ($N=5,264$ in this setup, left RGB camera, left event camera, LiDAR, IMU). Our late-fusion encoder restricts $N$-way attention to the $L_x=4$ cross-modal blocks. The $3$ spatial blocks only see per-modality per-frame tokens ($N_s \in\{144,256\}$) and the 2 temporal blocks see only 8 tokens for each modality (16 for the IMU).
As \cref{tab:hmae_latency} shows, this architectural choice gives a substantial advantage. Along with caching above, we can reduce latency by a factor of 1.70$\times$ at $B=1$, $T=8$ on an NVIDIA RTX 5090, with both encoders compiled using torch.compile with CUDA Graphs to amortize kernel launch overhead. Note that the comparison was performed at roughly similar encoder parameter counts, so this gain is due to the architecture.

We also perform experiments on a Jetson Orin NX 16GB edge processor. Our late-fusion MAE encoder can run with a latency of 112 ms. This is below the 200 ms window of perception measurements we were targeting in this study.

\begin{table}[h] 
  \centering
  \caption{Latency of the forward pass of the encoder for $B{=}1$, $T{=}8$ on an NVIDIA RTX 5090.
  Late-fusion MAE uses crossmodal\_depth$=$4 to match early-fusion MAE's parameter count
  (the late-fusion MAE used in the paper uses a crossmodal\_depth$=$3). All variants were compiled with \texttt{torch.compile(mode="reduce-overhead")} (CUDA Graphs). Wall-clock time
  measured with CUDA events over 20 iterations after 25 warmup steps.}
  \vspace*{1ex}
  \label{tab:hmae_latency}
  \renewcommand{\arraystretch}{1.25}
  \begin{adjustbox}{width=0.8\linewidth}
  \begin{tabular}{lrrr}
  \toprule
  Model & Enc.\ params (M) & Latency (ms) & Speedup \\
  \midrule
  Early-Fusion MAE (flat, $W{=}768$, $D{=}12$)        & 85.0 & 11.35 & 1.00$\times$ \\
  Late-Fusion\ MAE ($W{=}576$, $3{+}2{+}4$)           & 83.7 & 7.61  & 1.49$\times$ \\
  Late-Fusion\ MAE + streaming cache                  & 83.7 & 6.68  & 1.70$\times$ \\
  \bottomrule
  \end{tabular}
  \end{adjustbox}
  \end{table}

   \begin{table}[h]
    \centering
    \caption{Latency of the forward pass of the late-fusion MAE with $B{=}1$, $T{=}8$ on an
    NVIDIA Jetson Orin NX 16\,GB (MAXN power mode, clocks pinned via
    \texttt{jetson\_clocks}, Ampere GPU).
    Single-stream per-call latency measured with CUDA
    events (synchronized per call) over 20 iterations after 25 warmup steps.
    Late-fusion MAE configured with crossmodal\_depth of 4.
    Eager is the uncompiled baseline,
    \texttt{compile} refers to the model using \texttt{torch.compile(mode="reduce-overhead")}.}
    \vspace*{1ex}
    \label{tab:hmae_latency_jetson}
    \renewcommand{\arraystretch}{1.25}
    \begin{adjustbox}{width=0.85\linewidth}
    \begin{tabular}{lrrr}
    \toprule
    Model & Enc.\ params (M) & Eager (ms) & \texttt{compile} (ms) \\
    \midrule
    Late-Fusion\ MAE ($W{=}576$, $L=3{+}2{+}4$)        & 83.7 & 235 & 136 \\
    \quad + streaming cache                    & 83.7 & 214 & 112 \\
    Early-Fusion MAE ($W{=}768$, $L=12$)        & 85 & - & 288 \\
    \bottomrule
    \end{tabular}
    \end{adjustbox}
  \end{table}


\section{Analysis of the pre-trained MAE and predictions on dense perception tasks}
\label{app:dense_visualization}

\subsection{Reconstruction of masked sensor data}
\label{app:masked_recon}

\begin{figure}[!h]
\centering
\begin{subfigure}[b]{0.49\linewidth}
    \centering
    \includegraphics[width=\linewidth]{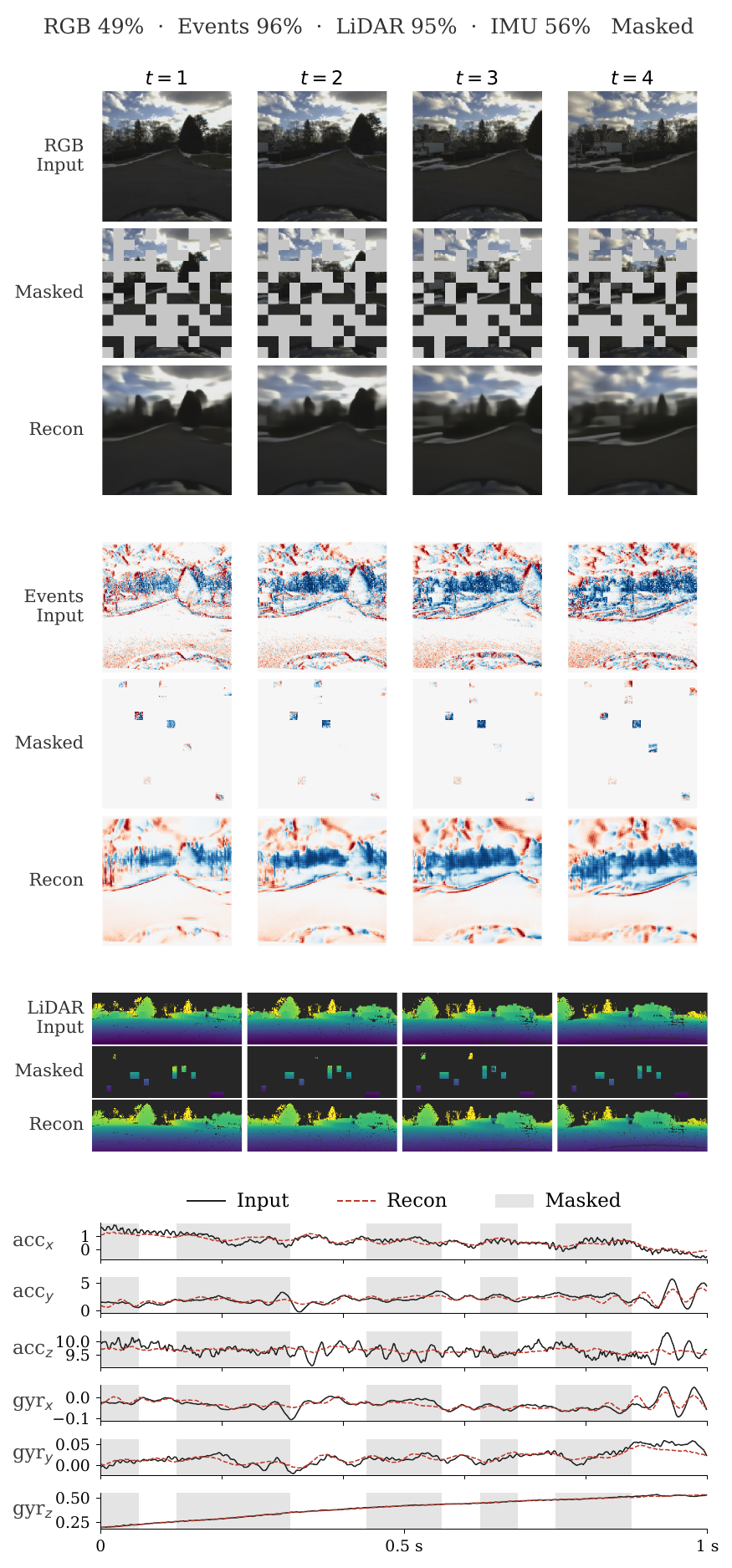}
    \caption{}
    \label{fig:event_masked}
\end{subfigure}
\begin{subfigure}[b]{0.49\linewidth}
    \centering
    \includegraphics[width=\linewidth]{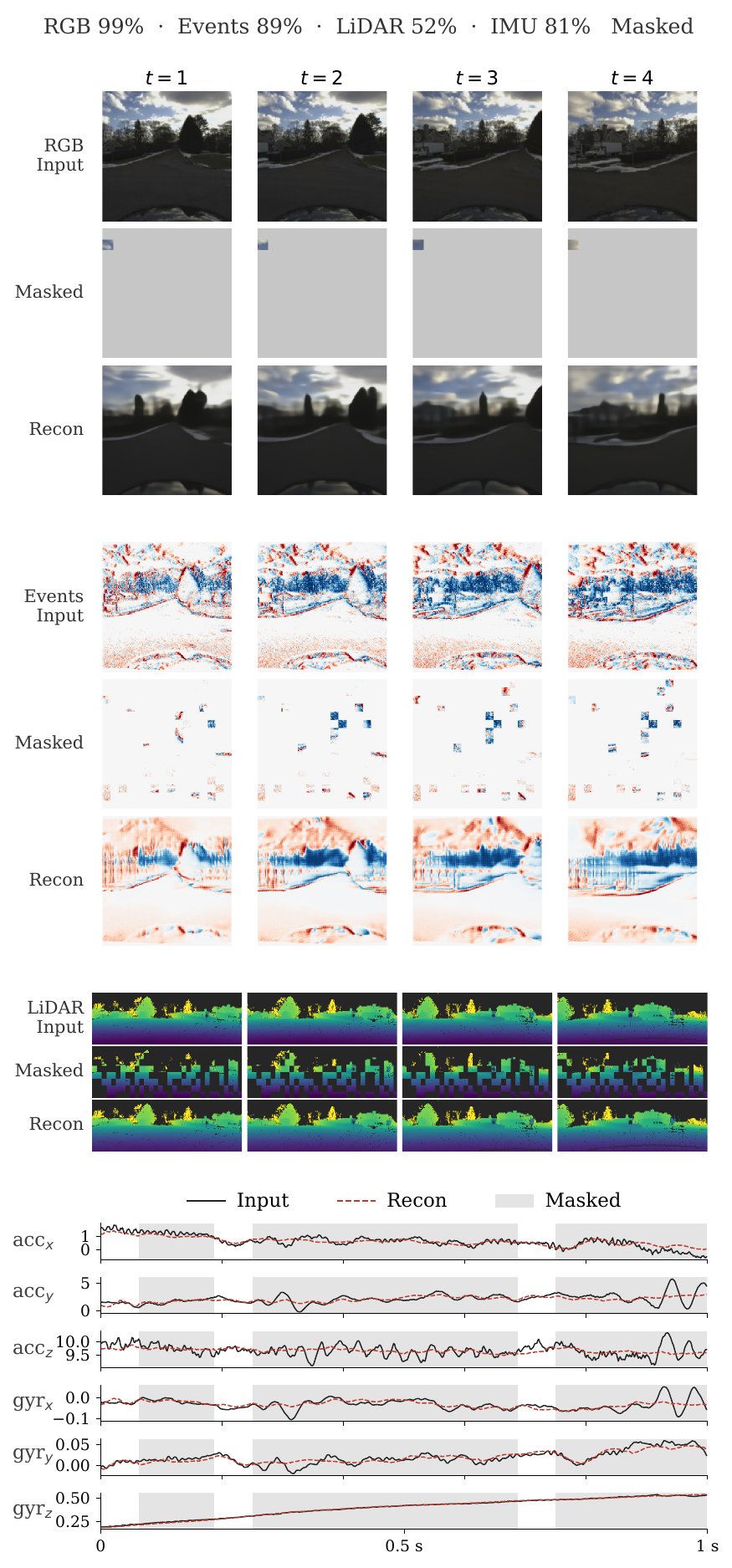}
    \caption{}
    \label{fig:rgb_masked}
\end{subfigure}
\caption{
Multi-modal reconstruction by predominantly masking event data (left panel) and RGB data (right panel).
}
\label{app:fig:maked_viz}
\end{figure}
\begin{figure}[!t]
\ContinuedFloat
\centering
\begin{subfigure}[b]{0.49\linewidth}
    \centering
    \includegraphics[width=\linewidth]{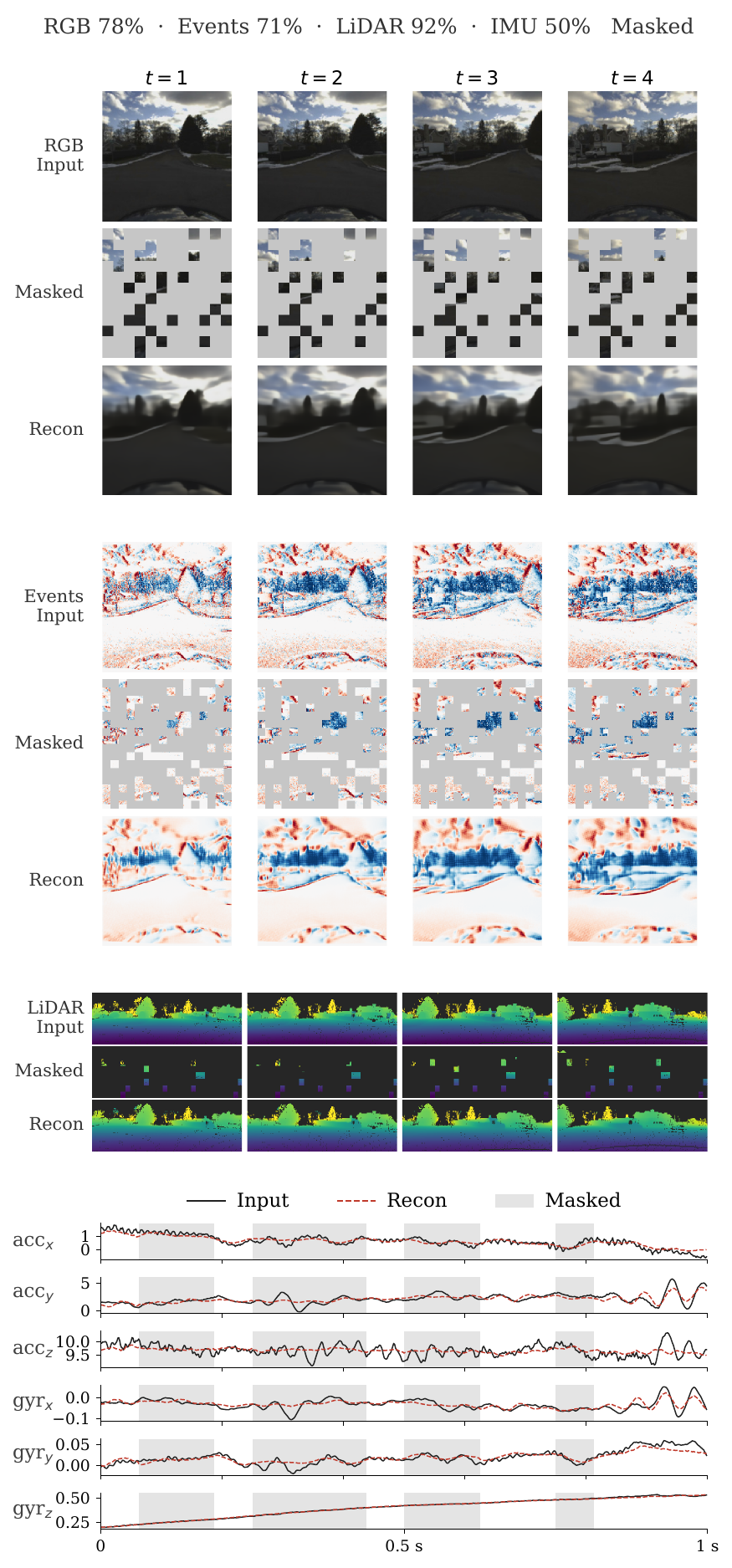}
    \caption{}
    \label{fig:lidar_masked}
\end{subfigure}
\begin{subfigure}[b]{0.49\linewidth}
    \centering
    \includegraphics[width=\linewidth]{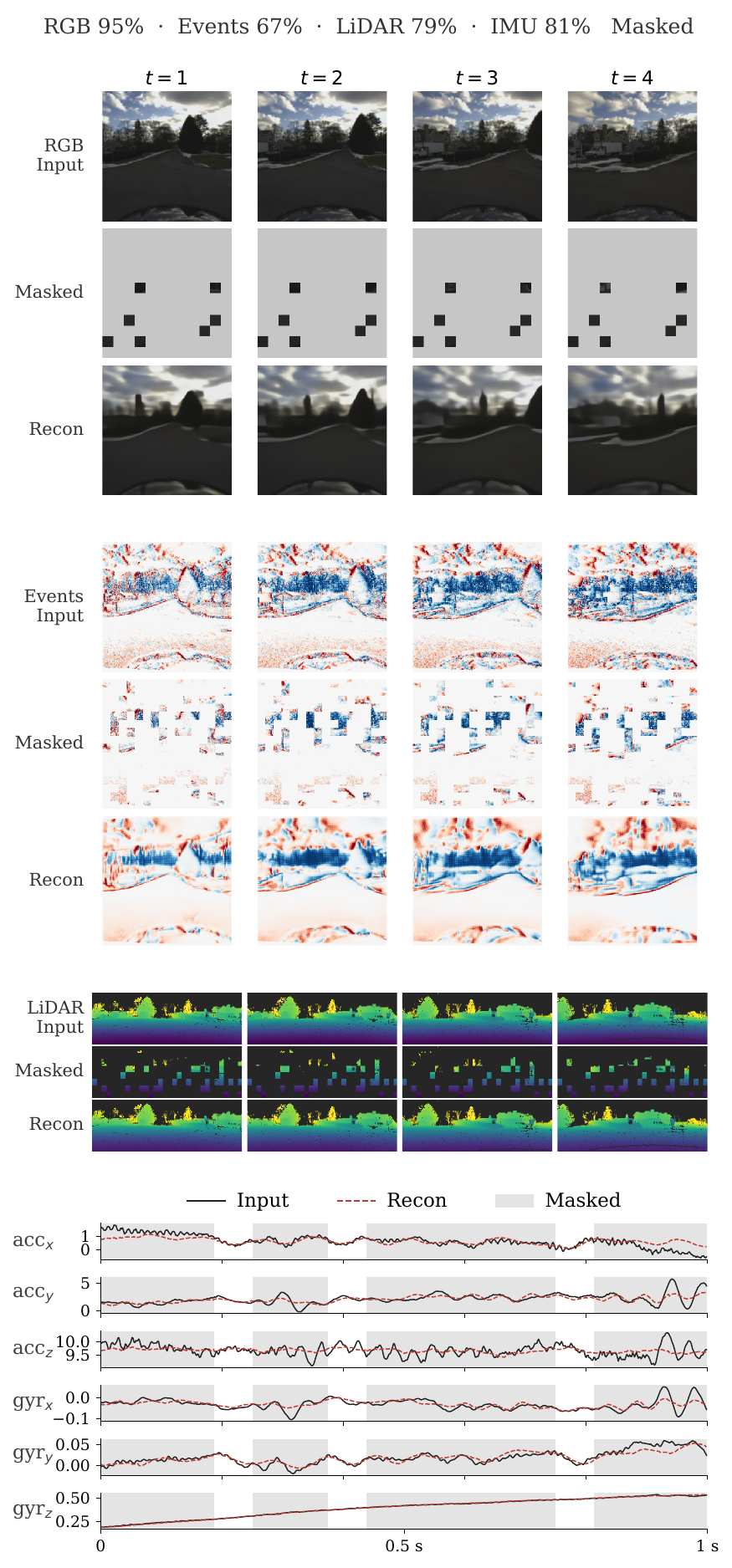}
    \caption{}
    \label{fig:imu_masked}
\end{subfigure}
\caption{
\textbf{Continued from the previous page.}
Multi-modal reconstruction by predominantly masking LiDAR data (left panel) and IMU data (right panel).
}
\end{figure}

In our paper, we trained various models to predict optical flow, semantic segmentation and depth using our procedure defined in \cref{app:eval}. In this section, we show a few examples of the dense perception outputs from the models we have tested.
Our late-fusion MAE learns inter-sensor correlations that enable it to reconstruct missing modalities. In this section, we present several examples of our approach reconstructing different masked configurations across modalities. This masked-sensor reconstruction can be valuable in robotics for handling sensor failure. For instance, if a perception algorithm relying on a particular sensor fails, the learned inter-sensor correlations can reconstruct the missing sensor data, and allow the perception algorithm to continue. We leave a full treatment of this application to future work, though we showcase an illustrative example of this missing sensor data reconstruction using our MAE. Furthermore, this approach can be readily extended to MaskGIT-style~\cite{Chang2022MaskGITMG} generative decoding, since we already have discrete latent tokenizers for each sensor.

\begin{figure}[!htpb]
    \centering
    \includegraphics[width=.95\linewidth]{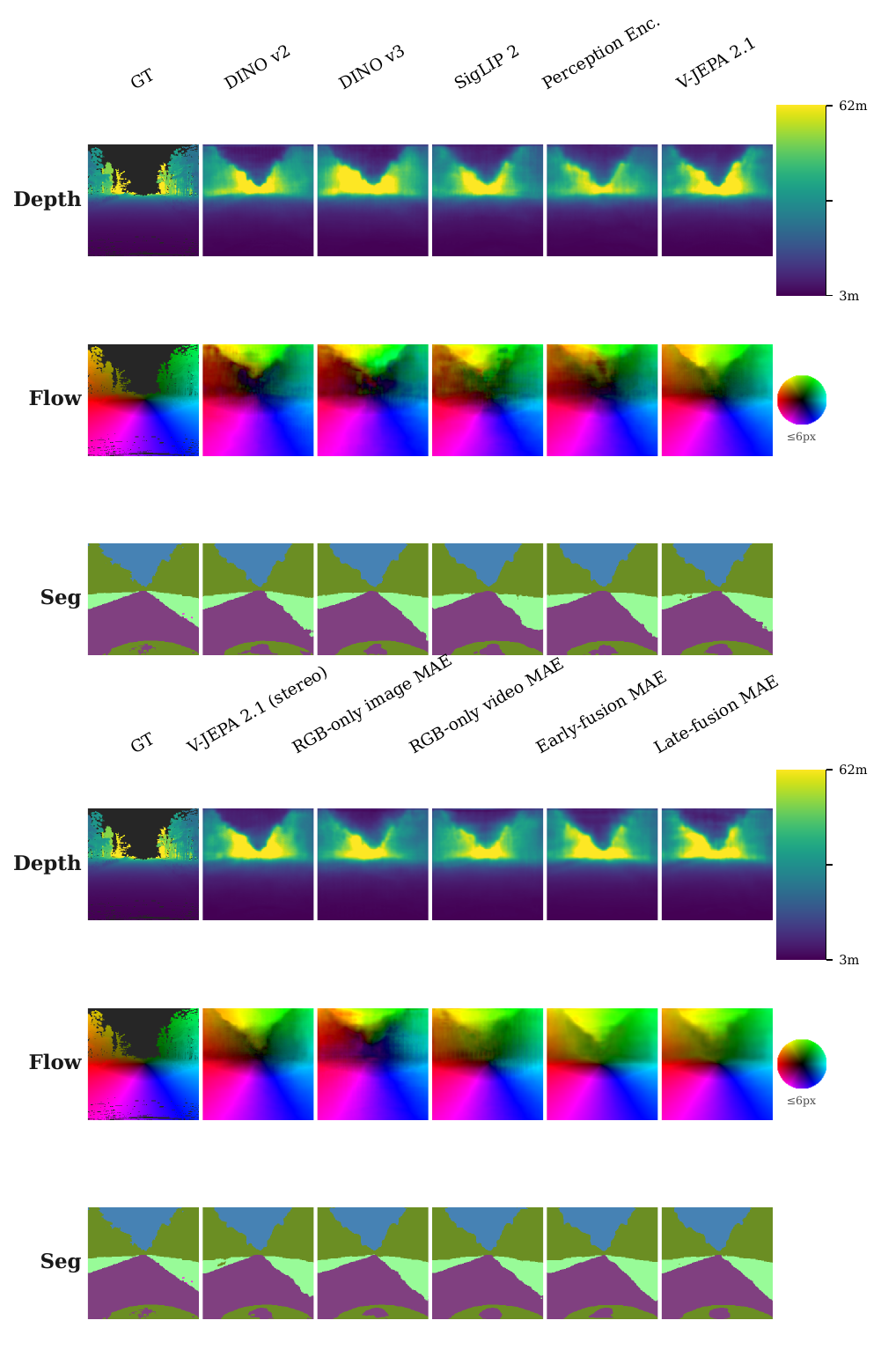}
    \caption{
    Predictions on dense prediction tasks for various models on the OctoSense test set: a sandy off-road region.
    }
    \label{fig:dense_perception_sand}
\end{figure}

\begin{figure}[!htpb]
\ContinuedFloat
    \centering
    \includegraphics[width=\linewidth]{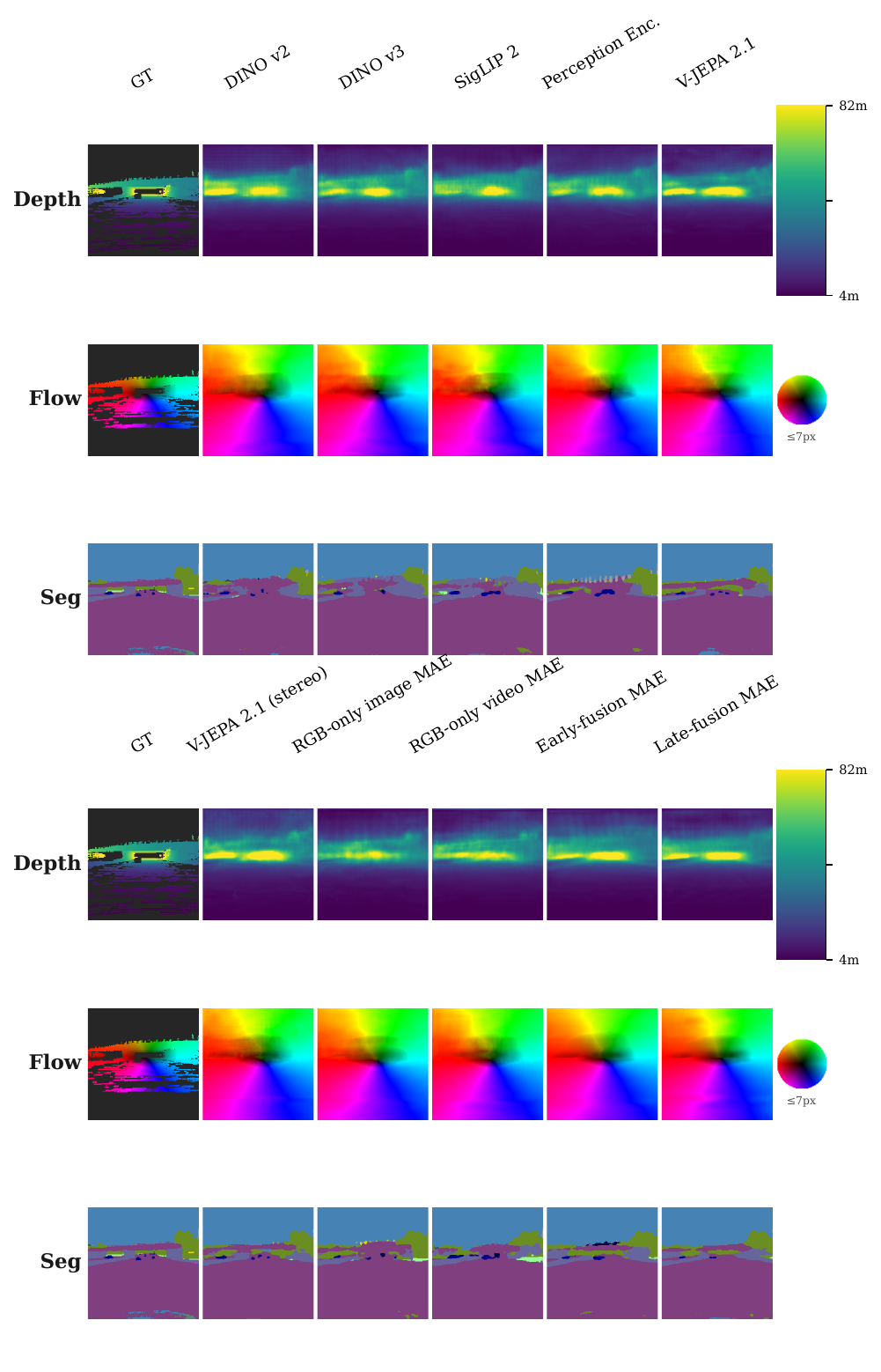}
    \caption{
    Predictions on dense prediction tasks for various models on the OctoSense test set: a highway setting.
    }
    \label{fig:dense_perception_highway}
\end{figure}

\begin{figure}[!htpb]
\ContinuedFloat
    \centering
    \includegraphics[width=\linewidth]{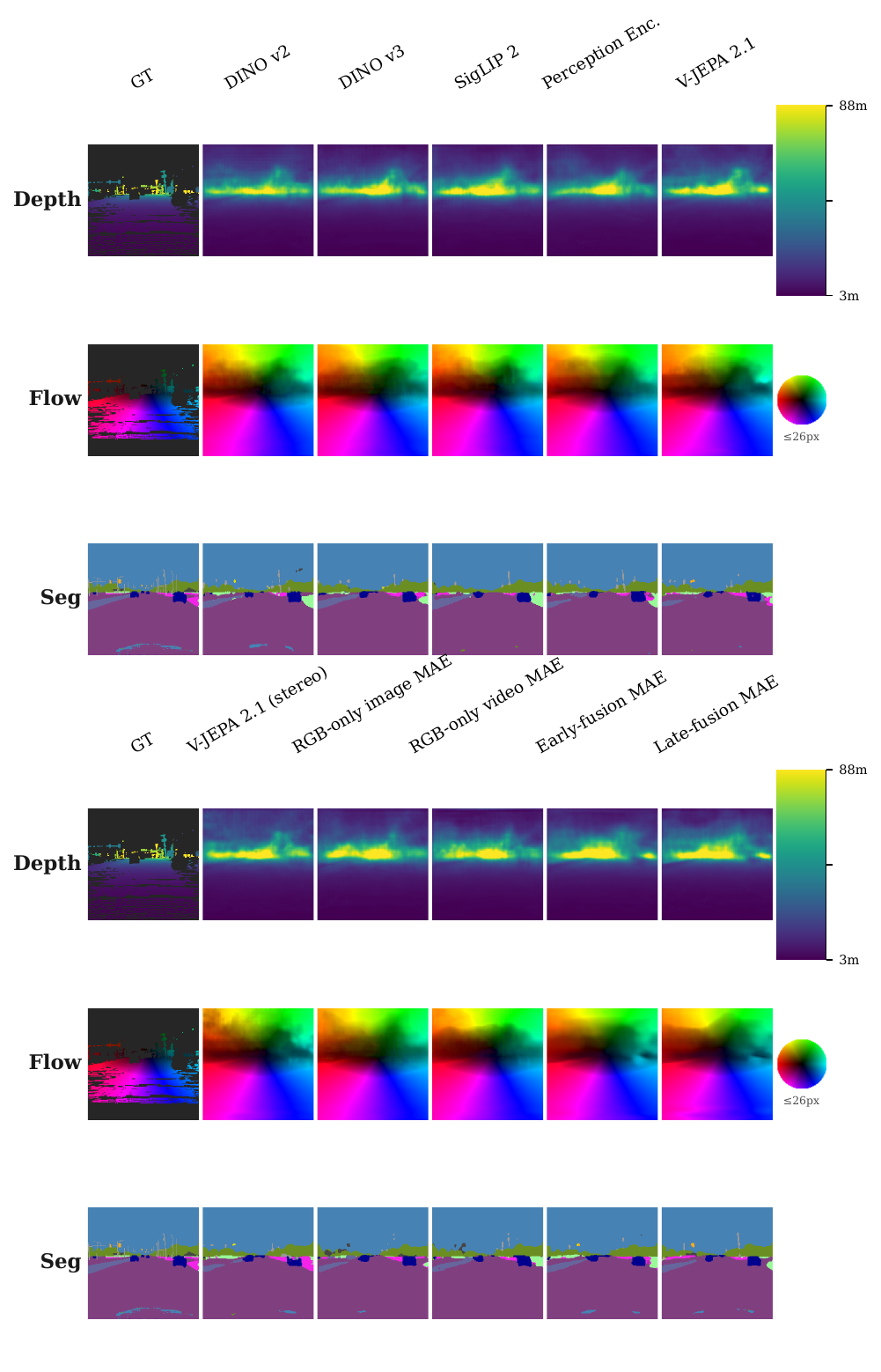}
    \caption{
    Predictions on dense prediction tasks for various models on the OctoSense test set: a city scene.
    }
    \label{fig:dense_perception_city}
\end{figure}

\clearpage
\subsection{Detailed metrics on dense perception tasks}
\label{app:error_analysis}

\begin{table}[h]
\centering
\caption{\textbf{Dense prediction task} — daytime (\texttt{test\_day}). All metrics reported on data where speed is at least 5~mph and described in \cref{app:eval}.}
\label{tab:dpt-test-day}
\vspace*{0.5ex}
\small
\renewcommand{\arraystretch}{1.1}
\begin{adjustbox}{max width=\linewidth}
\begin{tabular}{l rrrrrrrr}
\toprule
\textbf{Encoder} & \textbf{mIoU $\uparrow$} & \textbf{pix.\,acc.\,$\uparrow$} & \textbf{abs.\,rel.\,$\downarrow$} & \textbf{RMSE\,(m) $\downarrow$} & \textbf{SiLog $\downarrow$} & \textbf{$\delta\!<\!1.25$ $\uparrow$} & \textbf{EPE\,(px) $\downarrow$} & \textbf{outlier $\downarrow$} \\
\midrule
\textbf{Existing approaches}\\
DINOv2 & 0.410 & \textbf{0.921} & 0.10 & 6.82 & 0.17 & 0.879 & 19.46 & 0.54 \\
DINOv3 & 0.382 & 0.918 & 0.11 & 6.93 & 0.17 & 0.875 & 19.43 & 0.54 \\
SigLIP 2 & 0.377 & 0.912 & 0.11 & 7.22 & 0.18 & 0.864 & 20.22 & 0.55 \\
Perception Enc. & 0.388 & 0.915 & 0.11 & 7.12 & 0.18 & 0.867 & 19.90 & 0.56 \\
V-JEPA 2.1 & 0.402 & 0.920 & 0.09 & 6.38 & 0.16 & 0.896 & 9.13 & 0.31 \\
\addlinespace
\textbf{Our baselines}\\
V-JEPA 2.1 (stereo)$^*$ & 0.405 & 0.919 & 0.10 & 6.40 & 0.16 & 0.893 & 9.03 & 0.31 \\
RGB-only image MAE & 0.296 & 0.898 & 0.12 & 7.78 & 0.20 & 0.845 & 20.88 & 0.58 \\
RGB-only video MAE & 0.327 & 0.902 & 0.12 & 7.41 & 0.19 & 0.856 & 10.11 & 0.36 \\
Early-fusion MAE & 0.379 & 0.912 & 0.07 & 4.98 & \textbf{0.12} & 0.944 & 2.01 & \textbf{0.07} \\
\addlinespace
\textbf{Late-fusion MAE} & \textbf{0.411} & \textbf{0.921} & \textbf{0.06} & \textbf{4.73} & \textbf{0.12} & \textbf{0.950} & 1.97 & \textbf{0.07} \\
w/ full temporal attention & 0.405 & 0.920 & \textbf{0.06} & \textbf{4.73} & \textbf{0.12} & \textbf{0.950} & \textbf{1.96} & \textbf{0.07} \\
\addlinespace
\textbf{Leave one modality out}\\
w/o RGB & 0.366 & 0.906 & 0.06 & 4.70 & 0.12 & 0.950 & 2.02 & 0.07 \\
w/o EV & 0.403 & 0.920 & 0.07 & 4.88 & 0.12 & 0.947 & 2.00 & 0.07 \\
w/o LiDAR & 0.351 & 0.906 & 0.10 & 6.74 & 0.16 & 0.888 & 7.19 & 0.25 \\
w/o IMU & 0.402 & 0.919 & 0.06 & 4.67 & 0.11 & 0.950 & 2.05 & 0.08 \\
w/ only RGB & 0.310 & 0.896 & 0.11 & 7.20 & 0.18 & 0.867 & 11.31 & 0.36 \\
\bottomrule
\end{tabular}
\end{adjustbox}
\end{table}

\begin{table}[h]
\centering
\caption{\textbf{Dense prediction task} — nighttime (\texttt{test\_night}). All metrics reported on data where speed is at least 5~mph and described in \cref{app:eval}.}
\label{tab:dpt-test-night}
\vspace*{0.5ex}
\small
\renewcommand{\arraystretch}{1.1}
\begin{adjustbox}{max width=\linewidth}
\begin{tabular}{l rrrrrr}
\toprule
\textbf{Encoder} & \textbf{abs.\,rel.\,$\downarrow$} & \textbf{RMSE\,(m) $\downarrow$} & \textbf{SiLog $\downarrow$} & \textbf{$\delta\!<\!1.25$ $\uparrow$} & \textbf{EPE\,(px) $\downarrow$} & \textbf{outlier $\downarrow$} \\
\midrule
\textbf{Existing approaches}\\
DINOv2 & 0.13 & 9.95 & 0.22 & 0.828 & 18.36 & 0.54 \\
DINOv3 & 0.14 & 10.03 & 0.23 & 0.826 & 16.97 & 0.54 \\
SigLIP 2 & 0.14 & 10.36 & 0.24 & 0.817 & 18.12 & 0.55 \\
Perception Enc. & 0.14 & 10.20 & 0.23 & 0.819 & 17.84 & 0.56 \\
V-JEPA 2.1 & 0.13 & 9.51 & 0.22 & 0.836 & 10.90 & 0.39 \\
\addlinespace
\textbf{Our baselines}\\
V-JEPA 2.1 (stereo)$^*$ & 0.13 & 9.51 & 0.22 & 0.834 & 10.55 & 0.39 \\
RGB-only image MAE & 0.15 & 10.58 & 0.24 & 0.810 & 18.98 & 0.56 \\
RGB-only video MAE & 0.14 & 10.20 & 0.24 & 0.816 & 11.56 & 0.42 \\
Early-fusion MAE & \textbf{0.07} & 6.30 & 0.14 & 0.934 & 2.45 & \textbf{0.09} \\
\addlinespace
\textbf{Late-fusion MAE} & \textbf{0.07} & 6.08 & \textbf{0.13} & \textbf{0.939} & 2.39 & \textbf{0.09} \\
w/ full temporal attention & \textbf{0.07} & \textbf{6.07} & \textbf{0.13} & \textbf{0.939} & \textbf{2.38} & \textbf{0.09} \\
\addlinespace
\textbf{Leave one modality out}\\
w/o RGB & 0.07 & 5.99 & 0.13 & 0.941 & 2.43 & 0.09 \\
w/o EV & 0.07 & 6.32 & 0.13 & 0.935 & 2.42 & 0.09 \\
w/o LiDAR & 0.13 & 9.55 & 0.21 & 0.839 & 8.05 & 0.31 \\
w/o IMU & 0.07 & 6.00 & 0.13 & 0.940 & 2.49 & 0.09 \\
w/ only RGB & 0.14 & 10.09 & 0.23 & 0.819 & 12.81 & 0.43 \\
\bottomrule
\end{tabular}
\end{adjustbox}
\end{table}

\begin{table}[h]
\centering
\caption{\textbf{Dense prediction task} — sensor-degraded (\texttt{test\_degraded}). All metrics reported on data where speed is at least 5~mph and described in \cref{app:eval}.}
\label{tab:dpt-test-degraded}
\vspace*{0.5ex}
\small
\renewcommand{\arraystretch}{1.1}
\begin{adjustbox}{max width=\linewidth}
\begin{tabular}{l rrrrrr}
\toprule
\textbf{Encoder} & \textbf{abs.\,rel.\,$\downarrow$} & \textbf{RMSE\,(m) $\downarrow$} & \textbf{SiLog $\downarrow$} & \textbf{$\delta\!<\!1.25$ $\uparrow$} & \textbf{EPE\,(px) $\downarrow$} & \textbf{outlier $\downarrow$} \\
\midrule
\textbf{Existing approaches}\\
DINOv2 & 0.14 & 7.41 & 0.23 & 0.823 & 19.97 & 0.61 \\
DINOv3 & 0.14 & 7.49 & 0.23 & 0.822 & 18.82 & 0.60 \\
SigLIP 2 & 0.15 & 7.55 & 0.23 & 0.816 & 19.02 & 0.61 \\
Perception Enc. & 0.15 & 7.56 & 0.23 & 0.816 & 19.52 & 0.61 \\
V-JEPA 2.1 & 0.14 & 7.40 & 0.23 & 0.827 & 10.85 & 0.45 \\
\addlinespace
\textbf{Our baselines}\\
V-JEPA 2.1 (stereo)$^*$ & 0.14 & 7.36 & 0.22 & 0.828 & 9.22 & 0.42 \\
RGB-only image MAE & 0.15 & 7.90 & 0.24 & 0.807 & 20.84 & 0.61 \\
RGB-only video MAE & 0.15 & 7.59 & 0.24 & 0.815 & 11.38 & 0.47 \\
Early-fusion MAE & \textbf{0.07} & 4.99 & 0.14 & 0.932 & \textbf{2.10} & \textbf{0.08} \\
\addlinespace
\textbf{Late-fusion MAE} & \textbf{0.07} & \textbf{4.77} & \textbf{0.13} & \textbf{0.939} & 2.12 & \textbf{0.08} \\
w/ full attention & \textbf{0.07} & 4.79 & \textbf{0.13} & \textbf{0.939} & 2.15 & \textbf{0.08} \\
\addlinespace
\textbf{Leave one modality out}\\
w/o RGB & 0.07 & 4.76 & 0.13 & 0.941 & 2.15 & 0.08 \\
w/o EV & 0.07 & 4.87 & 0.14 & 0.935 & 2.18 & 0.08 \\
w/o LiDAR & 0.13 & 7.27 & 0.22 & 0.837 & 6.75 & 0.33 \\
w/o IMU & 0.07 & 4.75 & 0.13 & 0.939 & 2.24 & 0.09 \\
w/ only RGB & 0.15 & 7.76 & 0.24 & 0.812 & 12.68 & 0.48 \\
\bottomrule
\end{tabular}
\end{adjustbox}
\end{table}

\begin{table}[h]
\centering
\caption{\textbf{Ego-motion} — daytime (\texttt{test\_day}). All metrics reported on data where speed is at least 5~mph and described in \cref{app:eval}.}
\label{tab:ego-test-day}
\vspace*{0.5ex}
\small
\renewcommand{\arraystretch}{1.1}
\begin{adjustbox}{max width=\linewidth}
\begin{tabular}{l rrrrrr}
\toprule
\textbf{Encoder} & \textbf{lin.\,vel.\,(m/s) $\downarrow$} & \textbf{ang.\,vel.\,(rad/s) $\downarrow$} & \textbf{trans.\,(m) $\downarrow$} & \textbf{rot.\,($^\circ$) $\downarrow$} & \textbf{speed (mph) $\downarrow$} & \textbf{steer ($^\circ$) $\downarrow$} \\
\midrule
\textbf{Existing approaches}\\
DINOv2 & 4.92 & 0.07 & 0.96 & 0.80 & 5.84 & 22.62 \\
DINOv3 & 4.54 & 0.07 & 0.93 & 0.79 & 5.50 & 25.24 \\
SigLIP 2 & 4.67 & 0.07 & 0.98 & 0.95 & 6.59 & 36.12 \\
Perception Enc. & 4.68 & 0.09 & 0.96 & 0.94 & 6.42 & 35.62 \\
V-JEPA 2.1 & 3.94 & 0.05 & 0.77 & 0.47 & 2.90 & 5.98 \\
\addlinespace
\textbf{Our baselines}\\
V-JEPA 2.1 (stereo)$^*$ & 3.92 & 0.05 & 0.77 & 0.48 & 2.87 & 5.69 \\
RGB-only image MAE & 4.87 & 0.09 & 0.92 & 0.88 & 6.26 & 31.57 \\
RGB-only video MAE & 3.99 & 0.05 & 0.76 & 0.50 & 2.56 & 12.96 \\
Early-fusion MAE & 0.68 & 0.05 & 0.14 & 0.56 & 1.18 & 4.58 \\
\addlinespace
\textbf{Late-fusion MAE} & \textbf{0.36} & \textbf{0.01} & 0.06 & 0.24 & \textbf{0.78} & 2.51 \\
w/ full temporal attention & \textbf{0.36} & 0.02 & \textbf{0.05} & \textbf{0.17} & 0.98 & \textbf{2.46} \\
\addlinespace
\textbf{Leave one modality out}\\
w/o RGB & 0.36 & 0.01 & 0.06 & 0.24 & 0.84 & 2.65 \\
w/o EV & 0.35 & 0.01 & 0.05 & 0.21 & 0.82 & 2.59 \\
w/o LiDAR & 3.74 & 0.02 & 0.73 & 0.46 & 1.45 & 3.97 \\
w/o IMU & 0.38 & 0.02 & 0.06 & 0.24 & 0.90 & 2.78 \\
w/ only RGB & 4.04 & 0.05 & 0.83 & 0.47 & 2.65 & 8.29 \\
\bottomrule
\end{tabular}
\end{adjustbox}
\end{table}

\begin{table}[h]
\centering
\caption{\textbf{Ego-motion} — nighttime (\texttt{test\_night}). All metrics reported on data where speed is at least 5~mph and described in \cref{app:eval}.}
\label{tab:ego-test-night}
\vspace*{0.5ex}
\small
\renewcommand{\arraystretch}{1.1}
\begin{adjustbox}{max width=\linewidth}
\begin{tabular}{l rrrrrr}
\toprule
\textbf{Encoder} & \textbf{lin.\,vel.\,(m/s) $\downarrow$} & \textbf{ang.\,vel.\,(rad/s) $\downarrow$} & \textbf{trans.\,(m) $\downarrow$} & \textbf{rot.\,($^\circ$) $\downarrow$} & \textbf{speed (mph) $\downarrow$} & \textbf{steer ($^\circ$) $\downarrow$} \\
\midrule
\textbf{Existing approaches}\\
DINOv2 & 4.22 & 0.06 & 0.84 & 0.75 & 5.79 & 23.11 \\
DINOv3 & 4.00 & 0.07 & 0.79 & 0.76 & 5.36 & 21.93 \\
SigLIP 2 & 4.47 & 0.06 & 0.78 & 0.94 & 6.92 & 36.00 \\
Perception Enc. & 4.09 & 0.08 & 0.91 & 0.93 & 6.76 & 34.54 \\
V-JEPA 2.1 & 3.81 & 0.05 & 0.76 & 0.47 & 3.97 & 6.83 \\
\addlinespace
\textbf{Our baselines}\\
V-JEPA 2.1 (stereo)$^*$ & 3.62 & 0.05 & 0.71 & 0.47 & 3.73 & 6.47 \\
RGB-only image MAE & 4.53 & 0.09 & 1.05 & 0.77 & 5.17 & 21.39 \\
RGB-only video MAE & 3.61 & 0.05 & 0.72 & 0.50 & 3.06 & 12.45 \\
Early-fusion MAE & 0.69 & 0.05 & 0.15 & 0.53 & 1.75 & 4.48 \\
\addlinespace
\textbf{Late-fusion MAE} & \textbf{0.34} & \textbf{0.01} & 0.06 & 0.23 & \textbf{0.93} & 2.27 \\
w/ full attention & 0.35 & 0.02 & \textbf{0.05} & \textbf{0.17} & 1.58 & \textbf{2.25} \\
\addlinespace
\textbf{Leave one modality out}\\
w/o RGB & 0.33 & 0.01 & 0.06 & 0.23 & 1.09 & 2.34 \\
w/o EV & 0.35 & 0.02 & 0.05 & 0.20 & 1.11 & 2.42 \\
w/o LiDAR & 2.88 & 0.02 & 0.57 & 0.45 & 1.79 & 4.00 \\
w/o IMU & 0.37 & 0.02 & 0.06 & 0.22 & 1.36 & 2.61 \\
w/ only RGB & 3.32 & 0.05 & 0.77 & 0.47 & 3.36 & 7.75 \\
\bottomrule
\end{tabular}
\end{adjustbox}
\end{table}

\begin{table}[h]
\centering
\caption{\textbf{Ego-motion} — sensor-degraded (\texttt{test\_degraded}). All metrics reported on data where speed is at least 5~mph and described in \cref{app:eval}.}
\label{tab:ego-test-degraded}
\vspace*{0.5ex}
\small
\renewcommand{\arraystretch}{1.1}
\begin{adjustbox}{max width=\linewidth}
\begin{tabular}{l rrrrrr}
\toprule
\textbf{Encoder} & \textbf{lin.\,vel.\,(m/s) $\downarrow$} & \textbf{ang.\,vel.\,(rad/s) $\downarrow$} & \textbf{trans.\,(m) $\downarrow$} & \textbf{rot.\,($^\circ$) $\downarrow$} & \textbf{speed (mph) $\downarrow$} & \textbf{steer ($^\circ$) $\downarrow$} \\
\midrule
\textbf{Existing approaches}\\
DINOv2 & 3.13 & 0.05 & 0.65 & 0.58 & 7.20 & 24.59 \\
DINOv3 & 3.34 & 0.06 & 0.65 & 0.61 & 7.14 & 30.19 \\
SigLIP 2 & 3.48 & 0.06 & 0.71 & 0.70 & 8.51 & 26.04 \\
Perception Enc. & 3.70 & 0.07 & 0.72 & 0.71 & 8.23 & 32.81 \\
V-JEPA 2.1 & 2.47 & 0.04 & 0.51 & 0.43 & 5.09 & 8.66 \\
\addlinespace
\textbf{Our baselines}\\
V-JEPA 2.1 (stereo)$^*$ & 2.20 & 0.04 & 0.44 & 0.41 & 4.49 & 8.14 \\
RGB-only image MAE & 3.71 & 0.07 & 0.69 & 0.65 & 7.93 & 24.81 \\
RGB-only video MAE & 2.48 & 0.04 & 0.45 & 0.45 & 4.14 & 10.67 \\
Early-fusion MAE & 0.75 & 0.04 & 0.14 & 0.49 & 1.12 & 7.10 \\
\addlinespace
\textbf{Late-fusion MAE} & \textbf{0.32} & \textbf{0.01} & 0.06 & 0.23 & 0.62 & \textbf{2.76} \\
w/ full attention & \textbf{0.32} & 0.02 & \textbf{0.05} & \textbf{0.17} & \textbf{0.56} & 2.88 \\
\addlinespace
\textbf{Leave one modality out}\\
w/o RGB & 0.31 & 0.01 & 0.06 & 0.23 & 0.62 & 2.99 \\
w/o EV & 0.33 & 0.01 & 0.06 & 0.21 & 0.66 & 2.87 \\
w/o LiDAR & 1.52 & 0.02 & 0.34 & 0.40 & 2.10 & 4.95 \\
w/o IMU & 0.35 & 0.02 & 0.06 & 0.21 & 0.65 & 3.18 \\
w/ only RGB & 2.61 & 0.04 & 0.54 & 0.42 & 4.35 & 8.60 \\
\bottomrule
\end{tabular}
\end{adjustbox}
\end{table}

\begin{table}[h]
\centering
\small
\caption{Per-class segmentation IoU on \texttt{test\_day} (motion-filtered $\geq 5$\,mph), sorted by class pixel frequency. $\Delta =$ DINOv2 $-$ Ours. Late-Fusion MAE is competitive with DINOv2 on frequent classes (road, sky, vegetation, building, car); the deficit (\textbf{bold}) is mostly rare classes ($<1\%$ of pixels): poles, buses, and persons. }
\vspace*{1ex}
\label{tab:seg_per_class}
\renewcommand{\arraystretch}{1.25}
\begin{adjustbox}{width=0.75\linewidth}
\begin{tabular}{l r r r r}
\toprule
\textbf{Class} & \textbf{Freq.\ (\%)} & \textbf{Ours (mIoU)} & \textbf{DINOv2 (mIoU)} & \textbf{$\Delta$} \\
\midrule
road & 42.76 & 0.928 & 0.924 & -0.004 \\
sky & 21.02 & 0.934 & 0.939 & +0.005 \\
vegetation & 19.66 & 0.833 & 0.838 & +0.005 \\
building & 5.85 & 0.690 & 0.708 & +0.018 \\
car & 3.34 & 0.684 & 0.684 & +0.000 \\
terrain & 2.62 & 0.595 & 0.599 & +0.004 \\
sidewalk & 1.70 & 0.426 & 0.419 & -0.007 \\
wall & 1.05 & 0.529 & 0.490 & -0.039 \\
\textbf{pole} & 0.75 & 0.290 & 0.318 & \textbf{+0.027} \\
fence & 0.44 & 0.302 & 0.248 & -0.054 \\
truck & 0.25 & 0.496 & 0.462 & -0.034 \\
traffic sign & 0.23 & 0.416 & 0.414 & -0.002 \\
\textbf{bus} & 0.19 & 0.379 & 0.426 & \textbf{+0.048} \\
traffic light & 0.08 & 0.269 & 0.215 & -0.054 \\
\textbf{person} & 0.04 & 0.026 & 0.099 & \textbf{+0.073} \\
\midrule
\textbf{Overall mIoU} & & 0.411 & 0.410 & -0.001 \\
\bottomrule
\end{tabular}
\end{adjustbox}
\end{table}

\begin{figure}[h]
    \centering
    \includegraphics[width=\linewidth]{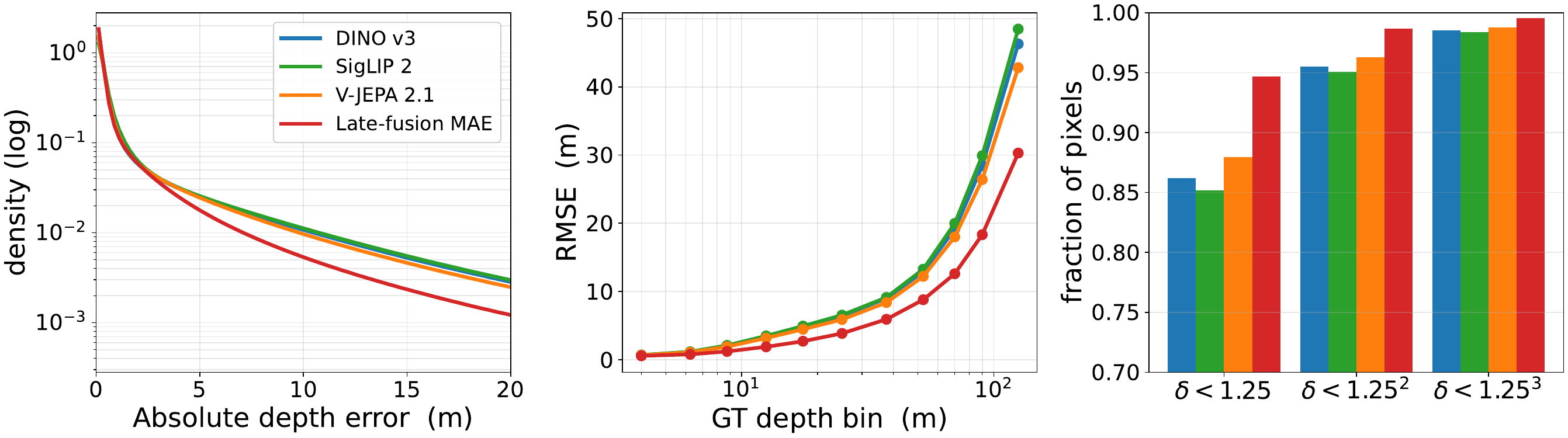}
    \caption{Depth Estimation error analysis figures across models on OctoSense test-set speed filtered~(5 mph):  (Left) Log-density of the per-pixel absolute depth error, (Middle) RMSE conditioned on ground truth magnitude, (Right) Threshold accuracy $\delta <1.25^k$ for $k= 1, 2, 3$  (\cref{app:eval}). Our late-fusion MAE has the smallest depth errors across the entire depth range and the highest threshold accuracy at every $\delta$. }
    \label{fig:error_depth_viz}
\end{figure}

\begin{figure}[h]
    \centering
    \includegraphics[width=\linewidth]{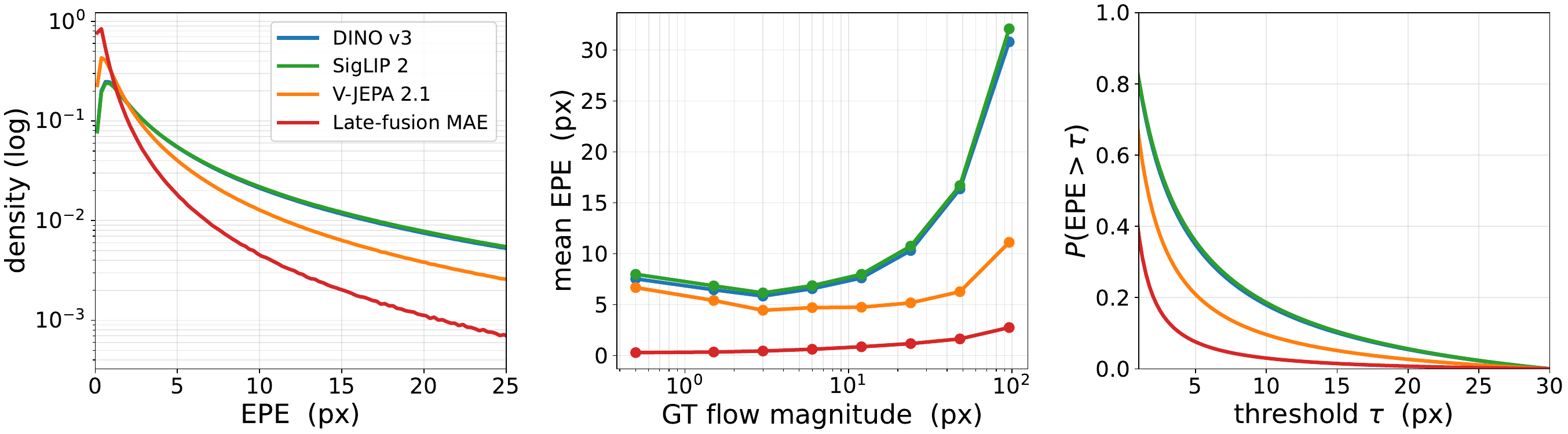}
    \caption{Optical Flow error analysis across models on OctoSense test-set speed filtered~(5 mph): (Left) Log-density of the per-pixel end-point error (EPE). (Middle) Mean EPE conditioned on the ground-truth flow magnitude. (Right) Outlier rate $P(\text{EPE}> \tau)$ swept over threshold $\tau$. Our late-fusion MAE attains lower mean EPE at every motion magnitude and a steeper initial decay than every baseline. }
    \label{fig:error_flow_viz}
\end{figure}

\begin{figure}[!htpb]
    \centering
    \includegraphics[width=\linewidth]{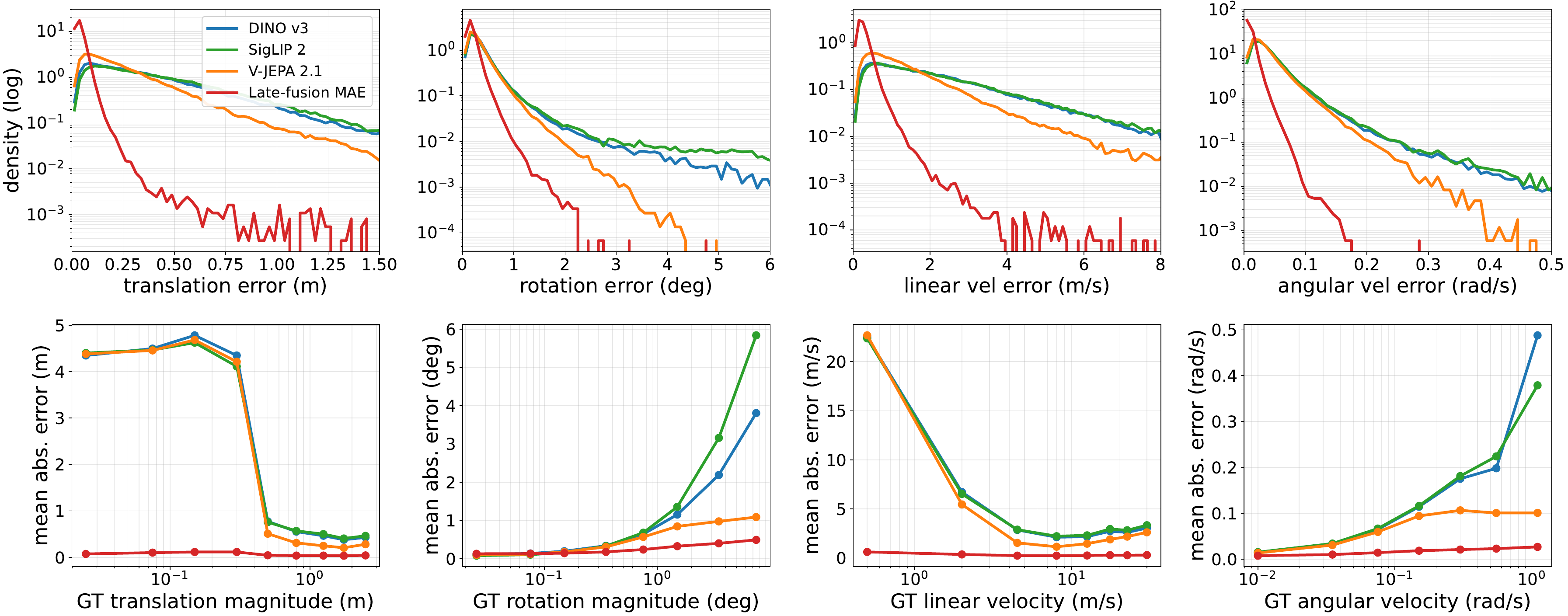}
    \caption{Ego-motion estimation analysis figures across models on OctoSense test-set speed filtered: (Top) Log-density of the error magnitudes for relative translation, relative rotation, linear velocity and angular velocity. (Bottom) Mean error conditioned on the ground-truth magnitude for each quantity. Our late-fusion MAE stays near zero translation error across the full motion range. RGB-only baselines degrade sharply under fast rotational motion.  }
    \label{fig:error_ego_viz}
\end{figure}

\begin{figure}[!htpb]
    \centering
    \includegraphics[width=0.6\linewidth]{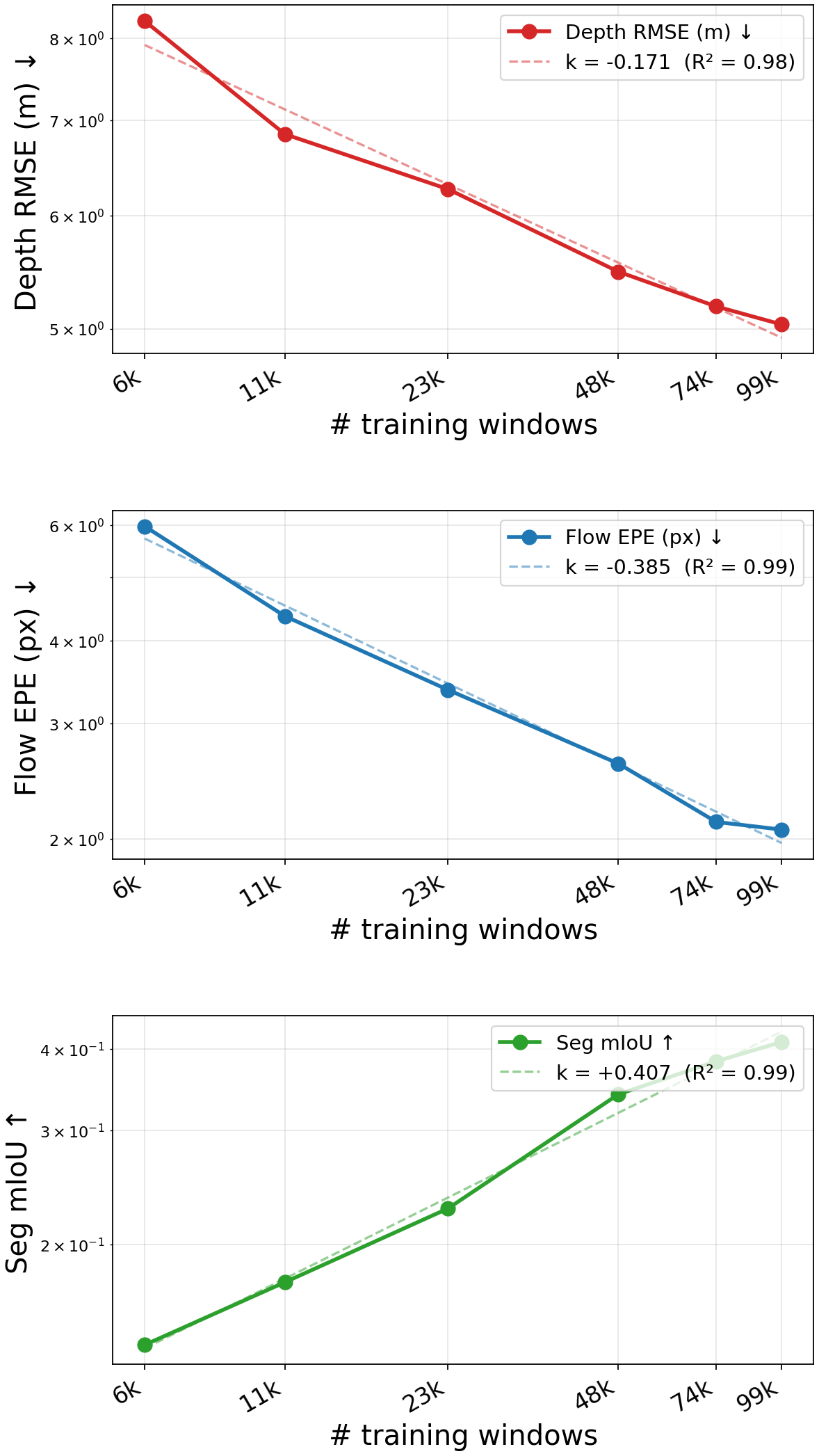}
    \caption{\textbf{Scaling laws on perception tasks on OctoSense test data.}
    Depth root-mean-square-error~(RMSE) in meters, optical-flow mean end-point error in pixels, and segmentation mIoU as a function of the amount of non-overlapping 1.4 s training windows for our late-fusion MAE evaluated on the full held-out test set (motion-filtered to ground truth speed above 5 mph). Both pre-training and probe training were performed using varying amounts of training data. We fit a power law: $y=Ax^k$, which is a linear trend in $\log(y)=\log(A)+k\log(x)$  in log-space.
    All three plots show that error improves monotonically with data.
    This motivates the need for large-scale datasets such as OctoSense. All other pretraining and probe hyper-parameters follow \cref{tab:hparams_late_fusion,tab:probe_arch}.
}
    \label{fig:scaling_laws}
\end{figure}
\end{document}